\RequirePackage{fix-cm}
\documentclass[smallextended]{svjour3}       % onecolumn (second format)
\smartqed  % flush right qed marks, e.g. at end of proof
\usepackage{graphicx}
\usepackage{fancyhdr}
\usepackage[utf8]{inputenc} % allow utf-8 input
\usepackage[T1]{fontenc}    % use 8-bit T1 fonts
\usepackage{hyperref}       % hyperlinks
\usepackage{url}            % simple URL typesetting
\usepackage{booktabs}       % professional-quality tables
\usepackage{amsfonts}       % blackboard math symbols
\usepackage{nicefrac}       % compact symbols for 1/2, etc.
\usepackage{microtype}      % microtypography
\usepackage{xcolor}         % colors
\usepackage{threeparttable}

\usepackage{amsmath}
\usepackage{amssymb}

\usepackage{sidecap}

\usepackage{tabularx}
\usepackage{multirow}
\usepackage{hhline}
\usepackage{algcompatible}
\usepackage{algorithm}
\usepackage{enumitem}
\usepackage{multicol}
\usepackage{bm}

\usepackage{epstopdf}
\AppendGraphicsExtensions{.tiff}

\def\va{{\bm{a}}}
\def\vb{{\bm{b}}}

\def\vf{{\bm{f}}}

\def\vh{{\bm{h}}}

\def\vm{{\bm{m}}}

\def\vu{{\bm{u}}}
\def\vv{{\bm{v}}}

\def\vx{{\bm{x}}}

\def\mA{{\bm{A}}}
\def\mB{{\bm{B}}}

\def\mI{{\bm{I}}}

\def\mK{{\bm{K}}}

\def\mM{{\bm{M}}}

\def\mU{{\bm{U}}}
\def\mV{{\bm{V}}}

\def\mX{{\bm{X}}}
\def\mY{{\bm{Y}}}

\def \RR {{\mathbb{R}}}

\newcommand{\R}{\mathbb{R}}
\newcommand*{\horzbar}{\rule[.5ex]{2.5ex}{0.5pt}}

\usepackage{verbatim}
\usepackage{wrapfig}

\newcommand{\bs}[1]{\boldsymbol #1}
\newcommand{\ppx}[1]{\frac{\partial}{\partial #1}}
\newcommand{\pfpx}[2]{\frac{\partial #1}{\partial #2}}
\newcommand{\bp}{\begin{pmatrix}}
\newcommand{\ep}{\end{pmatrix}}

\begin{document}

\title{Learning Proper Orthogonal Decomposition 
%POD 
of Complex Dynamics Using Heavy-ball Neural ODEs
}
%\subtitle{Do you have a subtitle?\\ If so, write it here}

%\titlerunning{Short form of title}        % if too long for running head

\author{Justin Baker\and
        Elena Cherkaev\and%etc.
        Akil Narayan\and
        Bao Wang
}

%\authorrunning{Short form of author list} % if too long for running head

\institute{J. Baker, E. Cherkaev \at
              Department of Mathematics\\
              University of Utah\\
              %Tel.: +801 581 6851\\
              %Fax: +123-45-678910\\
              \email{baker@math.utah.edu}, \email{elena@math.utah.edu}           %  \\
%             \emph{Present address:} of F. Author  %  if needed
%           \and
%           E. Cherkaev \at
%              Department of Mathematics\\
%              University of Utah\\
%              \email{elena@math.utah.edu}
           \and
           A. Narayan, B. Wang \at
              Department of Mathematics, and Scientific Computing and Imaging (SCI) Institute\\
              University of Utah\\
              \email{akil@sci.utah.edu}, \email{bwang@math.utah.edu}
        %\and
        %Please correspond to Bao Wang. Email: bwang@math.utah.edu
%           \and
%           B. Wang \at
%              Department of Mathematics, and Scientific Computing and Imaging (SCI) Institute\\
%              University of Utah\\
%              \email{bwang@math.utah.edu}
}

\date{Received: date / Accepted: date}
% The correct dates will be entered by the editor

\maketitle

\begin{abstract}
Proper orthogonal decomposition (POD) allows reduced-order modeling of complex dynamical systems at a substantial level, while maintaining a high degree of accuracy in modeling the underlying dynamical systems. Advances in machine learning algorithms enable learning POD-based dynamics from data and making accurate and fast predictions of dynamical systems. This paper extends the recently proposed heavy-ball neural ODEs (HBNODEs) [Xia et al. NeurIPS, 2021] for learning data-driven reduced-order models (ROMs) in the POD context, in particular, for learning dynamics of time-varying coefficients generated by the POD analysis on training snapshots constructed by solving full-order models. HBNODE enjoys several practical advantages for learning POD-based ROMs with theoretical guarantees, including 1) HBNODE can learn long-range dependencies effectively from sequential observations, which is crucial for learning intrinsic patterns from sequential data, and 2) HBNODE is computationally efficient in both training and testing. We compare HBNODE with other popular ROMs on several complex dynamical systems, including the von K\'{a}rm\'{a}n Street flow, the Kurganov-Petrova-Popov equation, and the one-dimensional Euler equations for fluids modeling. 
\keywords{Neural ODE \and Momentum \and Reduced-order modeling \and Deep learning}
\subclass{65P99 \and 68T07}
\end{abstract}

\section{Introduction}
Numerical long-time simulation of full-order models (FOMs) of complex dynamical systems is computationally costly. This is particularly true for physical systems that contain a wide range of spatial and temporal scales, including direct numerical simulation (DNS) \cite{moin1998direct} or large eddy simulation (LES) in fluid mechanics \cite{germano1991dynamic,you2007dynamic,craster2009dynamics} and chaotic systems \cite{10.1143/PTPS.64.346,10.2307/2100687,SIVASHINSKY19771177}. Reduced-order models (ROMs) have been utilized as alternative scientific simulation tools, which are computationally much more efficient than FOMs and retain comparable accuracy for simulating complex dynamical systems. ROMs have played crucial roles in designing, optimizing, and controlling dynamical systems \cite{gugercin_survey_2004,antoulas_approximation_2005,antoulas_interpolatory_2010,benner_survey_2015}.

Several data-driven numerical algorithms have been proposed for reduced-order modeling, including dynamic mode decomposition (DMD) \cite{schmid_2010} and proper orthogonal decomposition (POD) \cite{benner2015survey}. These models leverage some FOM simulation data to construct low-dimensional simplified models that describe the underlying dynamics, with the goal of using these simplified models in generalization regimes to predict the unseen dynamics. Classical projection-based reduced-order modeling techniques (of which DMD and POD are examples) are among the most popular approaches for constructing ROMs of dynamical systems. This approach transforms the simulation results of FOM into a suitable low-dimensional subspace that preserves the largest variance of the training data. In, e.g., POD, classical numerical algorithms (such as Galerkin methods), are subsequently used to rewrite the state variable in the governing equation of the underlying dynamics into a system of ODEs, resulting in a substantially reduced degree of freedom for describing the complex dynamics. Both DMD and POD have been widely used in scientific simulations, particularly for fluid simulations.

ROMs generated from projection-based approaches can preserve crucial physical structures of the dynamics system. However, inappropriate truncation of the POD modes in governing equations can severely degrade modeling accuracy and result in unexpected, unphysical predictive results. Moreover, the precise strategy for mode truncation is task-dependent and is typically limited to explicit and closed definitions of the mathematical models \cite{SAN2019271}. Another drawback of direct projection-based approaches is that they require knowledge of governing equations that model the dynamical system, and this information is often absent for real-world problems. As such, data-driven reduced-order modeling has drawn significant recent attention. For instance, the learning of closure models to compensate for information loss due to mode truncation \cite{SAN2018681,san2017neural,mou2020data,https://doi.org/10.1002/fld.4684}, and data-driven reduced basis representations have been learned from simulation data that provides significantly improved predictive performance of the dynamics compared to classical models \cite{murata_fukami_fukagata_2020,lui2019construction,7572934}. More recently, ``vanilla'' versions of machine learning approaches such as neural ODEs (NODEs) and recurrent neural networks (RNNs) have been used to learn temporal coefficients of the POD of a given complex dynamical system \cite{rom_node,kani2017dr,kani2019reduced}. A well-known issue of the vanilla NODEs and RNNs is that they lack the capability of capturing long-range dependencies from data, making these machine learning models fail to learn the intrinsic physics of complex systems \cite{bengio1994learning,HBNODE:2021}.

\subsection{Our contribution}
We employ the recently developed \emph{heavy-ball neural ODE} (HBNODE) \cite{HBNODE:2021}, an extension of NODE \cite{chen2018neural}, to learn the temporal coefficients of the POD of complex physical systems with a focus on time-dependent simulations in scientific computing. In particular, our examples include the von K\'{a}rm\'{a}n Street (VKS) flow, the Kurganov-Petrova-Popov (KPP) equation, and the one-dimensional Euler equations for fluids modeling. We provide numerical validation on the above three benchmark tasks and a detailed empirical and analysis of why HBNODEs are beneficial for learning the dynamics of POD modes. Our numerical results show the adjoint state of HBNODEs does not vanish, confirming that HBNODEs do learn long-range dependencies, which results in remarkable performance gain over the baseline NODEs. Moreover, our experimental results show significant computational advantages in training and testing HBNODEs over the baseline ROM models.

\subsection{Related work}
There is a healthy amount of recent work on learning POD mode dynamics using deep neural networks, particularly RNNs and vanilla NODEs. Perhaps the most related papers to this article are \cite{rom_node,dutta2021neural,dutta2021data}, which study the NODE framework for learning ROMs. In \cite{rom_node}, the authors developed a POD-NODE ROM framework for learning POD coefficients, which starts from FOM snapshots and then uses an autoencoder to encode the POD representations of FOM snapshots, followed by NODE training and forecasting. The POD-NODE ROM framework achieves appealing results for learning reduced dynamics of the VKS model, and it significantly outperforms the direct application of a long short-term memory (LSTM) network for sequential learning. In \cite{dutta2021neural,dutta2021data}, the authors study the effectiveness of NODEs for reduced-order modeling and predicting environment hydrodynamics. On the one hand, they find that NODEs provide an elegant framework for the stable and accurate evolution of latent-space dynamics with promising generalizability. On the other hand, they noticed that in order to facilitate the widespread adoption of NODEs for large-scale systems, significant effort needs to be directed at accelerating training time. This limitation motivates this article's study and utilization of HBNODEs \cite{HBNODE:2021}, which is the machine-learning backbone of the reduced-order modeling pipeline proposed in this work. There are three major advantages of learning PODs using HBNODEs over the existing deep learning approaches:
\begin{itemize}
\item HBNODEs are a class of continuous-depth neural networks, and they are suitable for learning irregularly-sampled simulation data or physical observations. Hence, observation protocols that entail missing or sparse data are easily tackled in this framework.
\item Certain spectral properties of HBNODEs enable them to capture long-range dependencies from sequential data, which is crucial for learning PODs of complex dynamics. 
\item Both HBNODEs and their adjoint ODEs are computationally much more efficient than baseline NODEs.
\end{itemize}

In addition to the NODE paradigm of continuous-depth neural networks for reduced-order modeling, the RNN --- a natural sequential deep learning model --- has also been successfully used for learning-assisted model reduction. Many advanced RNN algorithms can also be leveraged to enhance learning ROMs, e.g., LSTM networks \cite{LSTM}. RNN-based ROMs have achieved remarkable success in many applied domains, including multiphase flow simulation \cite{kani2017dr,kani2019reduced}, learning advection-dominated systems \cite{maulik2021reduced}, learning chaotic dynamics \cite{ma2018model}, and learning nonlinear aeroelastic models \cite{mannarino2014nonlinear}. Compared to NODEs for learning ROMs, RNNs cannot learn irregularly-sampled time series effectively and can even depart from the underpinning physics due to their discrete nature.

% https://arxiv.org/pdf/2002.00470.pdf --- POD galerkin

\subsection{Organization}
We organize the paper as follows: In Sections~\ref{sec:POD} and~\ref{sec:NODE}, we briefly review the POD-based reduced-order modeling and HBNODE for continuous-depth deep learning, respectively. We present the benchmark physical models of the complex dynamical systems and full-order modeling for data generation in Section~\ref{sec:data}. Section~\ref{sec:model} shows the detailed deep learning model and pipeline for learning POD-based ROMs. We verify the efficacy of our proposed machine learning models and contrast them with several baseline models in Section~\ref{sec:results}, followed by concluding remarks.

\section{POD-based Reduced-order Modeling}\label{sec:POD}
In this section, we briefly review key ideas and procedures of POD-based reduced-order modeling. 

\subsection{Notation}
We denote vectors and matrices by lower- and upper-case boldface letters, respectively. For a vector $\vx = (x_1, \cdots, x_d)^\top\in \mathbb{R}^d$, where $(x_1,\cdots,x_d)^\top$ denotes the transpose of the row vector $(x_1,\cdots,x_d)$, we use $\|\vx\| = {(\sum_{i=1}^d x_i^2)^{1/2}}$ to denote its $\ell_2$ norm, and use $\mathbf{0}$ to denote the zero vector. In cases when $d=2$ and $\bs{x}$ is a spatial vector, we will write the components instead as $\bs{x} = (x,y)^\top$. For a matrix $\mA$, we use $\mA^\top$,  $\mA^{-1}$, and $\|\mA\|$ to denote its transpose, inverse, and spectral norm, respectively. We use $\mI$ to denote the identity matrix, whose size will be clear based on context.

We will consider the approximation of a space-time function $\vu = \vu(\vx,t)$ where $\vx$ is a spatial vector (typically of 1 or 2 dimensions) and $t$ is a scalar on $[0,T]$ for some fixed and finite terminal time $T$. The function $\vu$ may be vector-valued. In much of our discussion, we will take the concrete example of $\vu$ being a solution to a discretized VKS problem, whose details are given in Section \ref{ssec:vks}. For the VKS problem, $\vu\in\RR^2$ contains the horizontal ($x)$ and vertical ($y)$ components of a fluid velocity field. We will write $\vu = (u_x, u_y)$ to denote these two components.

\subsection{POD snapshots}
 \begin{figure}[!ht]
     \centering
     \includegraphics[width=\textwidth]{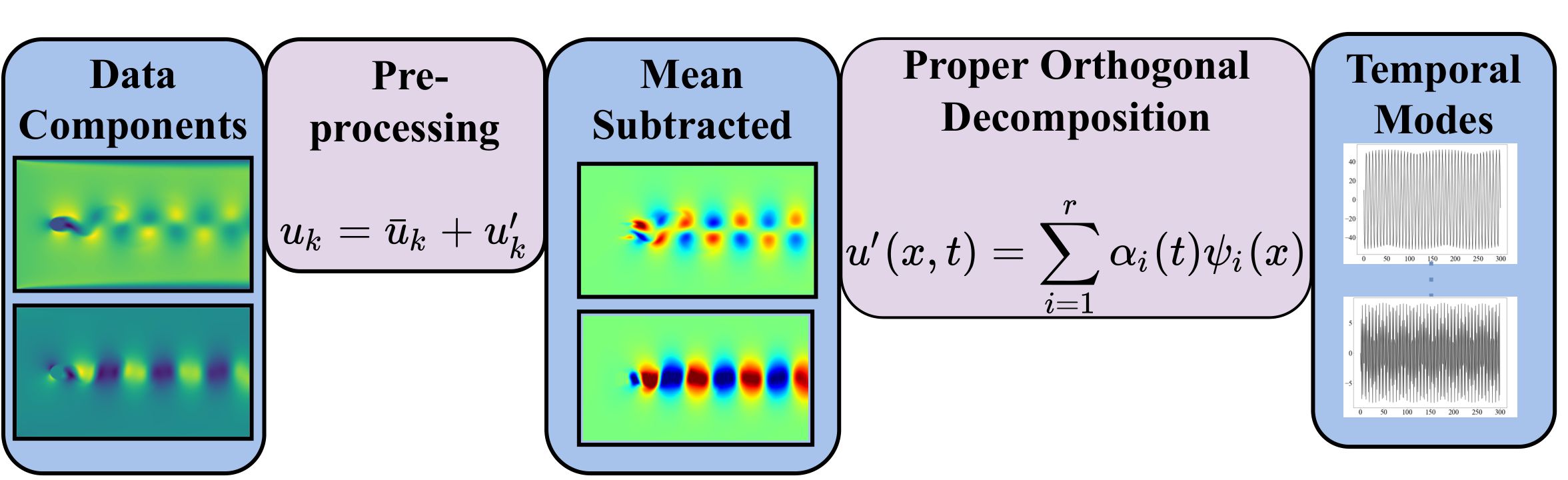}
     \caption{POD pipeline: We first pre-process the data, from experimental observation or FOM simulation, by subtracting the mean. Then we apply spectral decomposition of the covariance matrix and only keep the first $r$ eigenmodes, resulting in the reduced representation --- $\sum_{i=1}^r\alpha_i(t)\psi_i(\vx)$ --- of $\vu'(\vx,t)$, where $\alpha(t)$s are the temporal coefficients and $\psi(x)$s are the eigenmodes.}
     \label{fig:pod_process}
\end{figure}
POD shares a similar spirit and implementation as the celebrated principal component analysis (PCA), the latter of which has been a very popular tool for data analysis \cite{pearson1901liii,LIANG2002527}. The key idea of PCA is to project high-dimensional data into a lower-dimensional space that is spanned by the eigenvectors corresponding to the leading eigenvalues of the covariance matrix of the data. PCA preserves the largest variance of training data and thus contains the most important information contained in the originally high-dimensional data. POD has been introduced in accelerating fluids simulation and reducing the complexity of fluid models since the pioneering work of Berkooz et al. \cite{berkooz1993proper}. Once the leading eigenmodes are obtained via analysis of training data, it is possible to reduce the order (the computational complexity and degrees of freedom) of the complex FOMs. The POD-based dimension reduction approach starts with some training samples of physically observed or numerically simulated snapshots of dynamics. These sample snapshots are aggregated into an ensemble matrix $\mY$, where each row contains the state of a dynamical system at a fixed time step. Next, we compute the covariance matrix of the rows of the matrix $\mY$, and the eigenvectors --- sorted according to the corresponding decreasing-ordered eigenvalues --- are used as the new orthogonal basis for representing the ROM. Below we summarize the crucial steps of identifying the low-dimensional representations via the POD approach, which has also been visualized in Fig.~\ref{fig:pod_process}.
\begin{itemize}
  \item \emph{Data generation.} We simulate the FOM, which is computationally expensive, for a short time to obtain the training data at time steps $t_1,t_2,\cdots,t_{N_t}$. For the VKS problem, when our solution $u$ contains two components $\vu = (u_x,u_y)$ that depend on the two-dimensional spatial variable $\bs{x} = (x,y)$, we assume that FOM snapshots $\bs{u_x}(t_j)$, $\bs{u_y}(t_j)$ are vectorized representations of $N$ spatial degrees of freedom. Then we have
    \begin{align*}
      \bs{u_x}(t_j), \bs{u_y}(t_j) &\in \R^N, & j &=1,2,\ldots,N_t.
    \end{align*}
    For real-world dynamical systems for which we do not know the exact governing equation, we sample the true dynamics via experimental measurements as training data. In either case, we assume that training data is available to us (as the snapshots above), and our goal is to efficiently leverage this data to learn reduced-order dynamics without recourse to the FOM, which we assume is unknown.

\item \emph{Linearly center the data dynamics.} With our VKS-centric notation above, 
    according to the Reynolds decomposition of the flow, we have for fixed $t_j$,
\begin{equation}\label{eq:fluctuating}
  \bs{u_x}=\bar{\bs{u}}_x + \bs{u_x'};\quad \bs{u_y}=\bar{\bs{u}}_y + \bs{u_y'},
\end{equation}
    where $\bar{\bs{u}}_x$ and $\bar{\bs{u}}_y$ are the temporal mean of the solutions, computed over our $N_t$ snapshots. The components $\bs{u_x'}$ and $\bs{u_y'}$ are the \textit{fluctuating components} of the data.

\item \emph{Data assembling.} Concatenate the simulated and centered FOM snapshots into the following matrix $\mY$,
\begin{equation}\label{eq:data-matrix}
  \mY = \left(\begin{array}{ccccccc} 
    \horzbar & (\bs{u}'_x(t_1))^\top     & \horzbar & & \horzbar & (\bs{u}'_y(t_1))^\top     & \horzbar \\
    \horzbar & (\bs{u}'_x(t_2))^\top     & \horzbar & & \horzbar & (\bs{u}'_y(t_2))^\top     & \horzbar \\
             & \vdots                    &          & &          & \vdots                    &          \\
    \horzbar & (\bs{u}'_x(t_{N_t}))^\top & \horzbar & & \horzbar & (\bs{u}'_y(t_{N_t}))^\top & \horzbar 
  \end{array}\right),
\end{equation}
so that row $j$ contains the concatenated snapshot $\bs{u}'_x(t_j), \bs{u}'_y(t_j)$, i.e., the two flattened velocity components at time step $j$. The size of the matrix $\mY$ is $N_t \times 2 N$.

\item \emph{Perform a spectral decomposition of the covariance matrix.} We construct the covariance matrix $\mK$ of the rows of $\mY$ and compute its eigendecomposition:
\begin{align}\label{eq:cov-matrix}
  \mK &= \mY\mY^\top, & \mK &= \bs{A} \bs{\Lambda} \bs{A}^T, & \bs{A} &= \left( \bs{\alpha}_1, \ldots, \bs{\alpha}_{N_t} \right),
\end{align}
where $\bs{\alpha}_j$ is the $j$th eigenvector, and the matrix $\bs{\Lambda}$ is diagonal containing entries $\lambda_j$, the associated non-negative eigenvalues of $\bs{K}$. We assume the eigenvalues are listed in non-increasing order, $\lambda_j \geq \lambda_{j+1}$.

\item \emph{Identify reduced-order modes and truncate.} Larger eigenvalues of $\mK$ are directly related to the dominant characteristics of the dynamical system, while small eigenvalues correspond to small perturbations of the dynamical behavior. The matrix $\mK$ has $N_t$ eigenvalues, and we choose the \textit{order} of the reduced-order model to be $r\ll N_t$ by inspecting a relative information content $I(r)$, defined as follows
\begin{equation}\label{eq:I-def}
  I(r) = \frac{\sum_{i=1}^r\lambda_i}{\sum_{i=1}^{N_t}\lambda_i},
\end{equation}
so that $1 - I(r)$ is a relative Frobenius norm error between $\mK$ and its rank-$r$ spectral approximation. As we will see in Section~\ref{sec:results}, $I(r)$ is close to one for practical problems, even for very small $r$. As output of the procedure, we can construct the following (discretized) ROM of the fluctuating component of the dynamics
\begin{equation}\label{eq:ROM-fluc}
  \bs{u}'_*(t_j) \approx \sum_{i=1}^r (\alpha_i)_j \bs{\psi}_{*,i},
\end{equation}
where $* \in \{x, y\}$, and $(\bs{\psi}_{x,i}^{\top}, \bs{\psi}_{y,i}^{\top})^\top \in \R^{2 N}$ is a vector denoting a discretized spatial function; the entries of $\bs{\psi}_{*,i}$ correspond to the $N$ degrees of freedom in the snapshots $\bs{u}_\ast$, and is a subvector of the $i$-th right-singular vector of $\mY$. Equivalently, it is defined as,
$$
    \bs{\psi}_{*,i} = \frac{1}{\lambda_i} \bs{Y}^\top \bs{\alpha}_i = \frac{1}{\lambda_i} \sum_{j=1}^{N_t} (\alpha_i)_j \bs{u}'_{\ast}(t_j).
$$
\end{itemize}

A ``standard'' POD approach would next project the (assumed known) dynamical model onto $\mathrm{span}\{\bs{\psi}_{*,i}\}_{i=1}^r$. We will proceed to assume that such a dynamical model is unknown to us, and will instead use machine learning models to predict dynamics.

\begin{remark}
With the training data $\alpha_i$ available at time steps $t_1,t_2,\cdots,t_{N_t}$ through the above procedure, extrapolation of the FOM dynamics or experimental measurements amounts to predicting the POD coefficients $\alpha_i(t)$ for future time $t$ accurately, in our case using machine learning models. Notice that $\alpha_i(t)$ is observed sequentially and has a continuous profile, indicating the potential advantages of using NODE for learning $\alpha_i(t)$, as we describe next.
\end{remark}

\section{Heavy-ball Neural ODEs}\label{sec:NODE}

In this section, we briefly review NODE and HBNODE and algorithms for their training and testing. Moreover, we provide some simple analysis of why HBNODE is better for learning POD coefficients for reduced-order modeling leveraging insights from the acceleration theory of the classical momentum methods.

\subsection{Neural ODEs}
NODEs \cite{chen2018neural} are a class of continuous-depth (-time) neural networks \cite{rosenblatt1961principles,cohen1983absolute}. The continuous-time nature of NODEs makes them particularly suitable for learning complex dynamics from irregularly-sampled sequential data, see, e.g., \cite{chen2018neural,latentODE,NEURIPS2019_21be9a4b,massaroli2020dissecting,norcliffe2020_sonode}. Mathematically, a NODE is formulated as the following first-order ODE:
\begin{equation}\label{eq:NODE}
\frac{d{\vh}(t)}{dt}=\vf({\vh}(t),t,\theta),
\end{equation}
where $\vf({\vh}(t),t,\theta) \in \RR^d$ is specified by a neural network parameterized by $\theta$, e.g., a two-layer feed-forward neural network. Starting from the input ${\vh}(0)$, NODEs learn the representation and perform prediction by solving \eqref{eq:NODE} from $t=0$ to $T$ using a numerical integrator with a given error tolerance, often with adaptive step size solver or adaptive solver for short \cite{DORMAND198019}. Solving \eqref{eq:NODE} from $t=0$ to $T$ in a single pass with an adaptive solver requires evaluating $\vf({\vh}(t),t,\theta)$ at various timestamps, with computational complexity measured by the number of function evaluations in a time-forward sweep (``forward NFEs'') \cite{chen2018neural}. 

The \emph{adjoint sensitivity method}, or the adjoint method \cite{adjoint}, is a memory-efficient method for training NODEs through optimization of $\theta$. We regard the output ${\vh}(T)$ as the prediction and denote the loss between the prediction ${\vh}(T)$ and the ground truth as $\mathcal{L}$. Let ${\va}(t):={\partial \mathcal{L}}/{\partial {\vh}(t)}$ be the adjoint state, then we have (see \cite{chen2018neural,adjoint} for details)
\begin{equation}\label{eq:NODE:gradient}
\frac{d\mathcal{L}}{d\theta} = \int_0^T{\va}(t)^\top \frac{\partial \vf({\vh}(t),t,\theta)}{\partial\theta} dt,
\end{equation}
with ${\va}(t)$ satisfying the following adjoint ODE
\begin{equation}\label{eq:NODE:adjoint}
\frac{d{\va}(t)}{dt} = -{\va}(t)^\top \frac{\partial}{\partial {\vh}} \vf({\vh}(t),t,\theta),
\end{equation}
which is solved numerically from $t=T$ to $0$ and also requires the evaluation of the right-hand side of \eqref{eq:NODE:adjoint} at various timestamps, with the number of NFEs during this time-backward sweep (``backward NFEs'') measuring the computational complexity.

There are several critical problems with NODEs, including (i) Given an error tolerance, the NFEs required in a single forward pass can be excessive. Moreover, solving the adjoint ODE \eqref{eq:NODE:adjoint} often requires more NFEs than solving the forward ODE \eqref{eq:NODE}. (ii) In training NODEs, the adjoint state ${\va}(t)$ often vanishes, i.e., the norm of ${\va}(t)$ tends to $0$, impeding NODEs from learning long-range dependencies \cite{lechner2020learning}, resulting in poor predictive performance.

\subsection{Heavy-ball neural ODEs}\label{ssec:hbnode}
The authors of \cite{HBNODE:2021,wang2021does} proposed HBNODEs and their generalized version, named generalized HBNODEs (GHBNODEs). HBNODEs are motivated by ideas from momentum-accelerated gradient descent \cite{polyak1964some} and they can be regarded as the continuous limit of the MomentumRNN model \cite{MomentumRNN}. Mathematically, the HBNODE is a special second-order neural ODE of the following form
\begin{equation}\label{eq:HBNODE-1}
\frac{d^2\vh(t)}{dt^2} + \gamma\frac{d\vh(t)}{dt} = \vf\big(\vh(t),t,\theta\big),
\end{equation}
where $\gamma\geq 0$ is the damping parameter, which can be set as a tunable or a learnable hyperparameter with positivity constraint. In the trainable case, we adopt the one used in \cite{HBNODE:2021}, that is $\gamma = \epsilon \cdot \text{sigmoid}(\omega)$ for a trainable $\omega\in \mathbb{R}$ and a fixed tunable upper bound $\epsilon$, e.g., $\epsilon=1$. The HBNODE \eqref{eq:HBNODE-1} can be rewritten as the following system of first-order NODEs
\begin{equation}\label{eq:HBNODE-1st}
\frac{d\vh(t)}{dt} = \vm(t); 
\quad \frac{d\vm(t)}{dt} = -\gamma \vm(t) + \vf(\vh(t),t,\theta).
\end{equation}

\subsubsection{Computational advantages of HBNODE vs. NODE}\label{subsubsec-computational-advantages}
To show why HBNODE enjoys computational efficiency in training and testing, let us first consider the adjoint equation of \eqref{eq:HBNODE-1}, which will again be solved using adaptive numerical ODE solvers. First, the following theoretical result \cite{HBNODE:2021} shows that the adjoint of an HBNODE is also an HBNODE.

\begin{proposition}[Adjoint equation for HBNODE \cite{HBNODE:2021}]\label{prop:adjoint-HBNODE}
The adjoint state $\va(t):=\partial\mathcal{L}/\partial\vh(t)$ for the HBNODE \eqref{eq:HBNODE-1} satisfies the following HBNODE with the same damping parameter $\gamma$ as that in \eqref{eq:HBNODE-1},
\begin{equation}\label{eq:adjoint-HBNODE-2nd}
\frac{d^2\va(t)}{dt^2} - \gamma\frac{d\va(t)}{dt} = \va(t) \frac{\partial \vf}{\partial\vh}(\vh(t),t,\theta). 
\end{equation}
Notice that we solve the adjoint equation \eqref{eq:adjoint-HBNODE-2nd} from $t=T$ to $0$ via backward propagation. By letting $\tau=T-t$ and $\vb(\tau)=\va(T-\tau)$, we can rewrite \eqref{eq:adjoint-HBNODE-2nd} as follows,
\begin{equation}\label{eq:adjoint-HBNODE-2nd-b}
\frac{d^2\vb(\tau)}{d\tau^2} + \gamma \frac{d\vb(\tau)}{d\tau} = \vb(\tau)\frac{\partial \vf}{\partial\vh}(\vh(T-\tau),T-\tau,\theta).
\end{equation}
Therefore, the adjoint of the HBNODE is also an HBNODE and they have the same damping parameter.
\end{proposition}
The above result indicates that the adjoint problem for HBNODE is of the same type as the forward problem, accelerating backward propagation provided the forward propagation is accelerated.

Next, we provide theoretical insights into the computational efficiency of HBNODEs. Leveraging the acceleration theory of the heavy-ball momentum method in taming the oscillation of the optimization trajectory by reducing the condition number of the underlying problem. %\cite{goh2017momentum,recht2010cs726}. 
Furthermore, it is worth mentioning that the heavy-ball method only provably accelerates strongly convex quadratic optimization problems, while in practice, it accelerates general optimization problems. In analogy to the acceleration theory of the heavy-ball method, we consider the following two linearized high-dimensional ODE systems
\begin{equation}\label{eq:NODE-linear}
\frac{d\vh(t)}{dt} = \mA\vh(t),
\end{equation}
and
\begin{equation}\label{eq:HBNODE-linear}
\begin{aligned}
\frac{d\vh(t)}{dt} &= \vm(t)\\
\frac{d\vm(t)}{dt} &= -\gamma\vm(t) + \mA\vh(t)
\end{aligned} \Leftrightarrow \frac{d}{dt}\begin{pmatrix}
\vh(t) \\ \vm(t)
\end{pmatrix} = \underbrace{\begin{pmatrix}
{\bf 0} &\ \  \mI\\
\mA &\ \  (-\gamma\mI)
\end{pmatrix}}_{:=\mB}\begin{pmatrix}
\vh(t)\\ \vm(t)
\end{pmatrix},
\end{equation}
where we assume $\mA$ is positive definite to simplify our analysis and reveal intuition of the advantages of HBNODE over NODE. Let the eigenvalues and eigenvectors of $\mA$ be given by $\lambda_i,\vv_i$ respectively. Following the proof of the acceleration of heavy-ball momentum\footnote{see \url{http://www.math.utah.edu/~bwang/mathds/Lecture8.pdf} for details.}, we can show that ${\vert\tilde{\lambda}_{\max{}}\vert}/{\vert \tilde{\lambda}_{\min{}}\vert}\leq \sqrt{{\vert \lambda_{\max{}}\vert}/{\vert \lambda_{\min{}}\vert}}$, where $\lambda_{\max{}}$ and $\lambda_{\min{}}$ are the largest and smallest eigenvalues. Similarly, $\tilde{\lambda}_{\max}$ and $\tilde{\lambda}_{\min}$ are the largest and smallest eigenvalues, in magnitude, of $\mB$. 

Notice that the ratio ${\vert \lambda_{\max{}}\vert}/{\vert \lambda_{\min{}}\vert}$ and ${\vert\tilde{\lambda}_{\max{}}\vert}/{\vert \tilde{\lambda}_{\min{}}\vert}$ are the stiffness of the linear ODE model \eqref{eq:NODE-linear} and the corresponding linear HBNODE counterpart \eqref{eq:HBNODE-linear}. Thus the heavy-ball NODE can be much less stiff than the original NODE. If the stiffness of the ODE model is $\kappa$, using the heavy-ball model results in stiffness of at most $\sqrt{\kappa}$, which is a substantial reduction.

Recall that we use the adaptive step size explicit solver to solve both forward and backward ODEs, from $t=0$ to $T$, in training NODEs and HBNODEs. A less stiff model allows the adaptive solver to use a much large step size and thus can significantly reduce NFEs. Moreover, Proposition \ref{prop:adjoint-HBNODE} indicates that the adjoint equation of an HBNODE is also an HBNODE, and therefore we reap the computational advantages of relaxed stiffness in both forward and backward propagation phases. Our previous analysis only considers very simple linear ODE models. How to extend the analysis to the neural network is a very interesting future direction. One particular idea is analyzing the NODE and HBNODE when they are overparameterized, in which case one could leverage neural tangent kernel theory \cite{jacot2018neural}.

\subsubsection{Generalized HBNODEs (GHBNODEs)}
Compared to vanilla NODEs, high-order NODEs include HBNODEs usually suffer from the uncontrolled aggregation of the hidden state, deteriorating model performance at best, and blowing up training at worst. To alleviate this issue, in \cite{HBNODE:2021} the authors propose the following generalized HBNODE
\begin{equation}\label{eq:GHBNODE}
\begin{aligned}
\frac{d\vh(t)}{d t} = \sigma(\vm(t));\quad
\frac{d\vm(t)}{d t} = -\gamma\vm(t) + \vf(\vh(t),t,\theta) - \xi\vh(t), 
\end{aligned}
\end{equation}
where $\sigma(\cdot)$ is a nonlinear activation, which is set as $\tanh$ by default. The positive hyperparameters $\gamma,\xi>0$ are two tunable or learnable hyperparameters. In the trainable case, we let $\gamma = \epsilon \cdot \text{sigmoid}(\omega)$ as in HBNODE, and $\xi = \text{softplus}(\chi)$ to ensure that $\gamma,\xi \geq 0$. Compared to HBNODEs, GHBNODEs integrate two ideas to improve the neural network architecture design: (i) Incorporating the gating mechanism $\sigma$ used in LSTM \cite{hochreiter1997long} and GRU \cite{cho2014learning}, which can suppress the aggregation of $\vm(t)$; (ii) Following the idea of skip connections \cite{he2016identity}, HBNODEs add the term $\xi\vh(t)$ into the governing equation of $\vm(t)$, which benefits training and generalization of GHBNODEs. It has been extensively verified that GHBNODE can indeed control the growth of $\vh(t)$ effectively, which significantly improve the performance of machine learning models on various sequential learning tasks.

Another interesting result is that though the adjoint state of the GHBNODE does not satisfy the exact heavy-ball ODE, it also significantly reduces the backward NFEs in practice. We observe that sometimes GHBNODEs are computationally more efficient than HBNODEs.

\subsubsection{(G)HBNODEs learn long-range dependencies effectively}\label{subsubsec-long-term-dependencies}
Learning long-range dependencies is crucial for the success of deep learning for sequential data, and vanishing and exploding gradients are two bottlenecks for training RNNs to learn long-range dependencies \cite{bengio1994learning,pascanu2013difficulty}. The exploding gradients issue can be effectively resolved via gradient clipping, training loss regularization, etc \cite{pascanu2013difficulty}. The vanishing gradient phenomenon in training RNNs materializes in continuous-depth neural networks as vanishing of the adjoint state \cite{HBNODE:2021}. In particular, we consider $\va(t):=\partial\mathcal{L}/\partial\vh(t)$, and when the vanishing gradient phenomenon occurs, $\va(t)$ goes to ${\bm 0}$ quickly as $T-t$ increases, so that $d\mathcal{L}/d\theta$ in \eqref{eq:NODE:gradient} will be essentially independent of $\va(t)$ for larger $T-t$. We have the following expressions for the adjoint states of the NODE and HBNODE (see \cite{HBNODE:2021} for details):
\begin{itemize}[leftmargin=*]
\item For NODE, we have 
\begin{equation}\label{eq:NODE-gradient}
\frac{\partial\mathcal{L}}{\partial\vh_t} = \frac{\partial\mathcal{L}}{\partial\vh_T}\frac{\partial\vh_T}{\partial\vh_t} = \frac{\partial\mathcal{L}}{\partial\vh_T}\exp\Bigg\{-\int_T^t\frac{\partial \vf}{\partial \vh}(\vh(s),s,\theta)ds\Bigg\}.
\end{equation}

\item For GHBNODE\footnote{HBNODE can be seen as a special GHBNODE with $\xi=0$ and $\sigma$ be the identity map.}, we have
\begin{equation}\label{eq:HBNODE-gradient}
\begin{aligned}
\begin{bmatrix}
\frac{\partial\mathcal{L}}{\partial\vh_t}&\hspace{-0.1in} \frac{\partial\mathcal{L}}{\partial\vm_t}
\end{bmatrix} &= 
\begin{bmatrix}
\frac{\partial\mathcal{L}}{\partial\vh_T}&\hspace{-0.1in}  \frac{\partial\mathcal{L}}{\partial\vm_T} 
\end{bmatrix}
\begin{bmatrix}
\frac{\partial\vh_T}{\partial\vh_t} &\hspace{-0.1in} \frac{\partial\vh_T}{\partial\vm_t}\\
\frac{\partial\vm_T}{\partial\vh_t} &\hspace{-0.1in}
\frac{\partial\vm_T}{\partial\vm_t}\\
\end{bmatrix}\\
&=
\begin{bmatrix}
\frac{\partial\mathcal{L}}{\partial\vh_T} \  \frac{\partial\mathcal{L}}{\partial\vm_T} \end{bmatrix}\exp\Bigg\{-\underbrace{\int_T^t\begin{bmatrix}
{\bf 0} &\hspace{-0.05in} \frac{\partial \sigma}{\partial \vm}\\
\big(\frac{\partial \vf}{\partial\vh}-\xi\mI\big) &\hspace{-0.05in}
-\gamma\mI
\end{bmatrix}ds}_{:=\mM} \Bigg\}.
\end{aligned}
\end{equation}
\end{itemize}
For the matrix $\mM$, we have the following useful property about its spectrum.
\begin{proposition}[\cite{HBNODE:2021}]\label{lemma-eigan-M}
The eigenvalues of $-\mM$ can be paired so that the sum of each pair equals 
$(t-T)\gamma$. 
\end{proposition} 
Following the argument in \cite{HBNODE:2021}, Proposition~\ref{lemma-eigan-M} can be used to show that the adjoint state of NODE in \eqref{eq:NODE-gradient} may vanish when $T-t$ is large, but the adjoint state of (G)HBNODEs in \eqref{eq:HBNODE-gradient} will not vanish. This property supports the claim that HBNODEs benefit in learning long-range dependencies, which in turn further boosts the accuracy in learning POD of complex dynamical systems.

In Section~\ref{sec:results}, we will validate the above theoretical merits of HBNODEs over NODEs using the benchmark problems listed in Section~\ref{sec:data} below.

\section{Benchmarks and Data Preparation}\label{sec:data}
In this section, we will present some details of the three benchmark physical models --- VKS, KPP, and Euler equations --- used for validating the efficacy of learning POD with HBNODEs. Our training data for the KPP and Euler equations are generated by solving FOMs; the training data for the VKS dataset is adopted from a publicly available dataset.

\subsection{VKS model}\label{ssec:vks}
The von K\'{a}rm\'{a}n vortex street (VKS) is a fluid dynamics phenomenon where vortices appear in a periodic fashion in the wake of flow past a blunt object, frequently a cylinder. A very small Reynolds number results in a laminar smooth flow past the cylinder, and very large Reynolds numbers result in a turbulent flow. In an appropriate middle regime, the VKS phenomenon appears and can be simulated. The associated dynamical model is the two-dimensional Navier-Stokes equations; with ${\vu} = (u_x, u_y)$ the fluid velocity, these equations read,
\begin{align*}
  \pfpx{\bs{u}}{t} = \nu \nabla^2 \bs{u} - (\bs{u} \cdot \nabla) \bs{u} -\frac{1}{\rho_0} \nabla p,
\end{align*}
where $\rho_0$ is the spatially uniform pressure, and $\nu$ is the kinematic viscosity, which is inversely related to the Reynolds number. Our experimental setup concerns flow past a cylinder in two spatial dimensions with conditions that result in steady-state VKS flow after an initial transient period. We follow the experimental setting used in \cite{rom_node} to acquire simulation data.

\subsection{KPP model}\label{ssec:kpp}
The Kurganov-Petrova-Popov (KPP) model is a scalar, two-dimensional conservation law, first proposed in \cite{kurganov_adaptive_2007}. This system is difficult to simulate since it features a non-convex flux, and is given by,
\begin{align*}
  \pfpx{u}{t} + \nabla \cdot \bs{f}(u) = 0,  t > 0,\qquad x \in [-2, 2], \hskip 5pt y \in \left[-\frac{5}{2}, \frac{3}{2}\right],
\end{align*}
where $\bs{f}(u) = \left( \sin u, \cos u \right)^T$ and $\nabla := \left( \ppx{x}, \ppx{y} \right)$. Our setup mirrors that in \cite{kurganov_adaptive_2007}, so that we use the following initial data
\begin{align*}
  u(x,y,0) &= \left\{ \begin{array}{rl} \frac{ 14 \pi}{4}, & x^2 + y^2 < 1, \\ \frac{\pi}{4}, & \mathrm{else} \end{array}\right.
\end{align*}
We employ a finite volume scheme utilizing a Lax-Friedrichs flux with a 5th-order WENO reconstruction over the two-dimensional rectangular domain with a Cartesian mesh up to time $T = 10$. The simulation uses a tensorial grid with $N_x = 50$, $N_y = 50$ (corresponding to $N=N_x N_y = 2500$ total spatial degrees of freedom), and $N_t = 1250$.

\subsection{Euler equations for fluids modeling}
\label{ssec:ee}
The one-dimensional Euler equations of gas dynamics are a system of conservation laws. We consider the simulation of a \textit{parameterized} shock-entropy problem from this differential equation, whose setup is given by,
\begin{align*}
  \pfpx{\bs{u}}{t} + \pfpx{\bs{f}(\bs{u})}{x} &= 0,\quad  t > 0, x \in [-5, 5], \\ 
\mbox{with}\  \bs{f}(u) &= \left( \begin{array}{c} \rho u \\ \rho u^2 + p \\ (E+p) u \end{array}\right)^T,
\end{align*}
where $\bs{u}:= (\rho\  \rho u\  E)^\top \in \RR^3$ is the unknown with $(\rho, u, p, E)$ denoting the gas density, velocity, pressure, and energy, respectively. The system is closed via the following relationship between $E$ and $p$:
\begin{align*}
  p = (\gamma - 1) \left( E - \frac{1}{2} \rho u^2 \right),
\end{align*}
where $\gamma$ is the heat capacity ratio, a gas-dependent constant.\footnote{This $\gamma$ is distinct from the $\gamma$ discussed in Section \ref{ssec:hbnode}.} We take boundary conditions at $x = \pm 5$ as those given by the initial data. The shock-entropy problem features smoothly oscillating as well as discontinuous features. 

We again employ a finite volume scheme to solve the Euler equations, using a Harten-Lax-van Leer (HLL) flux, which is an approximated Riemann solver \cite{harten_upstream_1983}. Our simulations integrate up to terminal time $T=1.8$, with a uniform grid having $N=1000$ degrees of freedom in the scalar spatial variable $x$. 

This last example differs from the previous two in that we consider this a \textit{parametric} equation, where $\bs{\eta}=(\eta_u,\eta_\rho)\in\RR^2$ is a parameter for the initial conditions. We initialize the dynamics using the parameter $\bs{\eta}$ as follows, where $\eta_u$ varies on the interval $[2,3]$ and $\eta_\rho$ varies on the interval from $[3,4]$. The parametric initial data $(u(x,0), \rho(x,0), p(x,0)) = (u_0, \rho_0, p_0)$ are given by
\begin{equation*}
    u_0=\begin{cases} \eta_u& x<-4\\ 0&\text{else} \end{cases},\quad
    \rho_0=\begin{cases}\eta_\rho&x<-4\\1+0.2\sin(\pi x)&\text{else}\end{cases},\quad
    p_0=\begin{cases}\frac{31}{3}&x<-4\\1&\text{else}\end{cases}
\end{equation*}
\noindent
We generate training data by gathering an ensemble of trajectories for the above problem over a grid of $\bs{\eta}$ values and attempt to learn dynamics on unseen values of $\bs{\eta}$. Thus, in this example we not only seek to predict to future times, but also trajectories on parameter values not in the training set.

\section{Learning Pipeline}\label{sec:model}
In this section, we describe the detailed pipeline of using deep learning for reduced-order modeling accompanied by the baseline ROMs.

\subsection{Learning-based reduced-order modeling}
\begin{figure}[!ht]
     \centering
     \includegraphics[width=\textwidth]{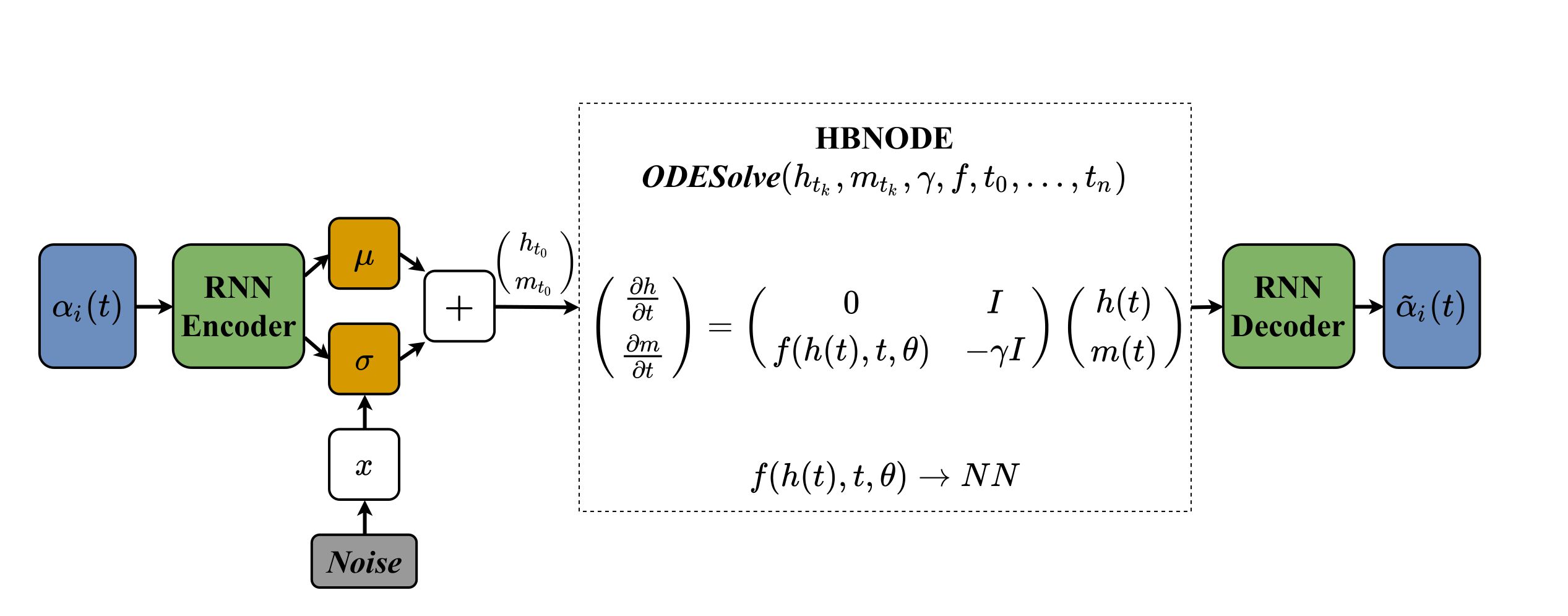}
     \caption{The pipeline of predicting the temporal coefficients for a single step forward using HBNODE leveraging a variational autoencoder. We first use a RNN encoder to encode the input data and then sample the states $\vh$ and $\vm$ and evolve them using an HBNODE. Finally, we apply an RNN decoder to the final representation to get the prediction.
     }
     \label{fig:vae_hbnode}
\end{figure}

\begin{figure}[!ht]
     \centering
     \includegraphics[width=\textwidth]{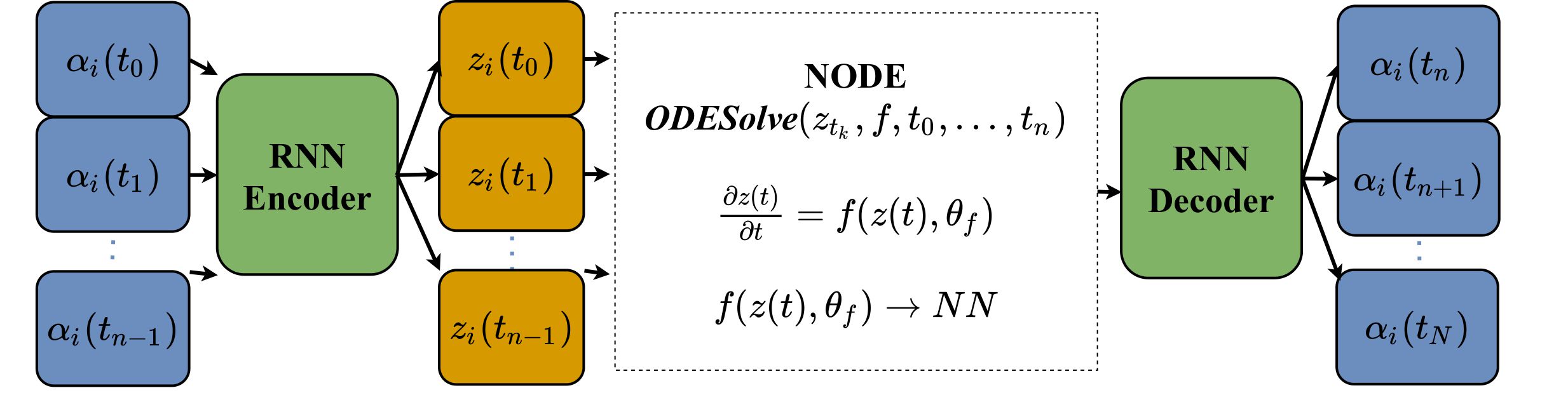}
     \caption{The pipeline of predicting the temporal coefficients for multi-steps ahead. First, we encode the input sequence $\{\alpha_i(t_j)\}_{j=0}^{n-1}$ using a RNN encoder to obtain the latent sequence $\{z_i(t_j)\}_{j=0}^{n-1}$. Second, we use a NODE to learn a ``good'' representation of the input sequence by evolving the latent sequence $\{z_i(t_j)\}_{j=0}^{n-1}$. Third, we apply a RNN decoder to the ``good'' representation to get the final prediction. Notice that NODE can be replaced with (generalized) HBNODE, in which case we need to obtain another sequence of momentum states from the RNN encoder.}
     \label{fig:rnn_node}
 \end{figure}

Our machine learning-based reduced-order modeling framework is flexible for machine learning model selection, e.g., using either HBNODE or NODE as shown in Fig.~\ref{fig:vae_hbnode} and Fig.~\ref{fig:rnn_node}, respectively. In our learning-based reduced-order modeling framework, we first apply POD outlined in Section~\ref{sec:POD} on the training data to extract (discretized) temporal coefficients $\bs{\alpha}(t)$'s and the eigenmodes $\bs{\psi}(\vx)$'s following \eqref{eq:ROM-fluc}. Next, we will use machine learning models to predict future dynamics $u(\vx,t)$ leveraging these coefficients and modes. In particular, the main task is an extrapolation of the temporal coefficients $\bs{\alpha}(t)$'s using NODEs or HBNODEs.

To predict future values of the POD data, we consider two different machine learning architectures, shown in Fig.~\ref{fig:vae_hbnode} and Fig.~\ref{fig:rnn_node}, respectively. The first architecture is a one-to-one architecture that predicts the value at $t_{k+1}$ based on the data at $t_{k}$. The second architecture is a sequence-to-sequence architecture that uses sequence data points to predict the following sequence of data points. The overlap in the sequence prediction can be adjusted so that the predicted sequence is entirely new or that only one new data point is predicted.

The first architecture under our study is adapted from \cite{rom_node}, which was originally used to compare the performance of NODE and LSTM in model reduction. We replace the vanilla NODE used in \cite{rom_node} with the HBNODE, and we depict the modified architecture in Fig.~\ref{fig:vae_hbnode}. Compared to the pipeline used in \cite{rom_node}, after the RNN encoding of the temporal coefficients we have to sample both $\vh$ and $\vm$ to accommodate learning using HBNODE. In contrast, the vanilla NODE used in \cite{rom_node} only needs to sample the state $\vh$. The above encoding and sampling procedure is accomplished via a variational autoencoder \cite{kingma2013auto1}.

We plot the second architecture in Fig.~\ref{fig:rnn_node}, where the vanilla NODE can be replaced with (generalized) HBNODE. For the second architecture, i.e., the sequence-to-sequence architecture, takes a sequence of length $n$ inputs and predicts a sequence of outputs, we encode the input sequence $\{\alpha_i(t_j)\}_{j=0}^{n-1}$ into the latent sequence $\{z_i(t_j)\}_{j=0}^{n-1}$ using an RNN encoder, then we use NODE or HBNODE to evolve the latent sequence to get the desired representation, followed by an RNN decoder to get the final long-term prediction $\{\alpha_i(t_j)\}_{j=n}^N$.

\subsection{A baseline comparison: Dynamic Mode Decomposition (DMD)}
We employ DMD as another baseline model reduction method to demonstrate the effectiveness of learning-based model reduction using HBNODEs. In this part, we briefly review the idea of DMD for reduced-order modeling. 
To compare DMD to the learning-based reduced-order modeling using HBNODE, we consider only modeling the fluctuating components $\vu'$ of the snapshots, see \eqref{eq:fluctuating}. The predictions of DMD are generated by a linear operator $\bs{A}$ corresponding to a linear difference equation $\vu'_{k+1} = \mA \vu'_k$, where $\bs{A}$ must be learned. In DMD, dominant eigenvalues and eigenvectors of $\mA$ are computed via the singular value decomposition (SVD). Although the true underlying dynamics may be nonlinear, the Koopman operator formalism concludes that a \textit{lifted} version of the dynamics is indeed linear. For nonlinear problems, DMD attempts to learn these lifted linear dynamics.

Let $\mU^{(k)}$ be the snapshot matrix for the time interval $t_0,\ldots,t_{\text{train}-1}$ and $\mU^{(k+1)}$ be the snapshot matrix for the time interval $t_1,\ldots,t_\text{train}$, i.e., column $j$ of $\bs{U}^{(k)}$ corresponding to time snapshot $t_{j-1}$. In particular, let $\mU^{(k+1)}\approx \mA\mU^{(k)}$ where $\mU^{(k)}$ is given by the SVD $\mU^{(k)}=\mX\Sigma \mV^*$. We further denote $\tilde{\bs{X}}$, $\tilde{\bs{V}}$, and $\tilde{\bs{\Sigma}}$ as the rank-$r$ truncation of $\bs{X}$, $\bs{V}$, and $\bs{\Sigma}$, respectively. Then we may compute an approximation $\tilde{\mA}$ directly from $\mU^{(k+1)}$ by the following,
\begin{equation}
    \tilde{\mA} = \tilde{\mX}^*\mU^{(k+1)}\tilde{\mV}\tilde{\bm{\Sigma}}^{-1}
    \label{eq:dmd_computation}
\end{equation}

\begin{figure}[H]
 \centering
     \includegraphics[width=\textwidth]{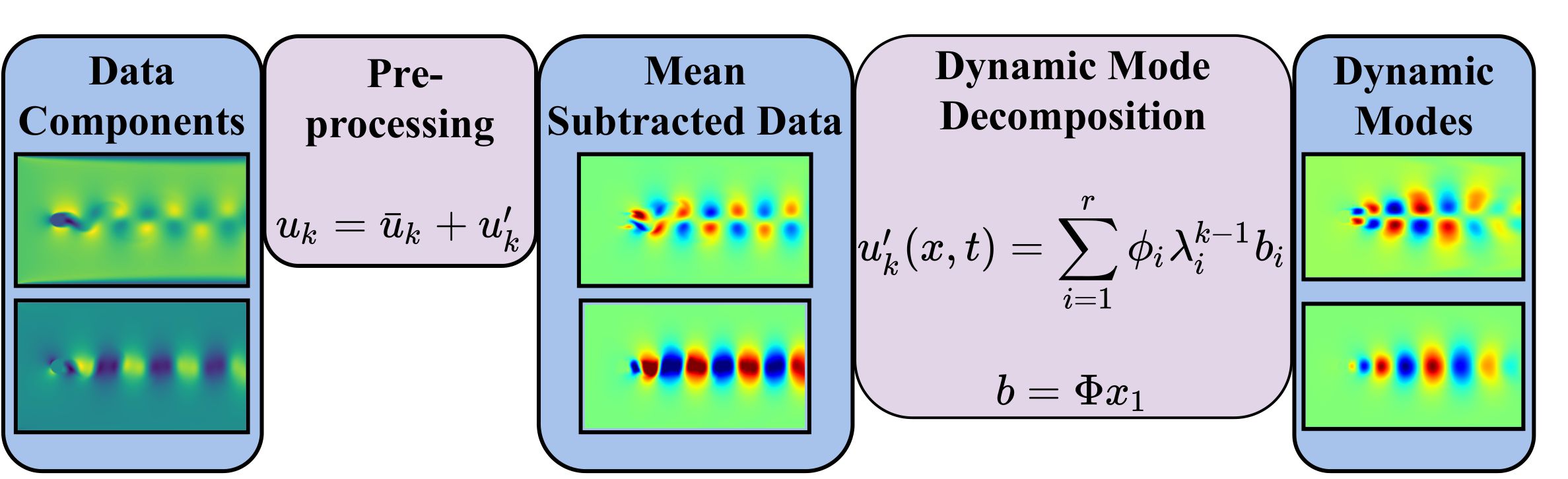}
     \caption{DMD pipeline: The data is pre-processed by subtracting the mean to capture the fluctuating components of the data. In addition, the following lifts $\{\cos(\vx),\sin(\vx),\vx^2,\vx^3\}$ were applied to the data, and then vectorized along the snap-shot axis. We then generate the full-order model according to the spectral decomposition of the linear transformation between the two snapshot matrices at subsequent time intervals. To reduce the order only the dominant $r$ eigenmodes are selected, resulting in a representation $u'(\vx,t)=\sum^r_{i=1}\phi_i\lambda_i^{k-1}b_i$ for the lifted data $u'(\vx,t)$.}
     \label{fig:dmd_process}
 \end{figure}

The reduced matrix $\tilde{\mA}$ is composed of the dominant $r$ eigenvalues $\lambda_1,\ldots,\lambda_r$ and eigenvectors $\bm{\phi}_1,\ldots,\bm{\phi}_r$. These eigenvectors are also known as the DMD modes. Given training data on the training interval $t_0,\ldots, t_{\text{train}}$, the matrix $\tilde{\mA}$ is formulated by partitioning the snapshot matrix $\vu'$ into two time intervals. Validation data on the interval $t_{\text{train}+1},\ldots, t_\text{valid}$ is generated by solving $u'(t_{\text{train}+k}) = \tilde{\mA}^ku'(t_{\text{train}})$. We depict DMD-based reduced-order modeling in Fig.~\ref{fig:dmd_process}. More details of DMD can be found at e.g., \cite{schmid_2010}.

\section{Experimental Results}\label{sec:results}

In each experiment below, we contrast the performance of HBNODE-based ROM to two baseline ROMs, namely, NODE-based and DMD-based ROMs. We observe consistently improved predictive performance of HBNODE over baseline ROMs. We interpret the improved performance using HBNODEs by inspecting the stiffness and adjoint state of HBNODEs, confirming the theoretical results. Animated comparisons of the data reconstructions can be found at \cite{github-animation}.

\begin{figure}[!ht]
\centering
 \begin{tabular}{cc}
 \includegraphics[clip, trim=0.01cm 0.01cm 0.01cm 0.01cm, width=0.42\columnwidth]{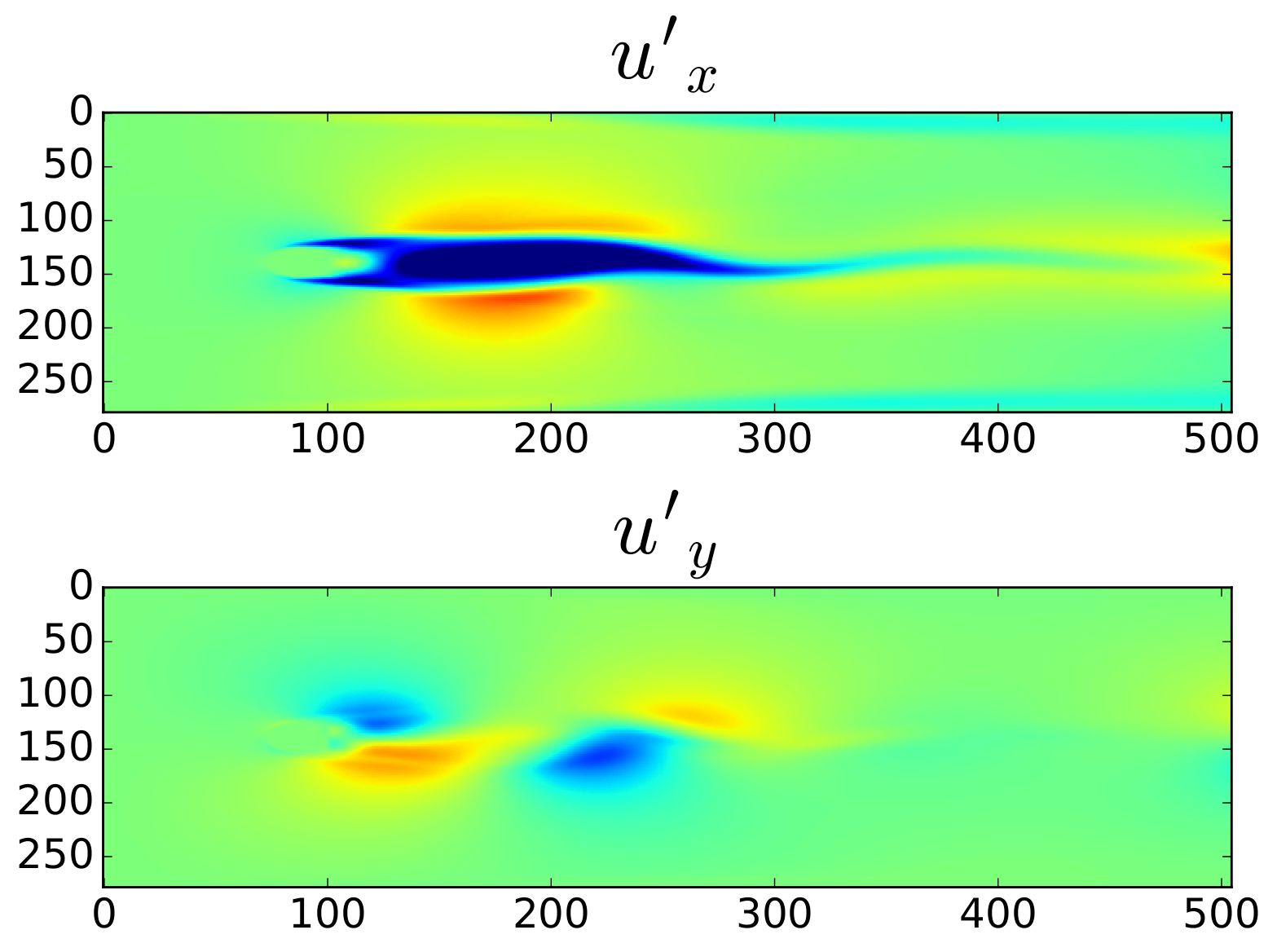}&
 \includegraphics[clip, trim=0.01cm 0.01cm 0.01cm 0.01cm, width=0.42\columnwidth]{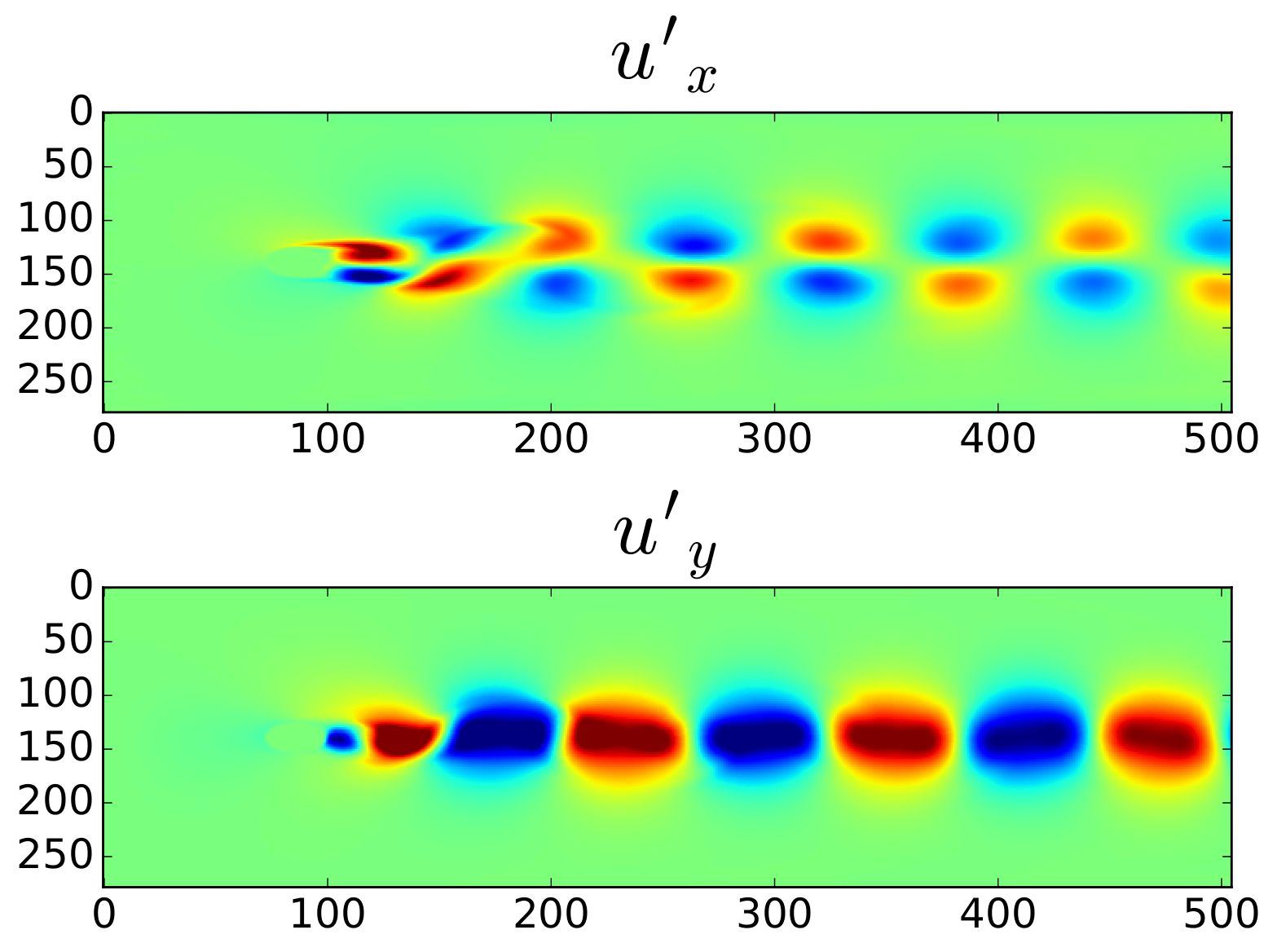}\\
 (a) Transient state $t=50$ &
 (b) Steady-state $t=150$
 \end{tabular}
 \caption{Comparison of transient and steady-state phases of the VKS dataset. The steady-state phase contains quasi-periodic solutions conducive to machine learning. The transient phase does not contain such well-behaved dynamics.}
\label{fig:vks_phases}
\end{figure}

\subsection{Transient and steady-state VKS}
The VKS dataset is obtained by simulating the FOM in Section~\ref{ssec:vks} on the time interval $[0,400]$, containing two different regimes. When $t<100$, the dynamics lie in the transient state and approach the steady state as $t$ increases; while the dynamics maintain a steady state when $t\geq 100$, as shown in Fig.~\ref{fig:vks_phases}.

\begin{figure}[!ht]
\centering
 \begin{tabular}{c}
 \includegraphics[clip, trim=0.01cm 0.01cm 0.01cm 0.01cm, width=0.8\columnwidth]{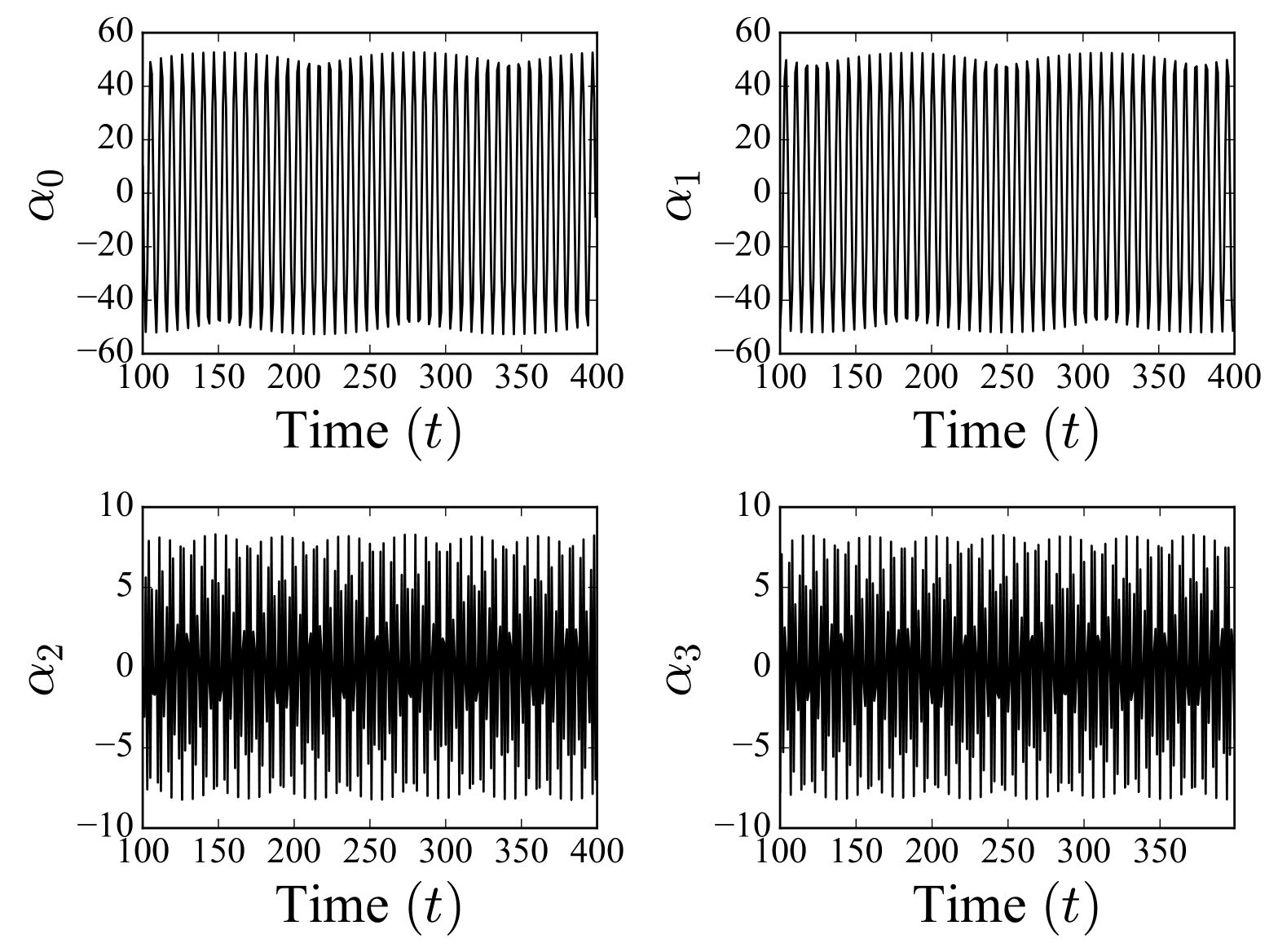}%{img/final/nonT_pred/vks_pod_modes.pdf}
 \end{tabular}
 \caption{The steady-state POD modes are highly oscillatory with quasi-periodic patterns.}
\label{fig:steady_pod_modes}
\end{figure}

\begin{figure}[!ht]
\centering
     \begin{tabular}{cc}
      \includegraphics[clip, trim=0.01cm 0.01cm 0.01cm 0.01cm, width=0.4\columnwidth]{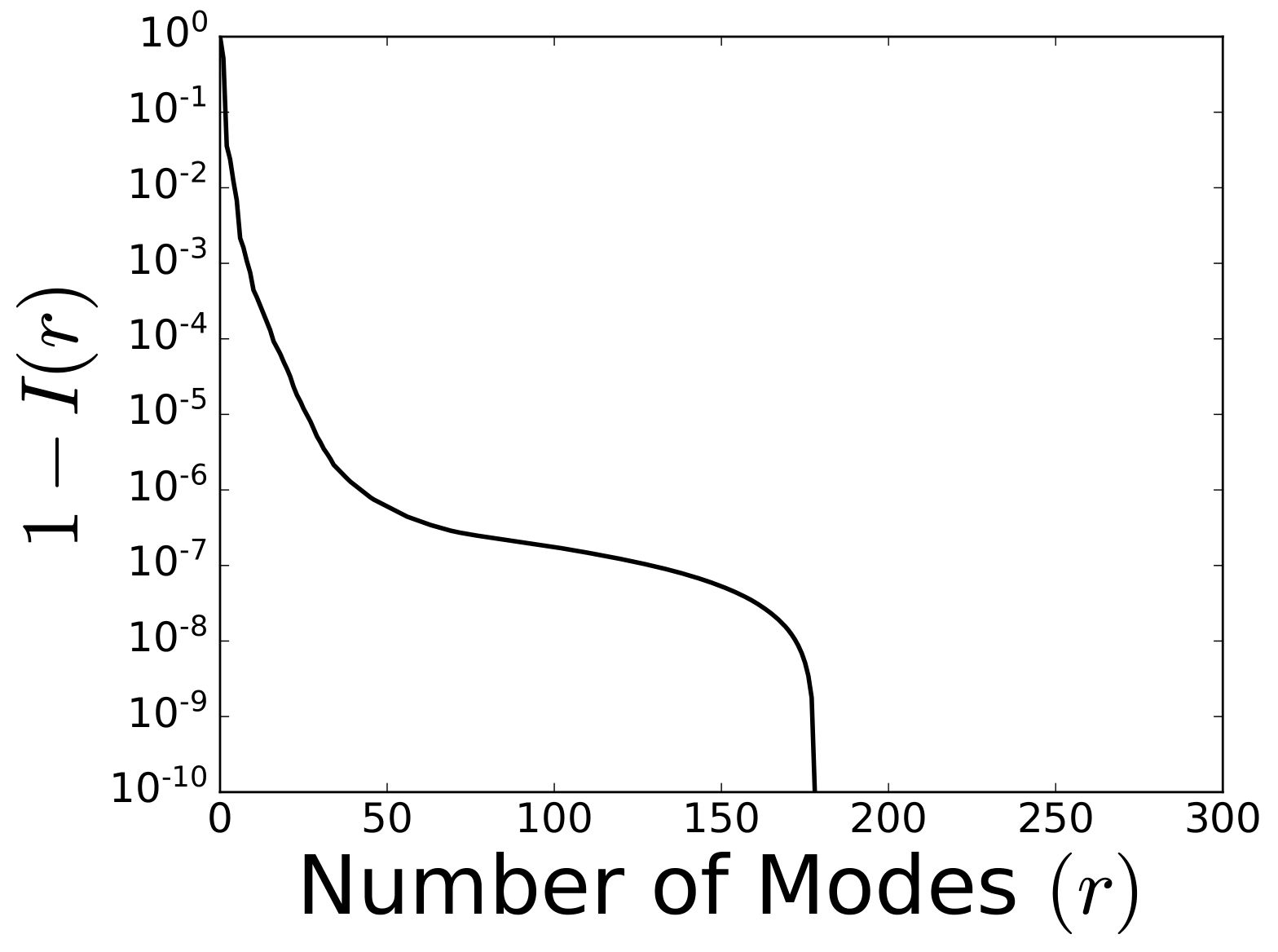}&
      %{img/final/nonT_pred/vks_pod_decay.pdf}
     \includegraphics[clip, trim=0.01cm 0.01cm 0.01cm 0.01cm, width=0.4\columnwidth]{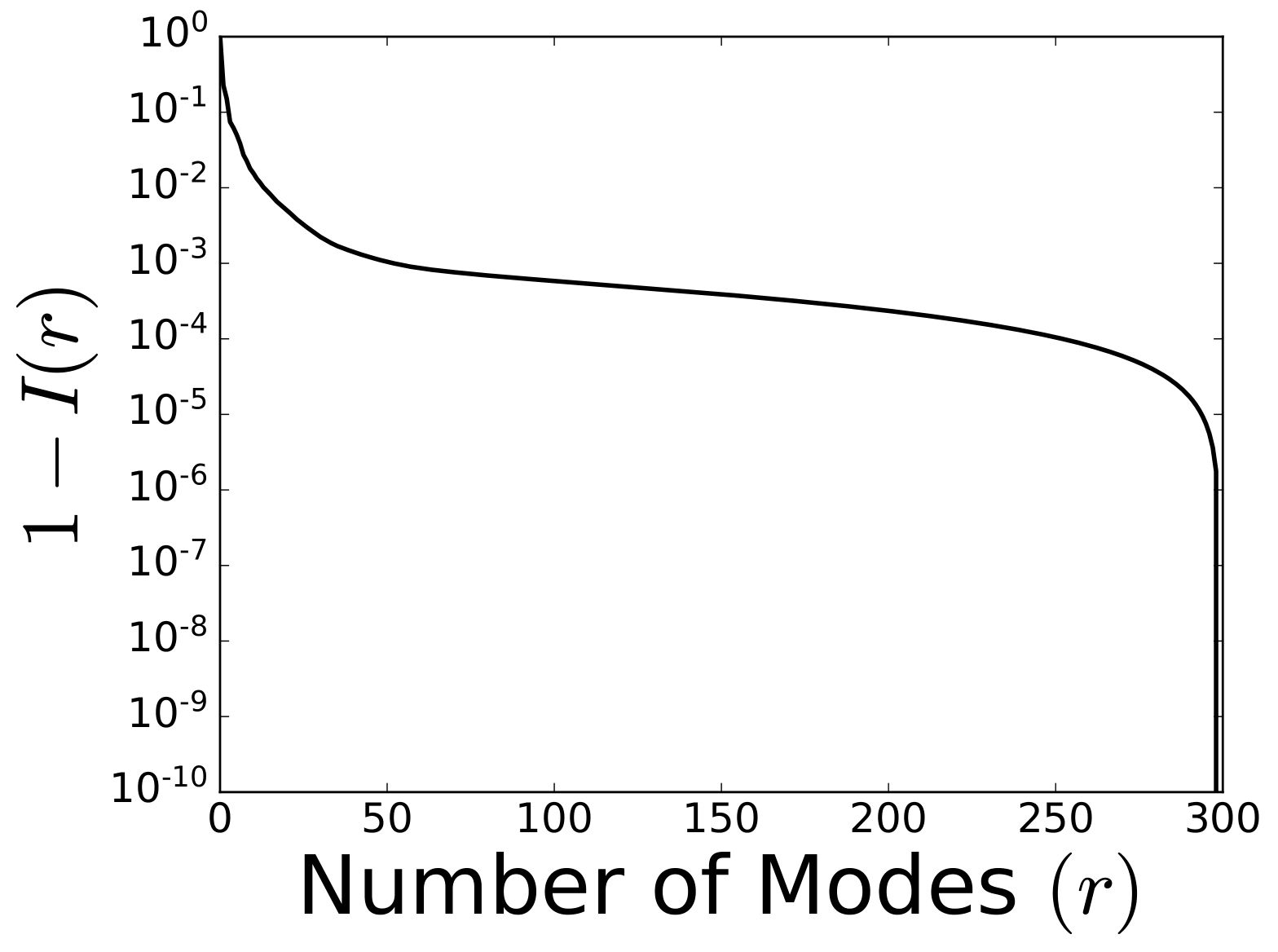}\\
     %{img/final/nonT_pred/vks_dmd_decay.pdf}
     (a) POD decay&
     (b) DMD decay\\
     \end{tabular}
     \caption{Comparison of the relative information decay for POD and DMD over the steady-state VKS. The POD modes decay far more rapidly than those of DMD. Therefore, one can expect better results with a smaller order of ROM using POD than DMD.} 
     \label{fig:steady_decay}
\end{figure}

\paragraph{ROMs for steady-state dynamics.}
We contrast different ROMs for simulating VKS flow in the steady-state regime. In particular, both the DMD and POD training is taken over the time interval from $t=100$ to $400$. The POD modes for the steady-state flow oscillate quasi-periodically, see Fig.~\ref{fig:steady_pod_modes}, and the relative information content $I(r)$ in \eqref{eq:I-def} decays rapidly in $r$. The POD relative information content for $8$ leading modes is $\sim 99\%$, as illustrated in Fig.~\ref{fig:steady_decay} (a). In contrast, the lifted DMD model, using the lifts $\{\cos(\vx),\sin(\vx),\vx^2,\vx^3\}$ with $\vx=(x,y)$, requires 24 modes to achieve $\sim 99\%$ relative information content, shown in Fig.~\ref{fig:steady_decay} (b). The quasi-periodic nature of the POD modes indicates that a model with high training accuracy will continue to perform well on the validation data.

\paragraph{ROMs for transient to steady state dynamics.}
ROMs behave very differently over the entire time interval from $t=0$ to $400$. The POD modes do not oscillate over the entire interval but only over the steady-state phase. For both POD and DMD, the relative information decays much slower. The POD relative information decreases to $\sim 96\%$ for the dominant $8$ modes.  While for the dominant $24$ lifted DMD modes, the relative information is reduced to $\sim 94\%$ over the full dynamics. This suggests that a machine learning-based ROM which is able to train on the transient phase to predict the steady-state phase accurately captures the intrinsic patterns of the underlying dynamics.

\begin{table}[!ht]
\fontsize{9.0}{9.0}\selectfont
\centering
\begin{threeparttable}
\begin{tabular}{cc}
\toprule[1.5pt]
\ \ \ \ \ \ \ \ \qquad {\bf Hyperparameter} \ \ \ \ \ \ \ \ \qquad  & \ \ \ \ \ \ \ \ \qquad Value \ \ \ \ \ \ \ \ \qquad \cr
\midrule[1.0pt]
Latent dimension & 6 \cr
     Layers encoder & 4 \cr
     Units encoder & 10 \cr
     Layers ODE & 12 \cr
     Units decoder & 41\cr
     Layers decoder & 4 \cr
     Learning rate & .00153 \cr
     Epochs & 2000\cr
\bottomrule[1.5pt]
\end{tabular}
\end{threeparttable}
\caption{The hyperparameters for the VAE architecture --- shown in Fig.~\ref{fig:vae_hbnode} --- for NODE and HBNODE-based ROMs. The parameters are tuned to the best NODE specification.}\label{Table:steady_hyper-vks}
\end{table}

\paragraph{Learning steady-state dynamics.}
In this task, we train the pipeline shown in Fig.~\ref{fig:vae_hbnode} for single-input-single-output dynamics prediction. Following the baseline in \cite{rom_node}, we train over the steady-state dynamics starting from $t=100$ using the dominant $8$ POD modes. The training data consists of the POD modes from $t=100$ to $174$, and the training labels consist of the POD modes from $t=101$ to $175$. The validation data consists of the POD modes from $t=175$ to $199$, with the objective to predict the POD modes at time steps from $t=176$ to $200$. We use the mean squared error to measure the loss between the labeled data and the predictions. We utilized an AdamW optimizer to train the network based on this loss criteria. For the black-box integration method, we selected DOPRI-5 \cite{DORMAND198019} with a relative tolerance of $1\mathrm{e}{-8}$. The model's hyperparameters are tuned to best the NODE as outlined in \cite{rom_node} and restated in Table~\ref{Table:steady_hyper-vks}. 

\begin{table}[!ht]
\fontsize{9.0}{9.0}\selectfont
\centering
\begin{threeparttable}
\begin{tabular}{cc}
\toprule[1.5pt]
\ \ \ \ \ \ \ \ \qquad {\bf Hyper-parameter} \ \ \ \ \ \ \ \ \qquad  & \ \ \ \ \ \ \ \ \qquad Value \ \ \ \ \ \ \ \ \qquad \cr
\midrule[1.0pt]
    Layers & 12 \cr
    Hidden layers & 64 \cr
    Sequence length & 9 \cr
    Learning rate & .001 \cr
    Epochs & 500 \cr
\bottomrule[1.5pt]
\end{tabular}
\end{threeparttable}
\caption{Hyperparameters of NODE and HBNODE for learning ROMs from transient to steady-state VKS dynamics.}\label{Table:full_hyper-vks}
\end{table}

\paragraph{Learning transient to steady-state dynamics.}
In this task, we train the pipeline outlined in Fig.~\ref{fig:rnn_node} for multi-input-single-output dynamics prediction. The objective of this task is to capture the phase transition at $t=100$. The data consisted of the dominant $8$ POD modes for the time interval from $t=0$ to $400$. The data was sequenced in a multi-input-single-output structure so that $9$ preceding time steps were used to predict the $10$-th time step. The training data consists of the POD modes for the transient time interval from $t=0$ to $79$. The training labels consisted of the POD modes from $t=10$ to $80$. The validation data utilizes the POD modes from steady-state time interval $t=80$ to $119$, and the validation labels consist of data from $t=90$ to $120$. The other experimental settings follow the above single-input-single-output scenario. The model's hyperparameters are the same for NODE and HBNODE components and are given in Table~\ref{Table:full_hyper-vks}.
\begin{figure}[!ht]
\centering
 \begin{tabular}{cc}
 \includegraphics[clip, trim=0.01cm 0.01cm 0.01cm 0.01cm, width=0.4\columnwidth]{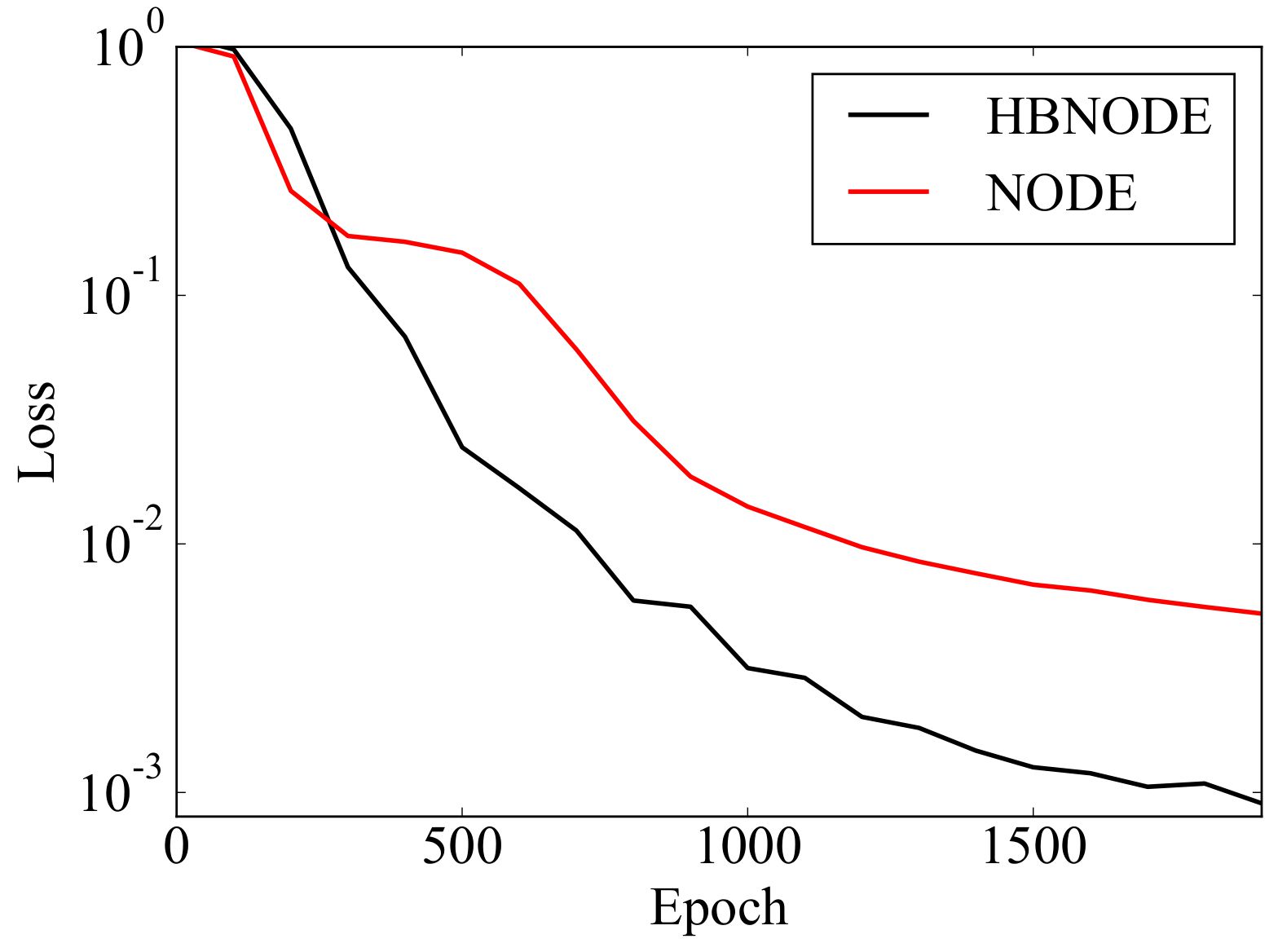}&
 %{img/final/nonT_pred/compare_tr_loss.pdf}
 \includegraphics[clip, trim=0.01cm 0.01cm 0.01cm 0.01cm, width=0.4\columnwidth]{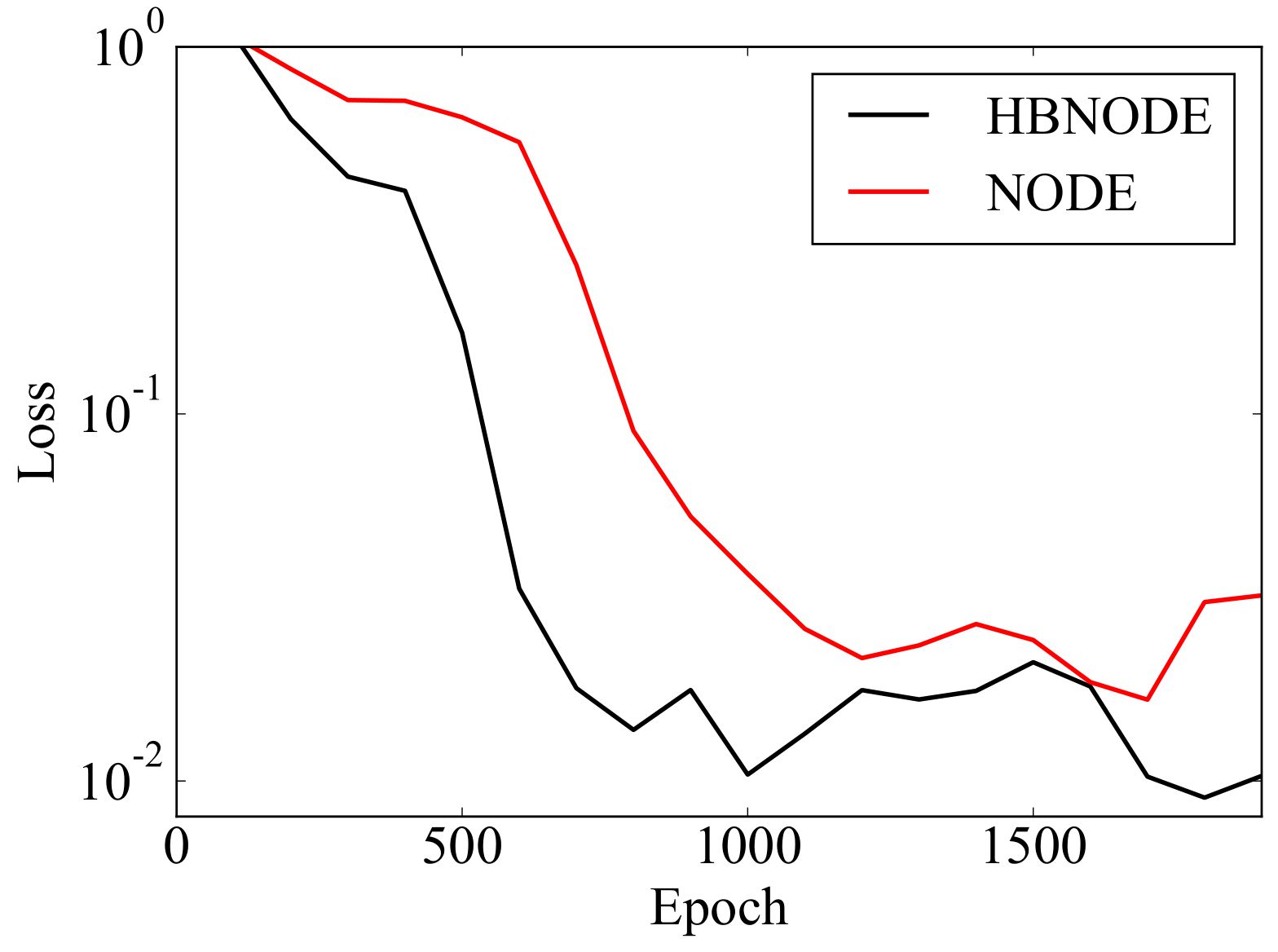}\\
 %{img/final/nonT_pred/compare_val_loss.pdf}
 (a) Training loss &
 (b) Validation loss
 \end{tabular}
 \caption{Contrasting NODE and HBNODE-based ROMs for learning steady-state VKS dynamics.
 HBNODE outperforms NODE in both training and validation loss.}
\label{fig:steady_compare_loss-vks}
\end{figure}

\subsubsection{Results and comparison to existing ROMs}
\begin{figure}[!ht]
\centering
 \begin{tabular}{cc}
 \includegraphics[clip, trim=0.01cm 7.5cm 10cm .01cm, width=0.4\columnwidth]{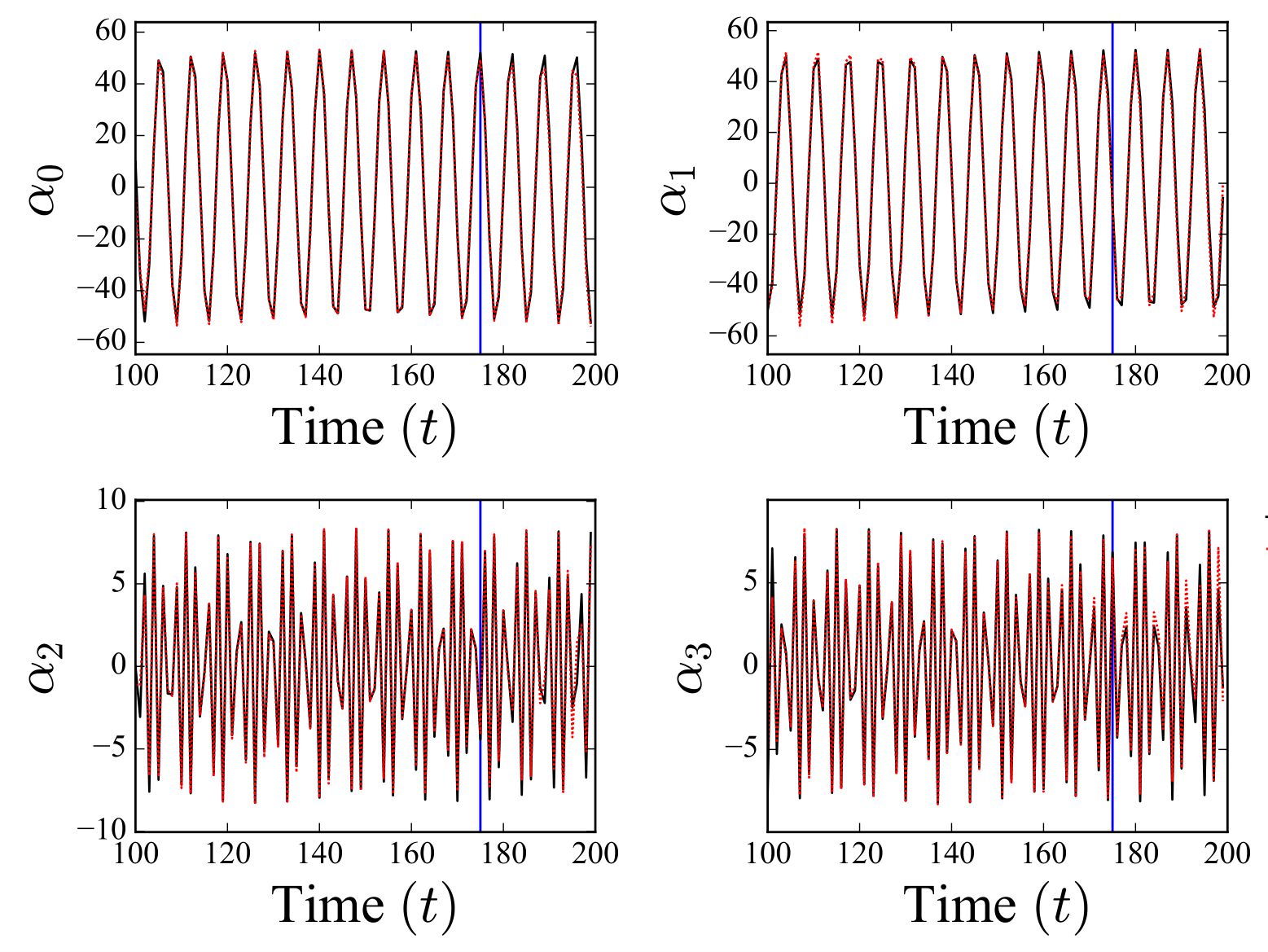}&
 %{img/new/nonT_pred/vks_vae_node_mode_pred.pdf}
 \includegraphics[clip, trim=0.01cm 7.5cm 10cm .01cm, width=0.4\columnwidth]{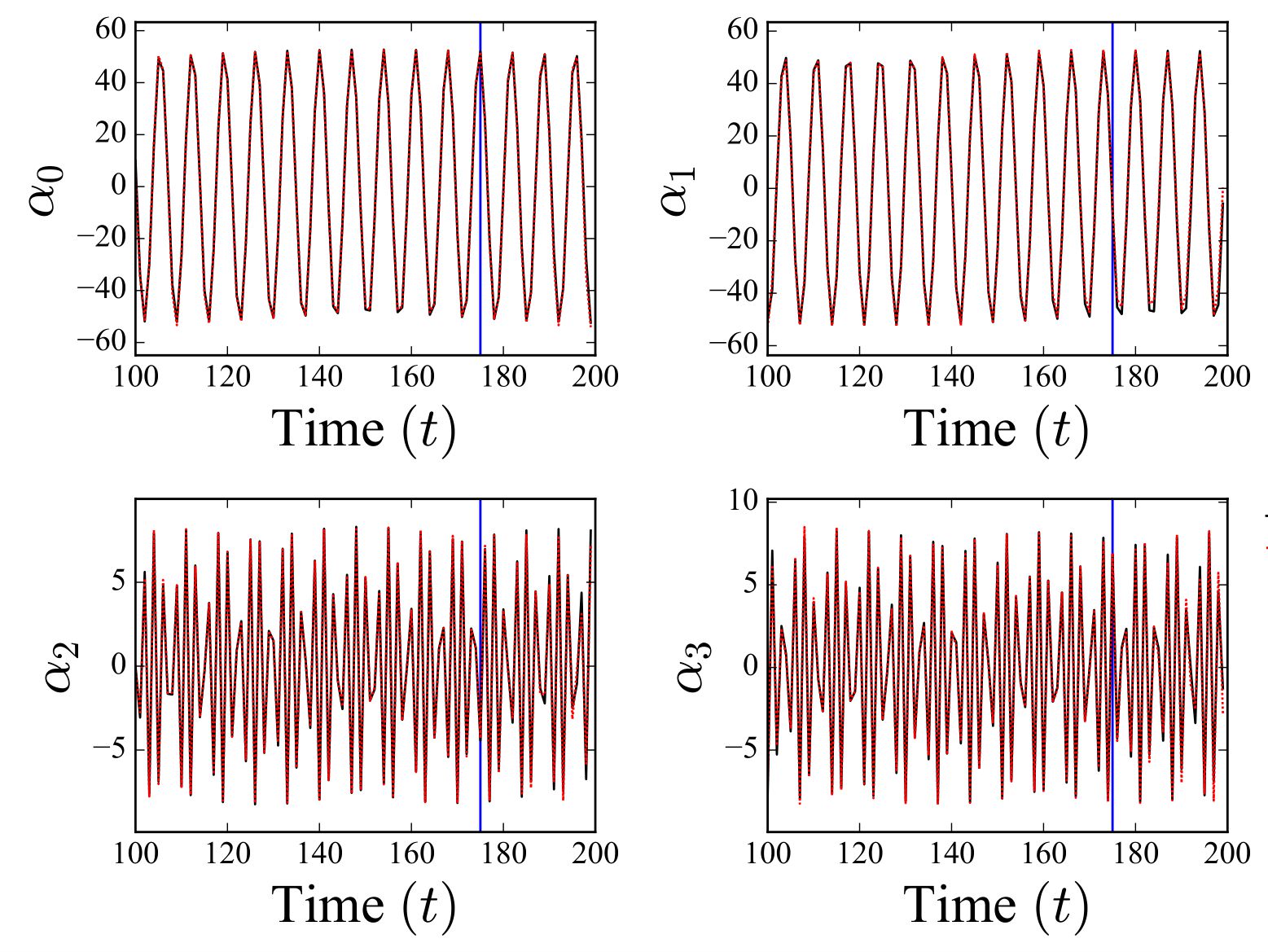}\\
 %{img/new/nonT_pred/vks_vae_hbnode_mode_pred.pdf}
 (a) VAE-NODE dominant mode &
  (b) VAE-HBNODE dominant mode\\
 \end{tabular}
 \caption{Comparison of modes reconstruction for NODE and HBNODE-based ROMs for learning steady-state VKS dynamics. HBNODE captures the peaks of the dominant POD modes better than NODE. Before and after the vertical blue line stands for training and validation, respectively.}
\label{fig:steady_compare_mode-vks}
\end{figure}

\paragraph{Results of learning steady-state dynamics.}
We contrast HBNODE and NODE-based ROMs in Fig.~\ref{fig:steady_compare_loss-vks} and Fig.~\ref{fig:steady_compare_mode-vks}. Figure~\ref{fig:steady_compare_loss-vks} shows that HBNODE-based ROM not only achieves remarkably smaller training loss but also significantly smaller validation loss than NODE-based ROM. In terms of the predictive performance, we see that HBNODE performs better at capturing several of the peaks of the oscillatory modes as shown in Fig.~\ref{fig:steady_compare_mode-vks}.

\begin{figure}[!ht]
\centering
 \begin{tabular}{cc}
 \includegraphics[clip, trim=0.01cm 0.01cm 0.01cm 0.01cm, width=0.4\columnwidth]{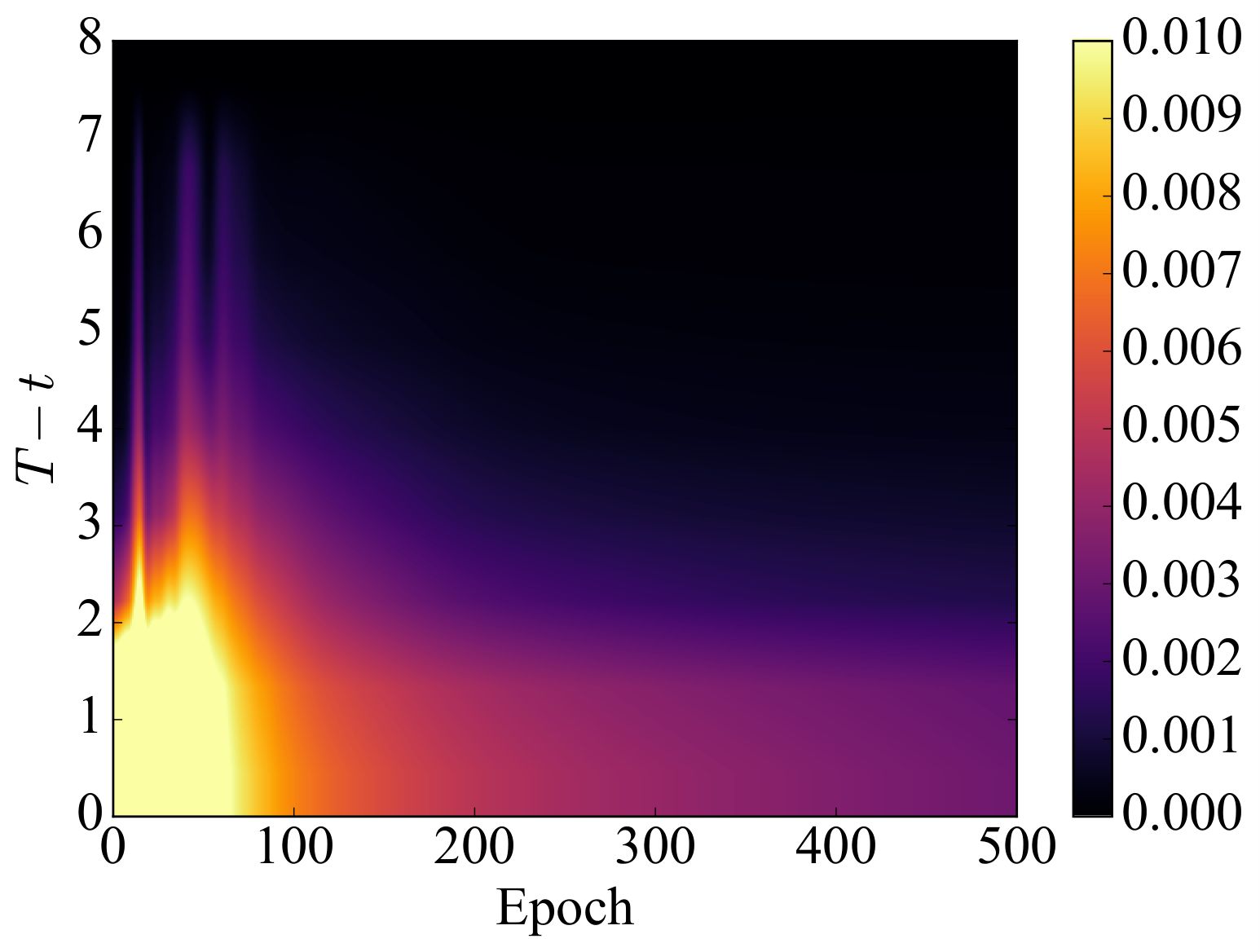}&
 %{img/final/full_pred/NODE_adjGrad.pdf}
 \includegraphics[clip, trim=0.01cm 0.01cm 0.01cm 0.01cm, width=0.4\columnwidth]{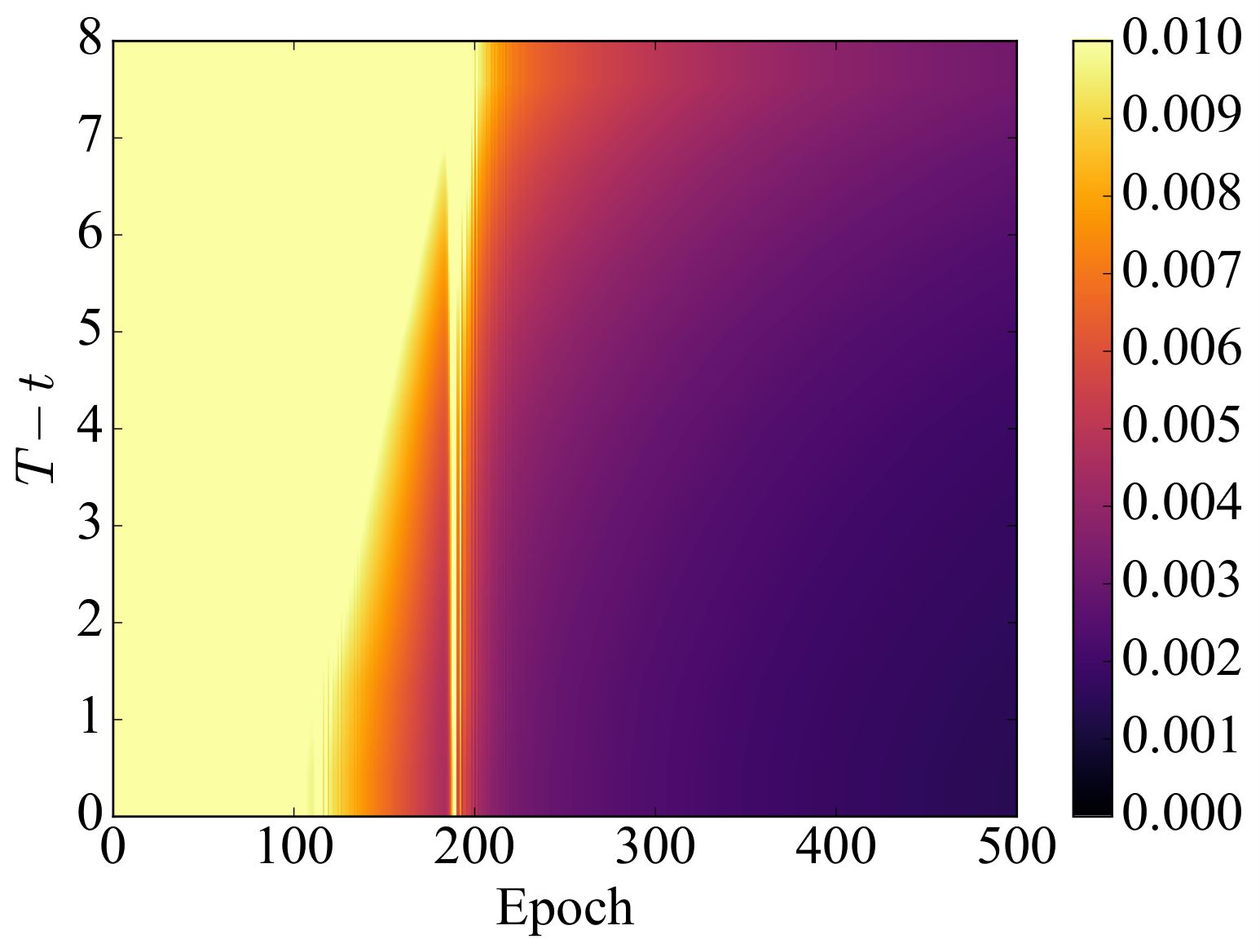}\\
 %{img/final/full_pred/HBNODE_adjGrad.pdf}
 (a) NODE adjoint state &
 (b) HBNODE adjoint state
 \end{tabular}
 \caption{Comparison of the adjoint states for the NODE and HBNODE in learning multi-input-single-output. The NODE adjoint state vanishes substantially faster than HBNODE.}
\label{fig:sfull_adjGrad-vks}
\end{figure}

\paragraph{Results of learning transient to steady-state dynamics.}
\begin{figure}[!ht]
\centering
 \begin{tabular}{cc}
\includegraphics[clip, trim=0.01cm 0.01cm 0.01cm 0.01cm, width=0.4\columnwidth]{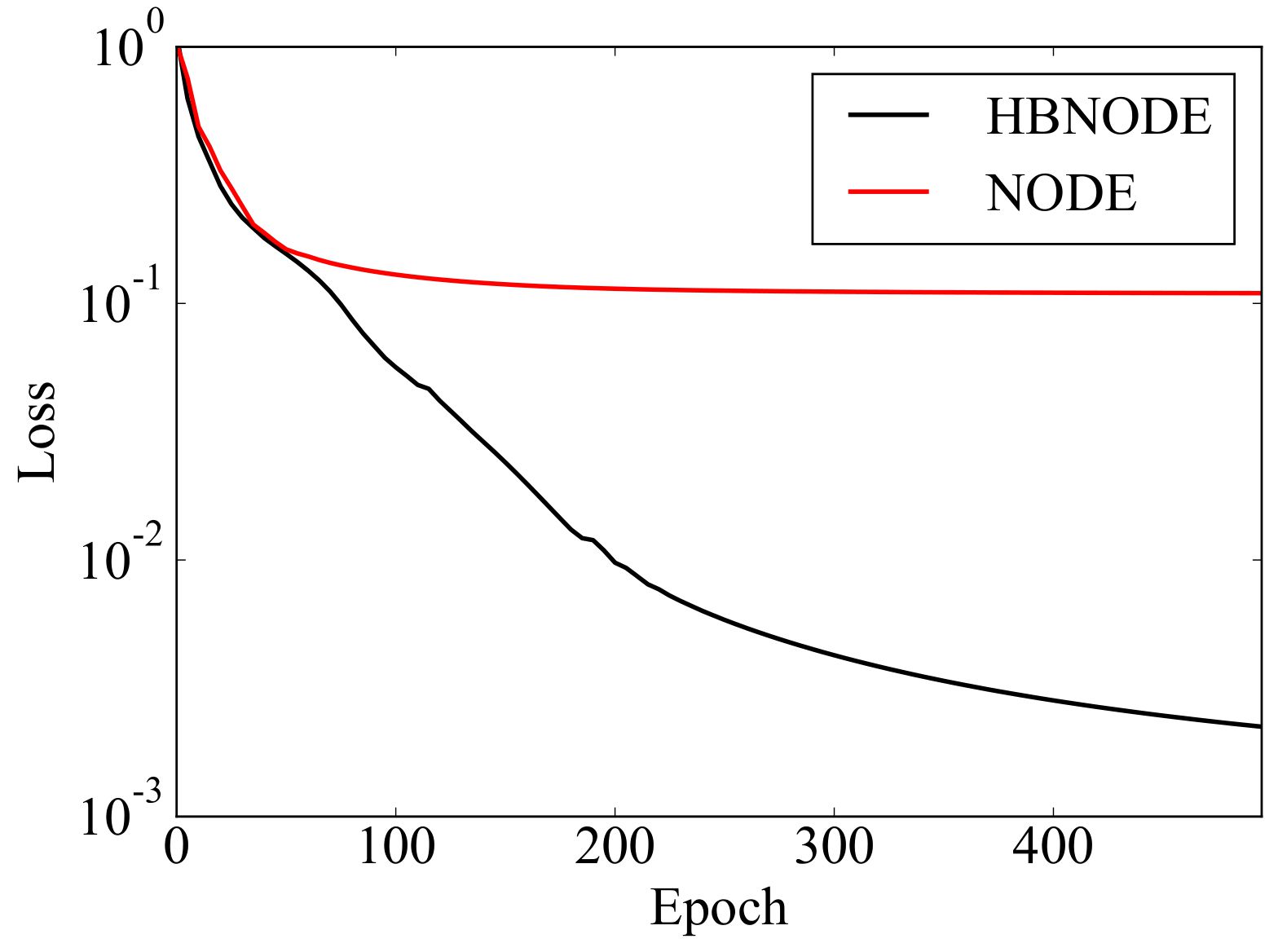}&
%{img/final/full_pred/compare_tr_loss.pdf}
 \includegraphics[clip, trim=0.01cm 0.01cm 0.01cm 0.01cm, width=0.4\columnwidth]{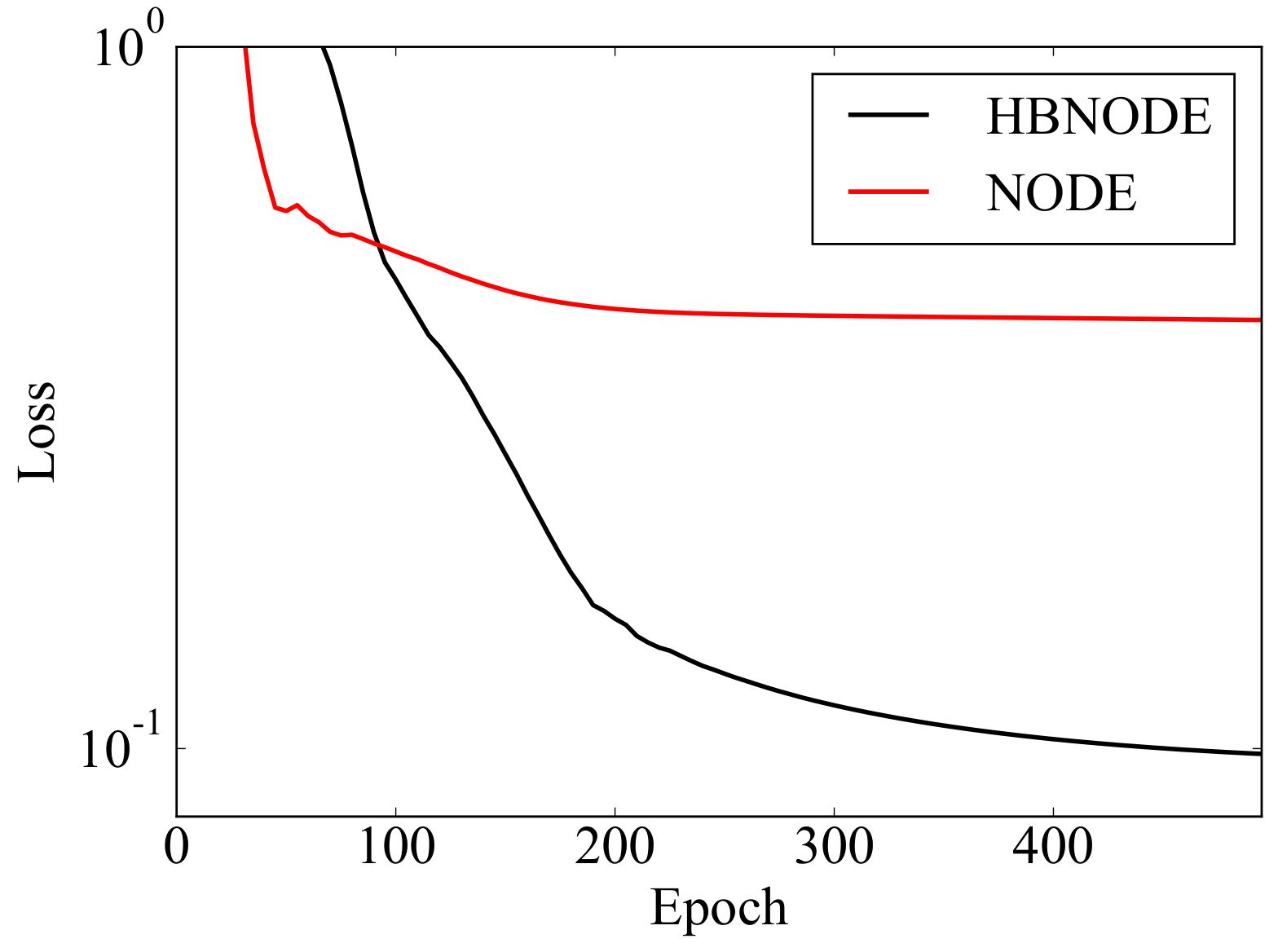}\\
 %{img/final/full_pred/compare_val_loss.pdf}
 (a) Training loss &
 (b) Validation loss
 \end{tabular}
 \caption{Contrasting training and validation loss of NODE and HBNODE for learning ROMs of transient to steady-state VKS dynamics. The progress made by the NODE is significantly reduced by the vanishing gradient. HBNODE has a much slower decaying gradient and is able to continue to make progress in both training and validation sets.}
\label{fig:full_loss-vks}
\end{figure}
Compared to learning steady-state VKS dynamics, HBNODE achieves more significant performance gain over NODE for learning transient to steady-state dynamics in terms of training and validation loss, as shown in Fig.~\ref{fig:full_loss-vks}. Since we are doing sequential learning, one interpretation of the improvement in learning dynamics is the effective learning of long-range dependencies. Indeed, the criterion of learning long-range dependencies has been widely used in measuring the efficacy of sequential learning models \cite{pascanu2013difficulty,HBNODE:2021}. In NODE and HBNODE, the effectiveness of learning long-range dependencies can be measured by whether the adjoint state vanishes quickly or not. We visualize the evolution of the magnitude of the adjoint states of NODE and HBNODE in Fig.~\ref{fig:sfull_adjGrad-vks}, which support the theoretical result in Section~\ref{subsubsec-long-term-dependencies}. In particular, we see that the adjoint state of NODE vanishes much more rapidly than that of HBNODE as $T-t$ increases. A more detailed connection between the adjoint state and learning long-range dependencies is provided in \cite{HBNODE:2021}.

\begin{figure}[!ht]
\centering
 \begin{tabular}{cc}
\includegraphics[clip, trim=0.01cm 0.01cm 0.01cm 0.01cm, width=0.4\columnwidth]{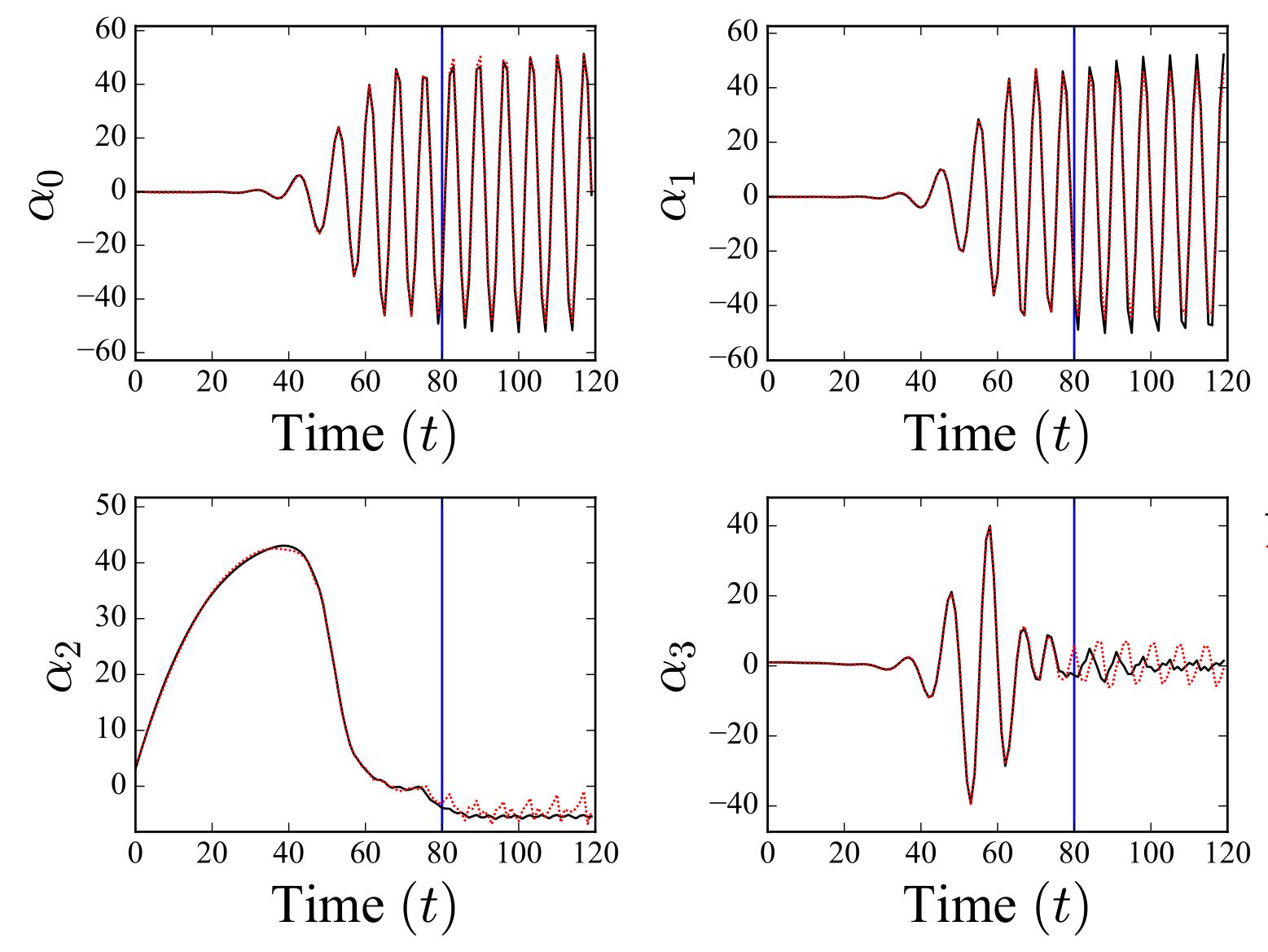}&
%{img/new/full_pred/vks_seq_node_mode_pred.pdf}
\includegraphics[clip, trim=0.01cm 0.01cm 0.01cm 0.01cm, width=0.4\columnwidth]{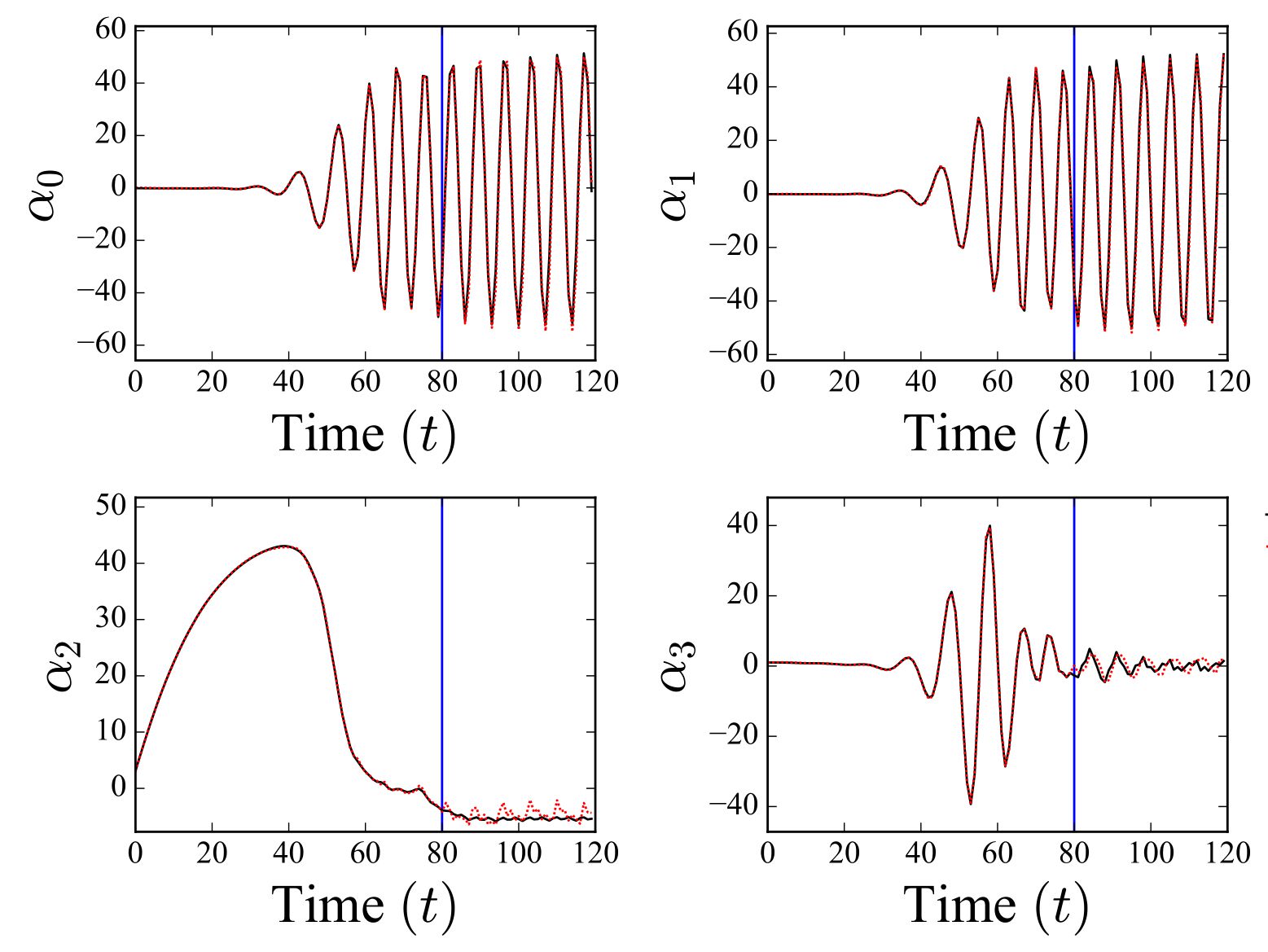}\\
%{img/new/full_pred/vks_seq_hbnode_mode_pred.pdf}
 (a) VAE-NODE modes &
 (b) VAE-HBNODE modes
 \end{tabular}
 \caption{Comparison of modes reconstruction for NODE and HBNODE-based ROMs for learning multi-input-single-output task. NODE is able to reliably learn the first two dominant modes but is unable to capture the steady-state dynamics, having a much larger frequency and introducing substantial lag into the oscillation frequency. HBNODE is able to more reliably capture the steady-state dynamics with slightly larger frequency and developing lag much later than the NODE component. Before and after the vertical blue line stands for training and validation, respectively.}
\label{fig:full_mode_comparison-vks}
\end{figure}

In terms of the predictive performance, as shown in Fig.~\ref{fig:full_mode_comparison-vks}, the HBNODE predictor captures the peaks of the oscillatory dynamics better than NODEs, especially in the first two modes $\alpha_1$ and $\alpha_2$. Moreover, the prediction error using NODE is much larger than that of HBNODE, and the prediction error amplifies as the prediction time goes, in particular, for modes $\alpha_3$ and $\alpha_4$.

\begin{figure}[!ht]
\centering
 \begin{tabular}{cc}
 \includegraphics[clip, trim=0.01cm 0.01cm 0.01cm 0.01cm, width=0.4\columnwidth]{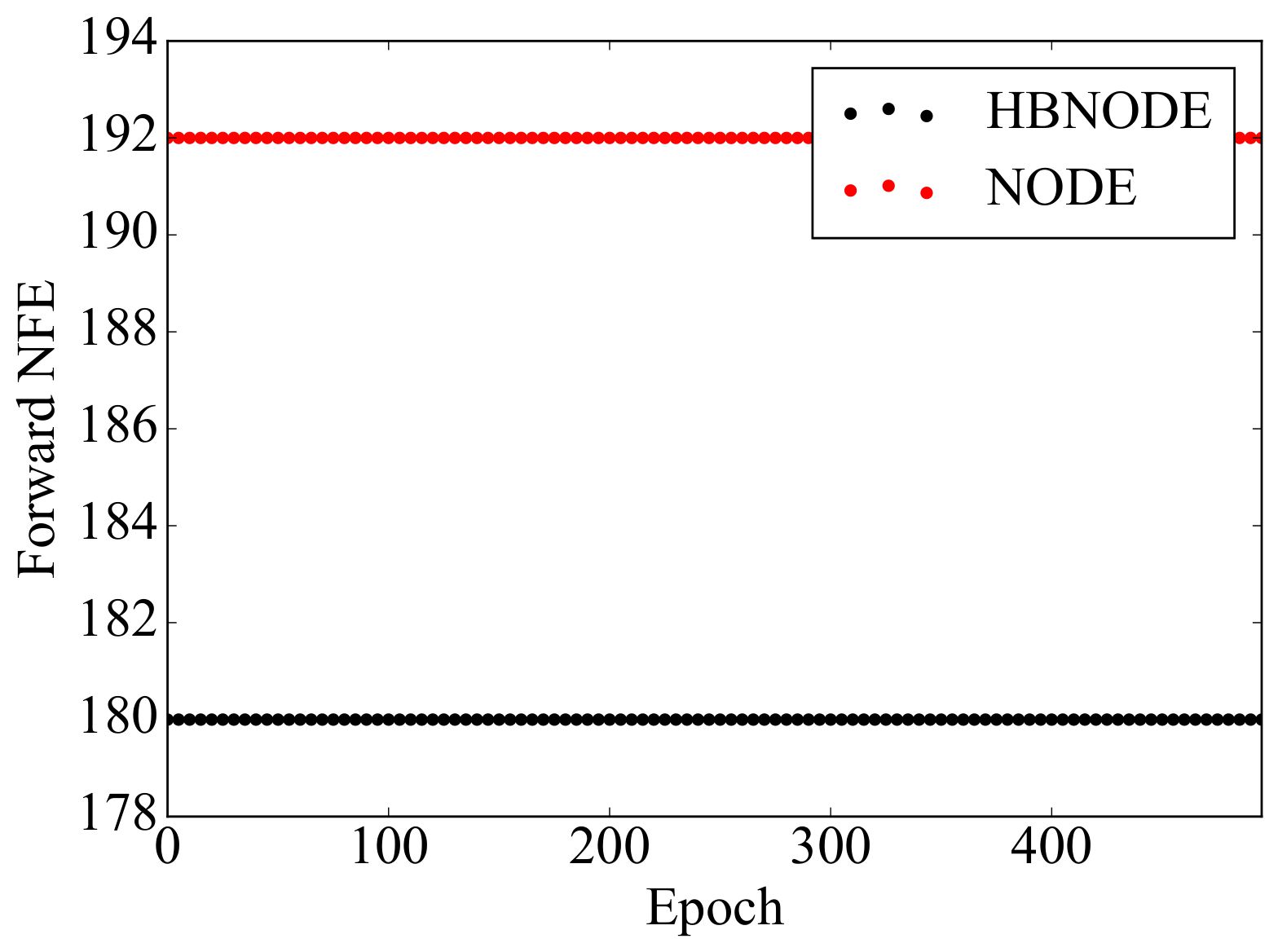}&
 %{img/final/full_pred/compare_forward_nfe.pdf}
 \includegraphics[clip, trim=0.01cm 0.01cm 0.01cm 0.01cm, width=0.4\columnwidth]{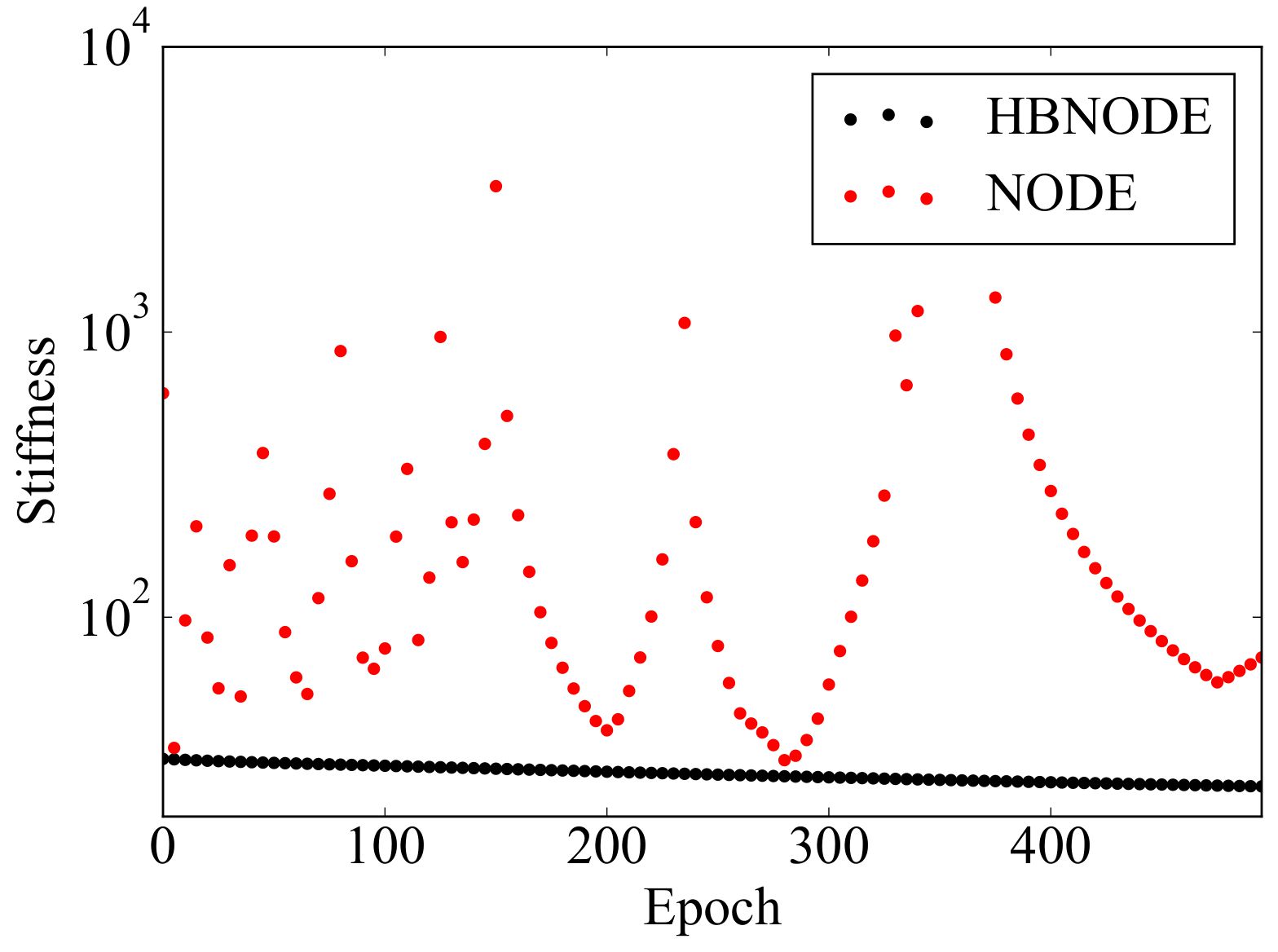}\\
 %{img/final/full_pred/compare_forward_stiff.pdf}
 (a) Forward NFE comparison &
 (b) Stiffness comparison
 \end{tabular}
 \caption{Comparison of NFEs and stiffness for the NODE and HBNODE in learning transient to steady-state VKS dynamics. NODE requires more NFEs in each forward pass than HBNODE as the NODE is much stiffer than HBNODE. The stiffness of NODE varies sharply as training goes on, while the stiffness of HBNODE decays during the training.
 %Furthermore, the NODE component is unstable in its stiffness while the stiffness of the NODE component decays over the training epochs. 
% \BW{x,y-labels}
 %\justin{I have comparison stiffness plots}
}
\label{fig:full_nfe_stiff-vks}
\end{figure}

Another primary advantage of HBNODE over NODE-based ROMs lies in computational efficiency, which is theoretically supported by the discussion in Section~\ref{subsubsec-computational-advantages}. As shown in Fig.~\ref{fig:full_nfe_stiff-vks} (a), the forward NFE required in each forward pass by HBNODE is consistent smaller than that of NODE. We also monitor the stiffness of both NODE and HBNODE during the learning process, and Fig.~\ref{fig:full_nfe_stiff-vks} (b) shows that the stiffness of NODE oscillates and maintains much larger than HBNODE.

\begin{figure}[!ht]
\centering
 \begin{tabular}{cc}
 \includegraphics[clip, trim=0.01cm 0.01cm 0.01cm 0.01cm, width=0.4\columnwidth]{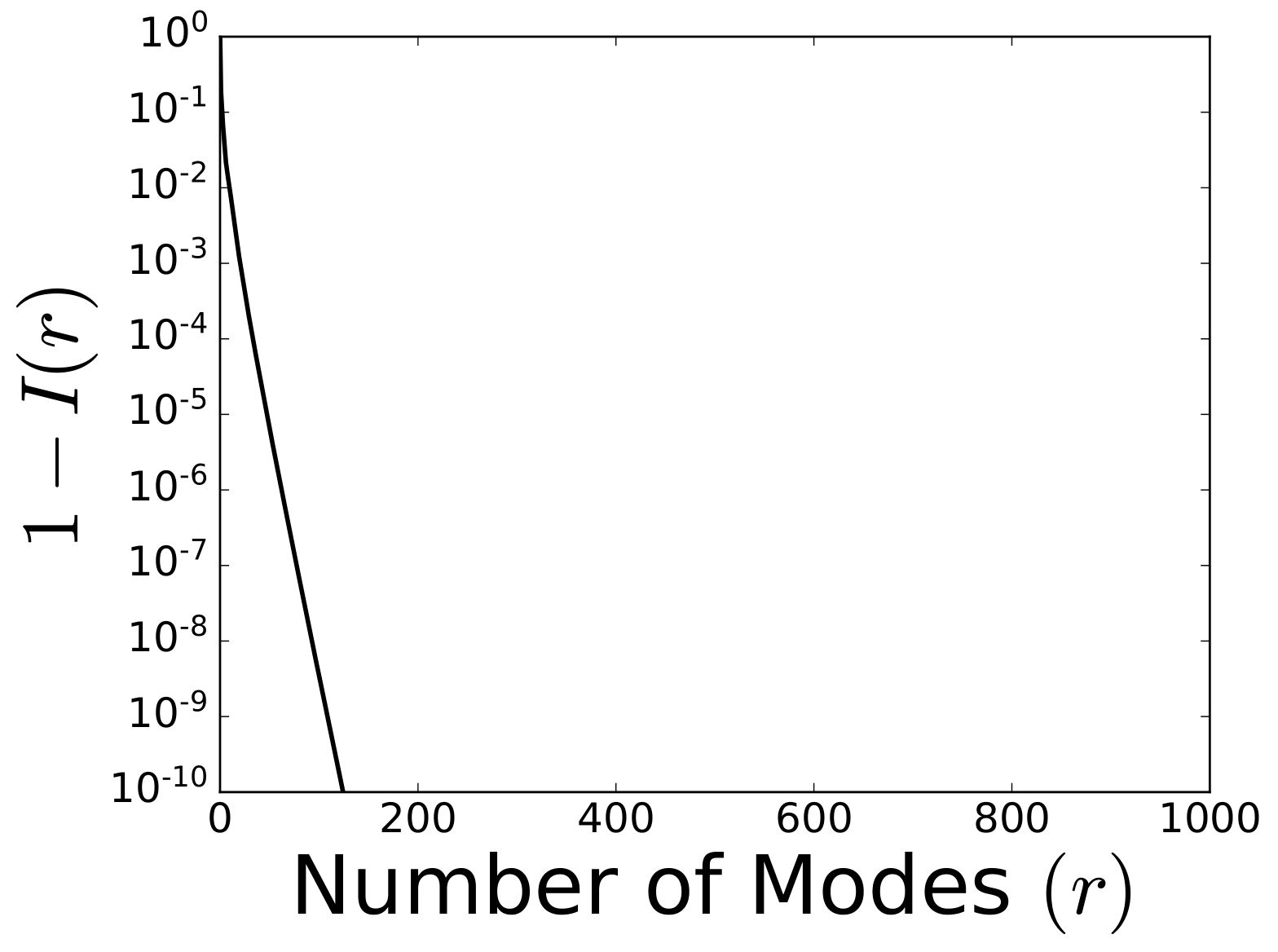}&
 %{img/final/kpp/kpp_pod_decay.pdf}
 \includegraphics[clip, trim=0.01cm 0.01cm 0.01cm 0.01cm, width=0.4\columnwidth]{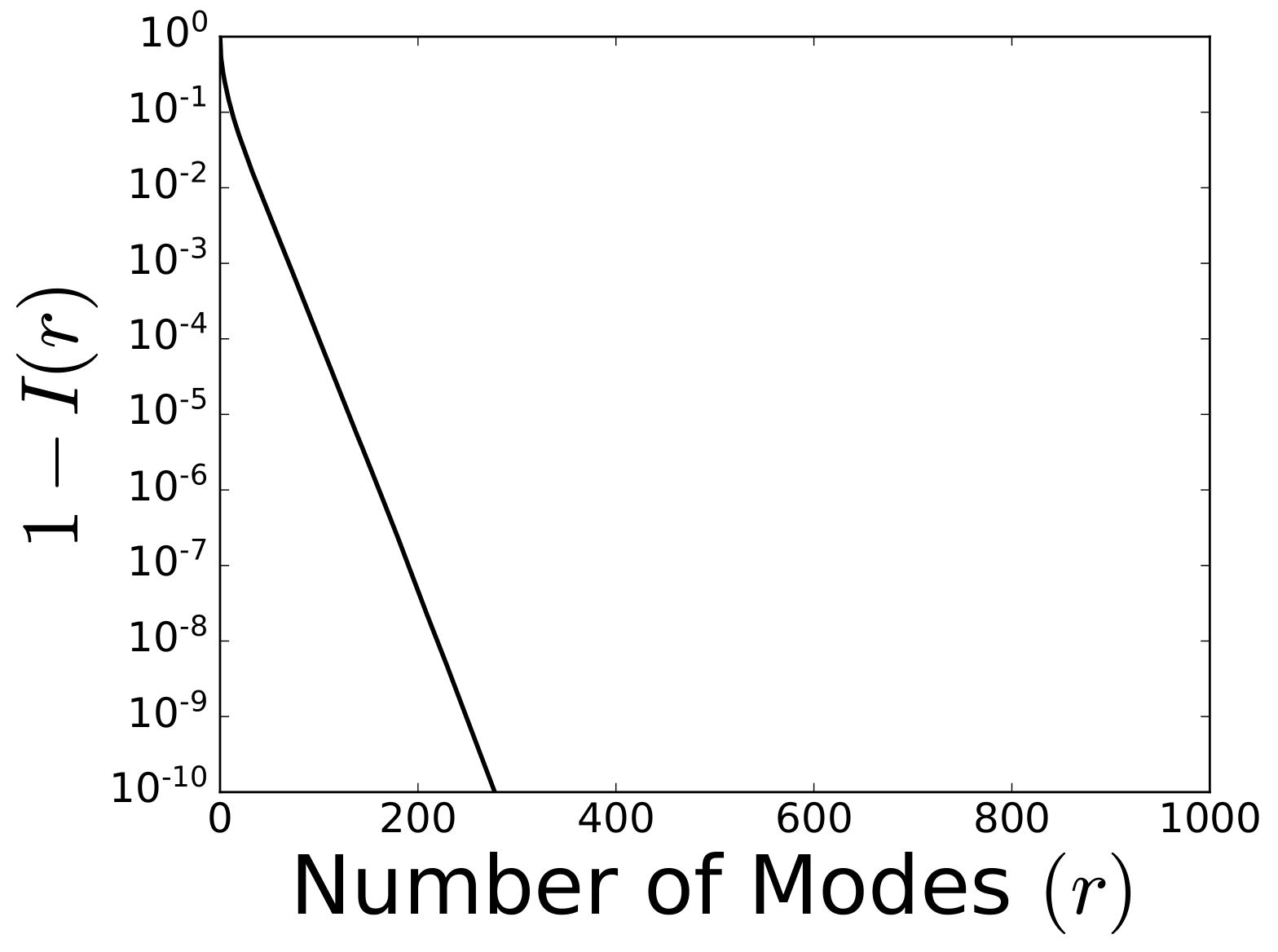}\\
 %{img/final/kpp/kpp_dmd_decay.pdf}
  (a) POD mode decay &
(b) DMD mode decay
 \end{tabular}
 \caption{The decay of relative information content for POD and DMD of the KPP dataset. The rapid decay in the modes indicates the problem is suitable for POD and DMD.}
\label{fig:kpp_decay}
\end{figure}
\subsection{KPP model}
\begin{figure}[!ht]
\centering
 \begin{tabular}{c}
 \includegraphics[clip, trim=0.01cm 0.01cm 0.01cm 0.01cm, width=0.8\columnwidth]{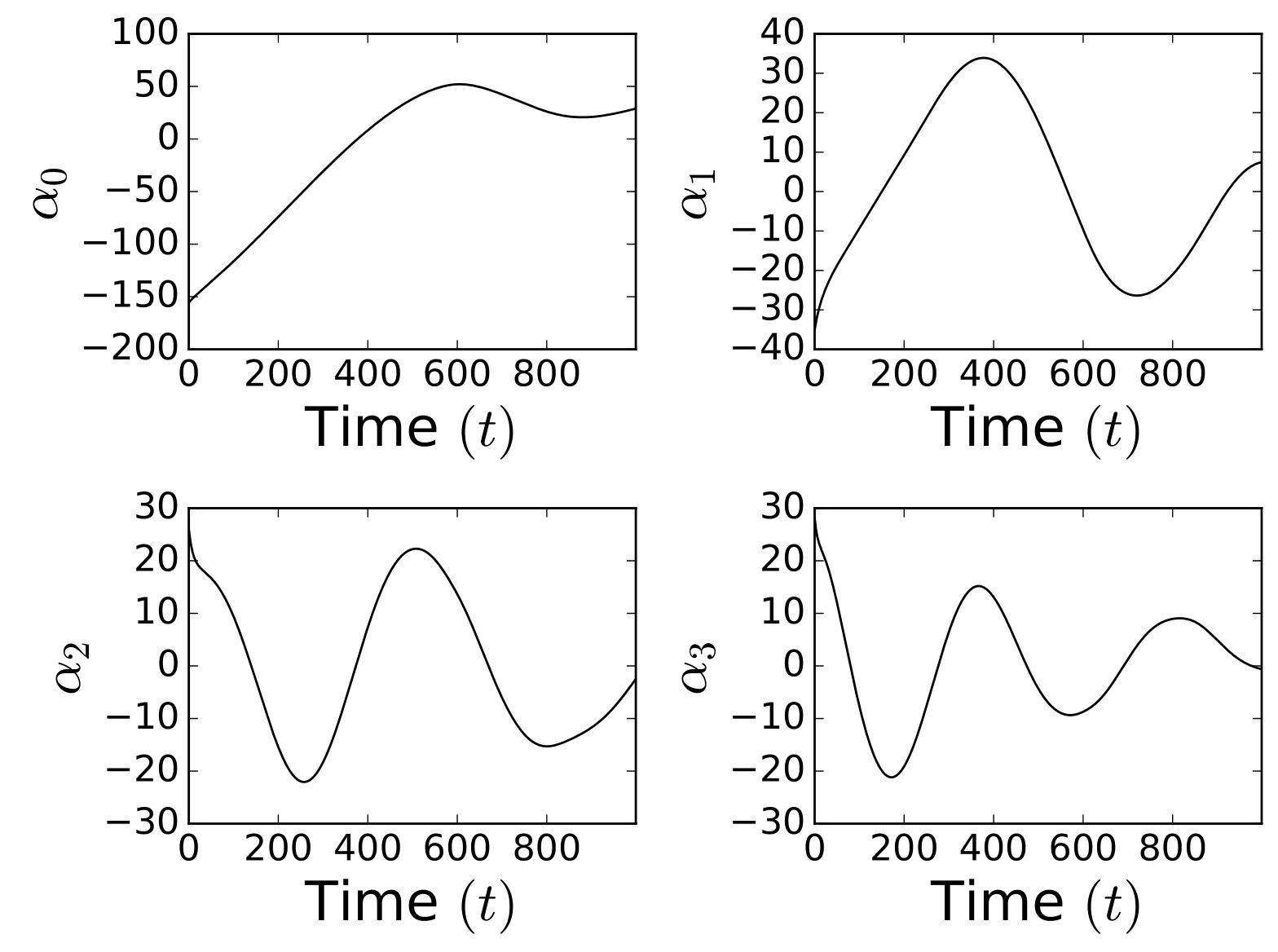}\\
 %{img/final/kpp/kpp_pod_modes.pdf}
 KPP POD modes
 \end{tabular}
 \caption{The dominant POD modes for the KPP data. The oscillations of the modes decay rapidly and have a very low frequency, making it challenging for learning compared to the quasi-periodic oscillations of the VKS dataset.}
\label{fig:kpp_modes}
\end{figure}

We obtain the KPP dataset by simulating the FOM presented in Section~\ref{ssec:kpp} for $1000$ timesteps ($N_t = 1000$). The KPP model is well-suited for reduced-order modeling due to the rapidly decaying eigenvalues in both POD and DMD, seeing Fig.~\ref{fig:kpp_decay}. However, we found in our experiments that it is particularly difficult to capture the dynamics using machine learning architectures due to the slow decaying ROM dynamics depicted in Fig.~\ref{fig:kpp_modes}.

In our experiments, we note that the non-lifted DMD continuously deforms the center of mass in a way that defies the physical constraints of the system. A comparison of lifted and non-lifted DMD predictions and POD predictions are shown in \cite{github-animation}. To lift DMD, we utilized the lifting functions $\{\cos(\vx),\sin(\vx),\vx^2,\vx^3\}$ with $\vx=(x,y)$. Figure~\ref{fig:kpp_decay} shows that the POD modes decay faster than the lifted DMD modes. The dominant $24$ DMD modes correspond to $97\%$ of the relative information content; in contrast, the dominant $8$ POD modes correspond to $99\%$ of the relative information value.

We train the pipeline depicted in Fig.~\ref{fig:rnn_node} for learning multi-input-single-output dynamics. The data is constructed from the $8$ dominant POD modes on the time interval from $t=0$ to $1000$. The data is sequenced so that every $4$ preceding time step is used to predict the $5$-th time step. The training data consists of the POD modes from $t=0$ to $799$ and the training labels consist of POD modes from $t=5$ to $800$. The validation data utilizes data from $t=800$ to $999$ and the validation labels consist of data from $t=805$ to $1000$. The model's hyperparameters are the same for NODE and HBNODE components and are given in Table~\ref{Table:full_hyper-kpp}.
\begin{table}[!ht]
\fontsize{9.0}{9.0}\selectfont
\centering
\begin{threeparttable}
\begin{tabular}{cc}
\toprule[1.5pt]
\ \ \ \ \ \ \ \ \qquad {\bf Hyper-parameter} \ \ \ \ \ \ \ \ \qquad  & \ \ \ \ \ \ \ \ \qquad Value \ \ \ \ \ \ \ \ \qquad \cr
\midrule[1.0pt]
    Layers & 2 \cr
    Hidden layers & 64 \cr
     Sequence length & 4 \cr
     Learning rate & .01 \cr
     Epochs & 500 \cr
\bottomrule[1.5pt]
\end{tabular}
\end{threeparttable}
\caption{The hyperparameters of NODE and NODE for the learning ROMs of KPP model.}
\label{Table:full_hyper-kpp}
\end{table}

\subsubsection{Results and comparison to existing ROMs}
\begin{figure}[!ht]
\centering
 \begin{tabular}{cc}
 \includegraphics[clip, trim=0.01cm 0.01cm 0.01cm 0.01cm, width=0.46\columnwidth]{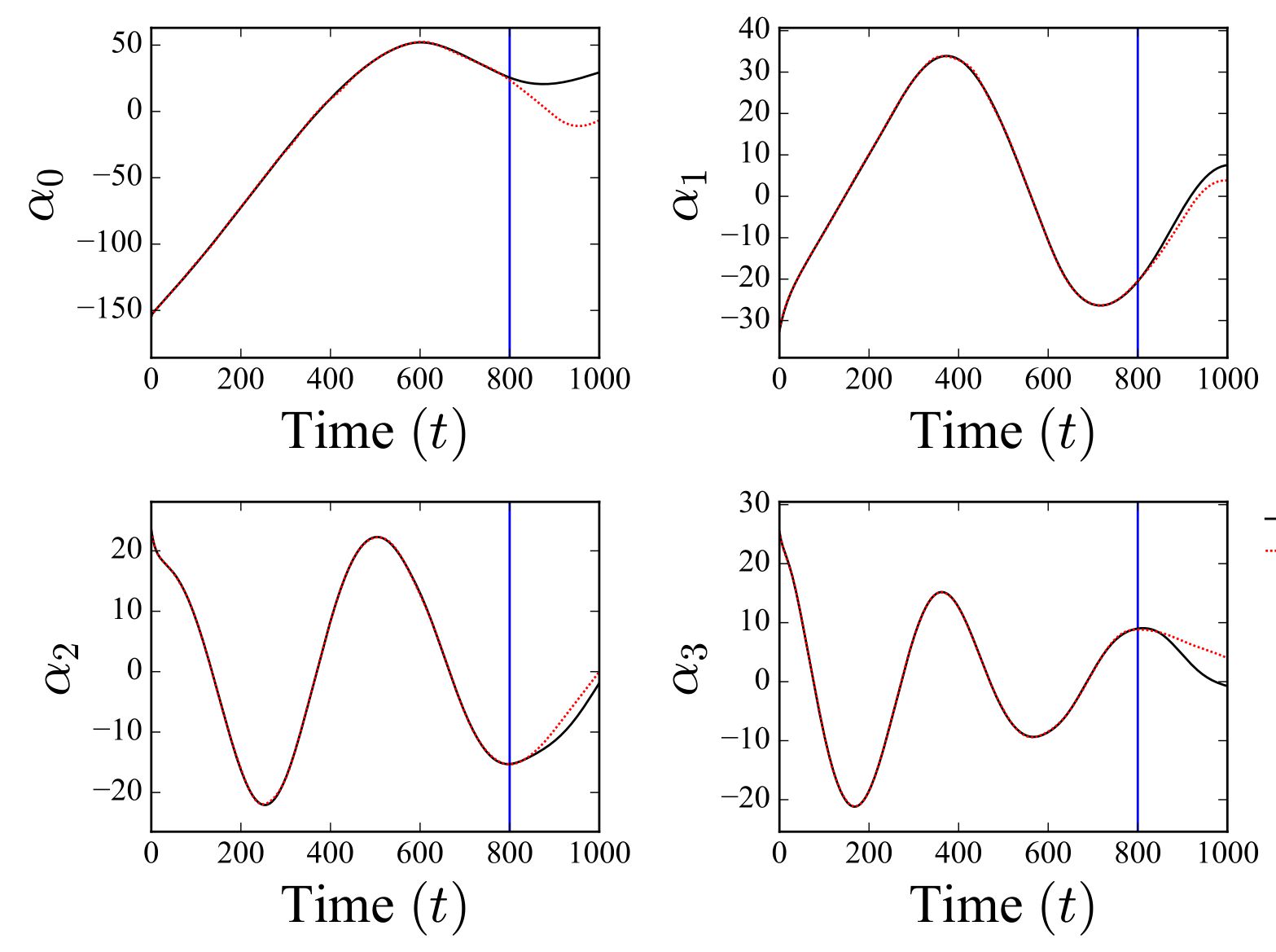}&
 %{img/new/kpp/kpp_seq_node_mode_pred.pdf}
 \includegraphics[clip, trim=0.01cm 0.01cm 0.01cm 0.01cm, width=0.46\columnwidth]{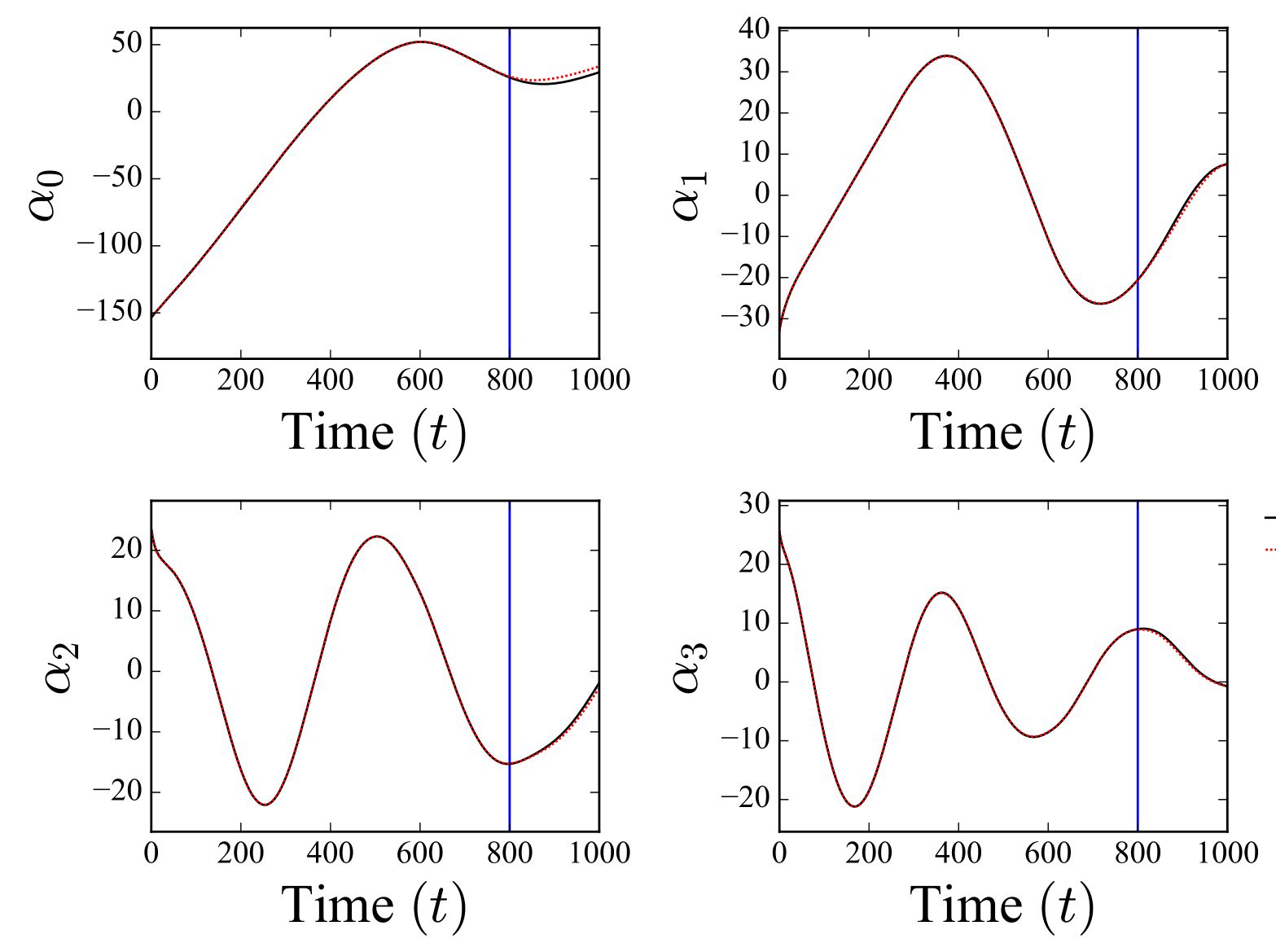}\\
 %{img/new/kpp/kpp_seq_hbnode_mode_pred.pdf}
 (a) NODE mode recapture &
 (b) HBNODE mode recapture
 \end{tabular}
 \caption{Comparison of the KPP modes prediction using NODE and HBNODE. HBNODE is better at predicting the dynamics for the validation set than NODE. Before and after the vertical blue line stands for training and validation, respectively.}
\label{fig:kpp_mode_prediction}
\end{figure}

\begin{figure}[!ht]
\centering
 \begin{tabular}{cc}
 \includegraphics[clip, trim=0.01cm 0.01cm 0.01cm 0.01cm, width=0.4\columnwidth]{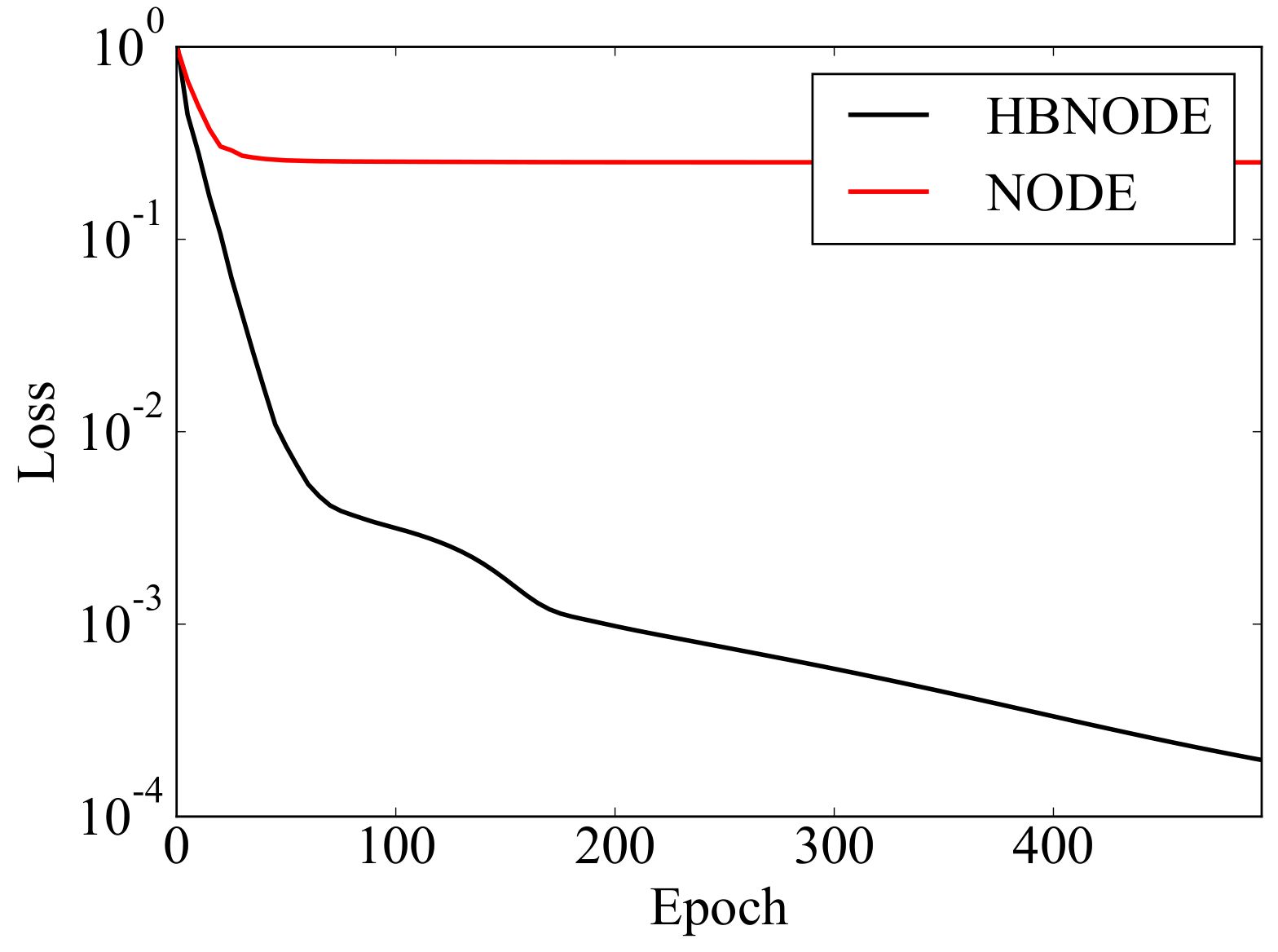}&
 %{img/final/kpp/compare_tr_loss.pdf}
 \includegraphics[clip, trim=0.01cm 0.01cm 0.01cm 0.01cm, width=0.4\columnwidth]{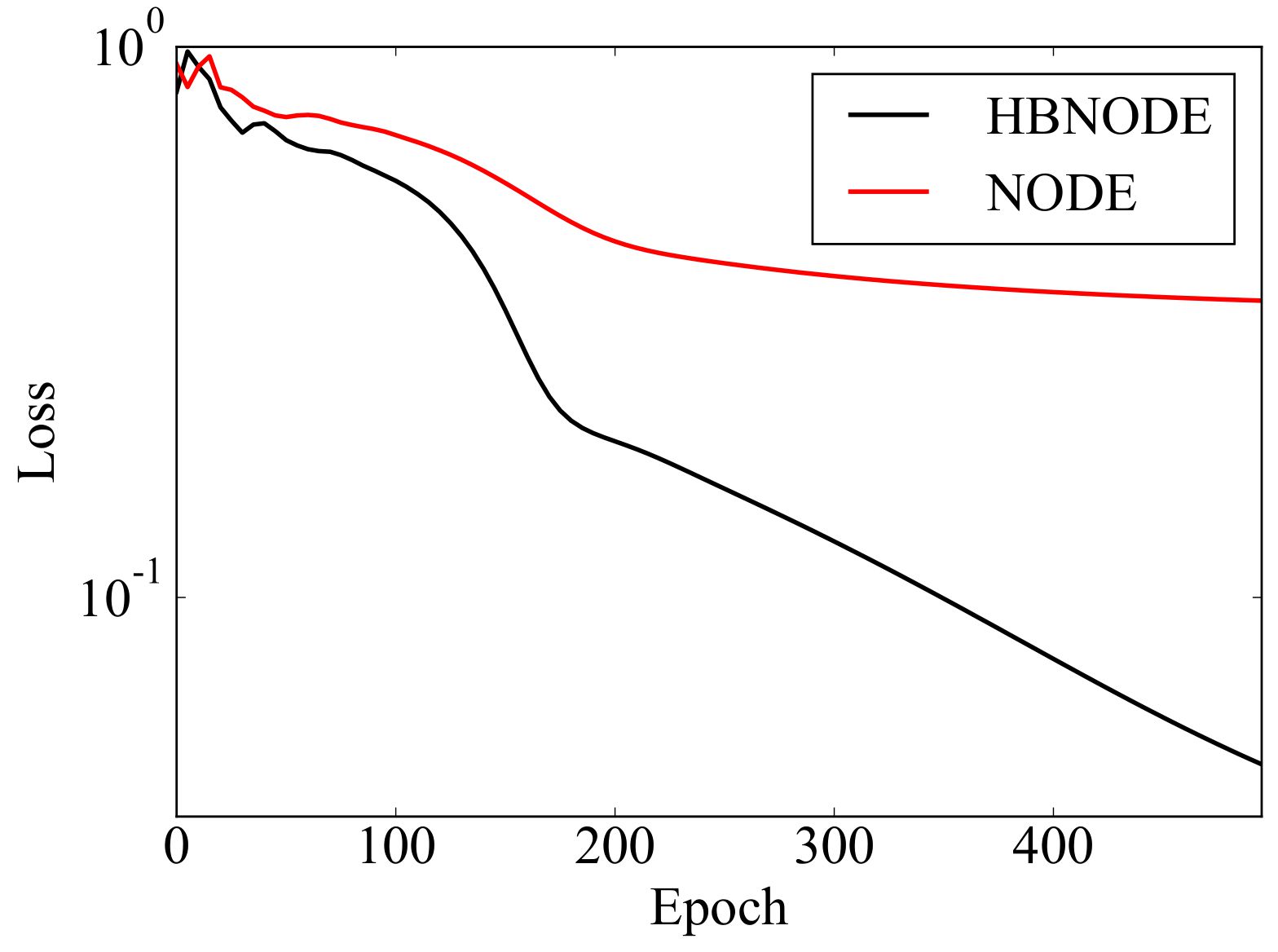}\\
 %{img/final/kpp/compare_val_loss.pdf}
 (a) Training loss &
 (b) Validation loss
 \end{tabular}
 \caption{Comparison of the training and validation loss of NODE and HBNODE for learning ROMs for the KPP model. NODE is unable to make progress due to a rapidly vanishing gradient, impeding learning long-range dependencies.}
\label{fig:kpp_loss}
\end{figure}
We compare the prediction of NODE and HBNODE against ground truth in Fig.~\ref{fig:kpp_mode_prediction}, and we see that HBNODE performs remarkably better than NODE in predicting the dynamics. In particular, HBNODE is able to properly capture the oscillation dynamics of the modes, unlike NODE.  Figure~\ref{fig:kpp_loss} shows that HBNODE has a much smaller training and validation loss than that of NODE.

\subsection{Euler equations for fluids modeling}
\begin{figure}[!ht]
     \centering
     \includegraphics[width=0.8\textwidth]{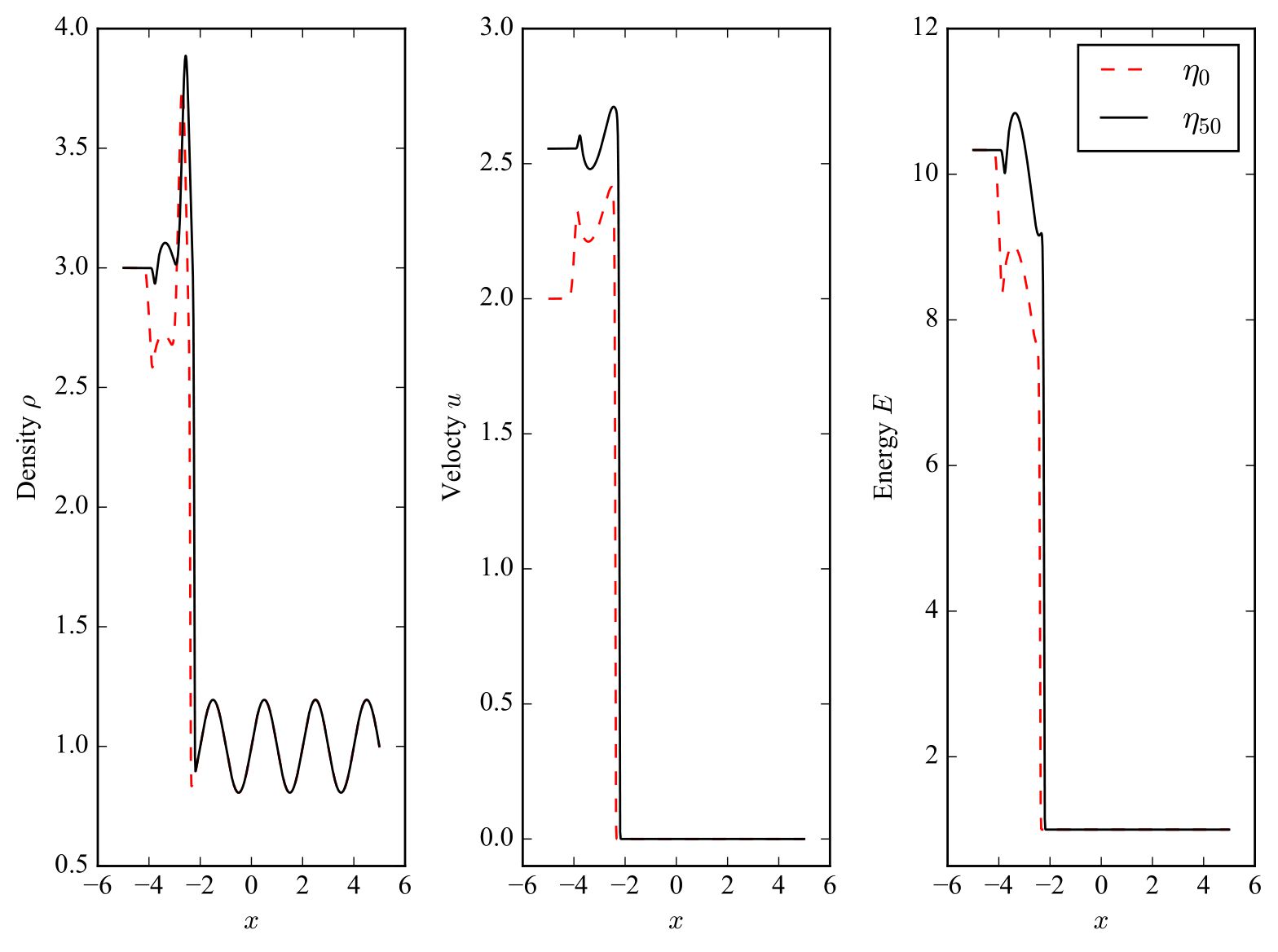} %{img/final/ee/ee_params.pdf}
     \caption{The Euler Equations data at time step $t=50$ with two different initial parameters, $\eta_0$ and $\eta_{50}$. Variations in the parameter $\eta$ produce widely varying dynamics and as a result varying POD modes. The objective of this task is to predict the dynamics for unseen parameters $\eta$ using a set of training parameters $\eta_{\mathrm{train}}$. 
     }
     \label{fig:ee_params}
\end{figure}
  We further consider learning reduced-order models for simulating the Euler equations, where the dataset is obtained by simulating the full-order model presented in Section~\ref{ssec:ee} with a discrete ensemble of parameters $\bs{\eta}_1,\ldots,\bs{\eta}_M$ with $M=100$, over 180 timesteps. Two different parameters $\bs{\eta}$ values can produce rather different dynamics, as evidenced in Fig.~\ref{fig:ee_params}. 
The Euler equations data is unique in the sense that it may be segmented based on these initial conditions. The ROM is generated by taking the dominant $8$ POD modes for each parameter $\bs{\eta}_i$ on the time interval from $t=0$ to $180$. This data is shuffled randomly among the initial parameter $\bs{\eta}_i$ to no longer increase sequentially. The average relative information content across all $100$ values of $\bs{\eta}$ is $\sim95\%$. 

In this task, we train the machine learning pipeline shown in Fig.~\ref{fig:rnn_node} for learning multi-input-multi-output dynamics. The training dataset comprises the dominant $8$ POD modes for each of the training parameters among $\eta_1,\ldots,\eta_{\mathrm{train}}$. We use $90$ of the $100$ parameters, $\eta_1,\ldots,\eta_{90}$, for training and the rest for validation. The training input consists of the dominant $8$ POD modes for each of the $90$ training parameters on the time interval from $t=0$ to $150$. The training labels consist of the dominant $8$ POD modes for each of the $90$ training parameters time steps from $t=151$ to $180$. The validation dataset is composed of the validation parameters $\eta_{91},\ldots,\eta_{100}$. The validation input and labels are segmented using the same intervals as the training data.

The model uses a GHBNODE component with a hyperbolic tangent activation function. All other experimental settings are the same as in the KPP dataset, and the tuned hyper-parameters are listed in table \ref{Table:full_ee_hyper}. The NODE and HBNODE models are trained and validated over the same data shuffling.

\begin{table}[!ht]
\fontsize{9.0}{9.0}\selectfont
\centering
\begin{threeparttable}
\begin{tabular}{cc}
\toprule[1.5pt]
\ \ \ \ \ \ \ \ \qquad {\bf Hyper-parameter} \ \ \ \ \ \ \ \ \qquad  & \ \ \ \ \ \ \ \ \qquad Value \ \ \ \ \ \ \ \ \qquad \cr
\midrule[1.0pt]
     Layers & 6 \\
     Hidden layers & 16 \\
     Learning rate & .01 \\
     Epochs & 100 \\
\bottomrule[1.5pt]
\end{tabular}
\end{threeparttable}
\caption{The hyperparameters of NODE and NODE for the learning ROMs of Euler equations.}
\label{Table:full_ee_hyper}
\end{table}

\subsubsection{Results and comparison to existing models}
\begin{figure}[!ht]
     \centering
     \begin{tabular}{cc}
     \includegraphics[clip, trim=0.01cm 0.01cm 0.01cm 0.01cm, width=0.46\columnwidth]{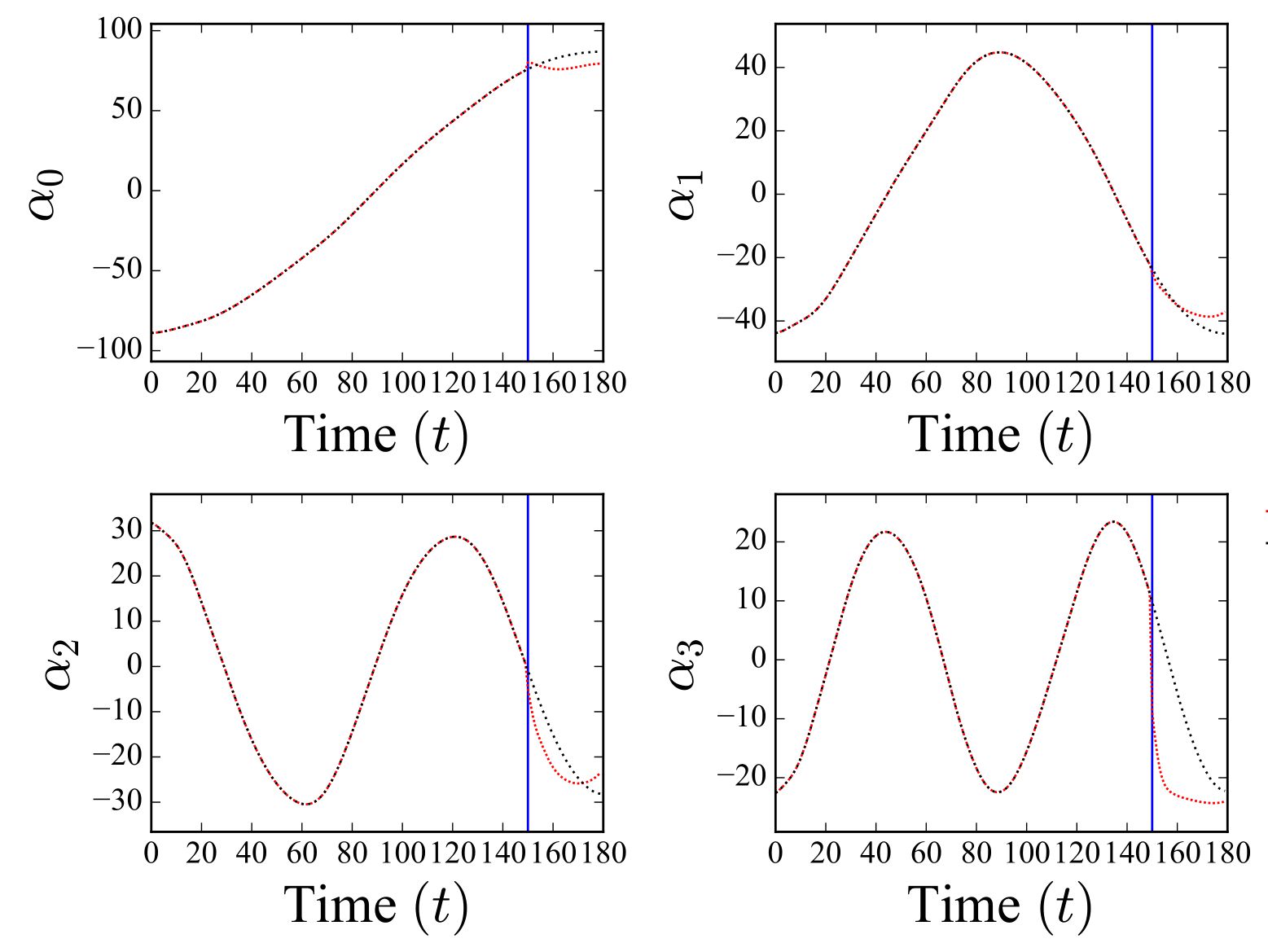}&
     %{img/final/ee/ee_param_node_mode_pred.pdf}
     \includegraphics[clip, trim=0.01cm 0.01cm 0.01cm 0.01cm, width=0.46\columnwidth]{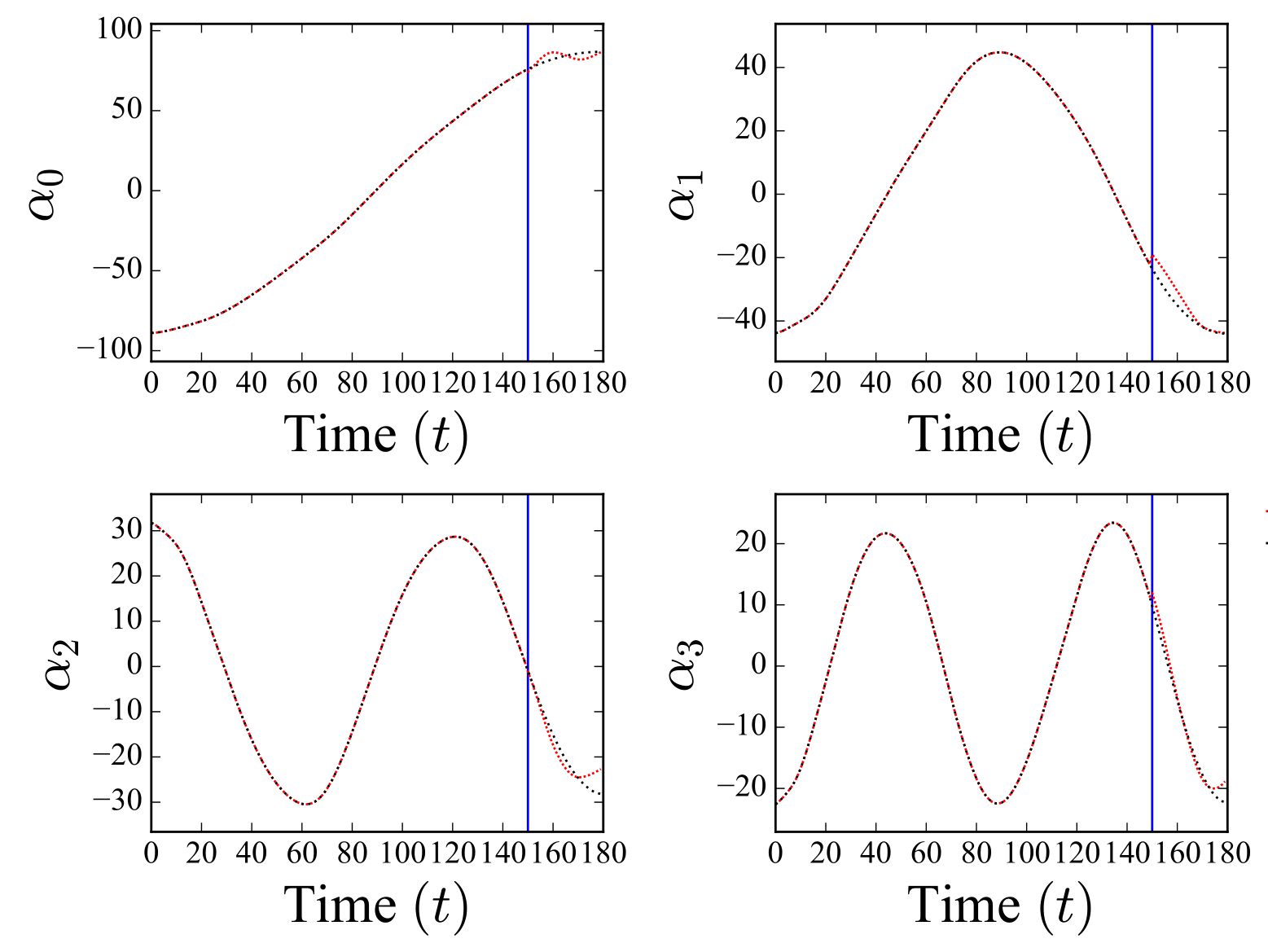}\\
     %{img/final/ee/ee_param_ghbnode_mode_pred.pdf}
     (a) NODE Training Mode Recapture &
     (b) GHBNODE Training Mode Recapture
     \end{tabular}
     \caption{For a randomly selected initial parameter $\eta_{\mathrm{train}}$, we plot the dominant four modes and their predicted data using NODE and GHBNODE. The blue vertical line separates the input and output data. The ground truth is in black, while the prediction data is in dashed red. GHBNODE can learn the dynamics of $\alpha_2$ and $\alpha_1$ for the parameterized data significantly better than NODE. This is evidenced by the fact that NODE predicts the inflection point of $\alpha_2$ too early.
     }
     \label{fig:ee_train}
\end{figure}

\begin{figure}[!ht]
     \centering
     \begin{tabular}{cc}
     \includegraphics[clip, trim=0.01cm 0.01cm 0.01cm 0.01cm, width=0.46\columnwidth]{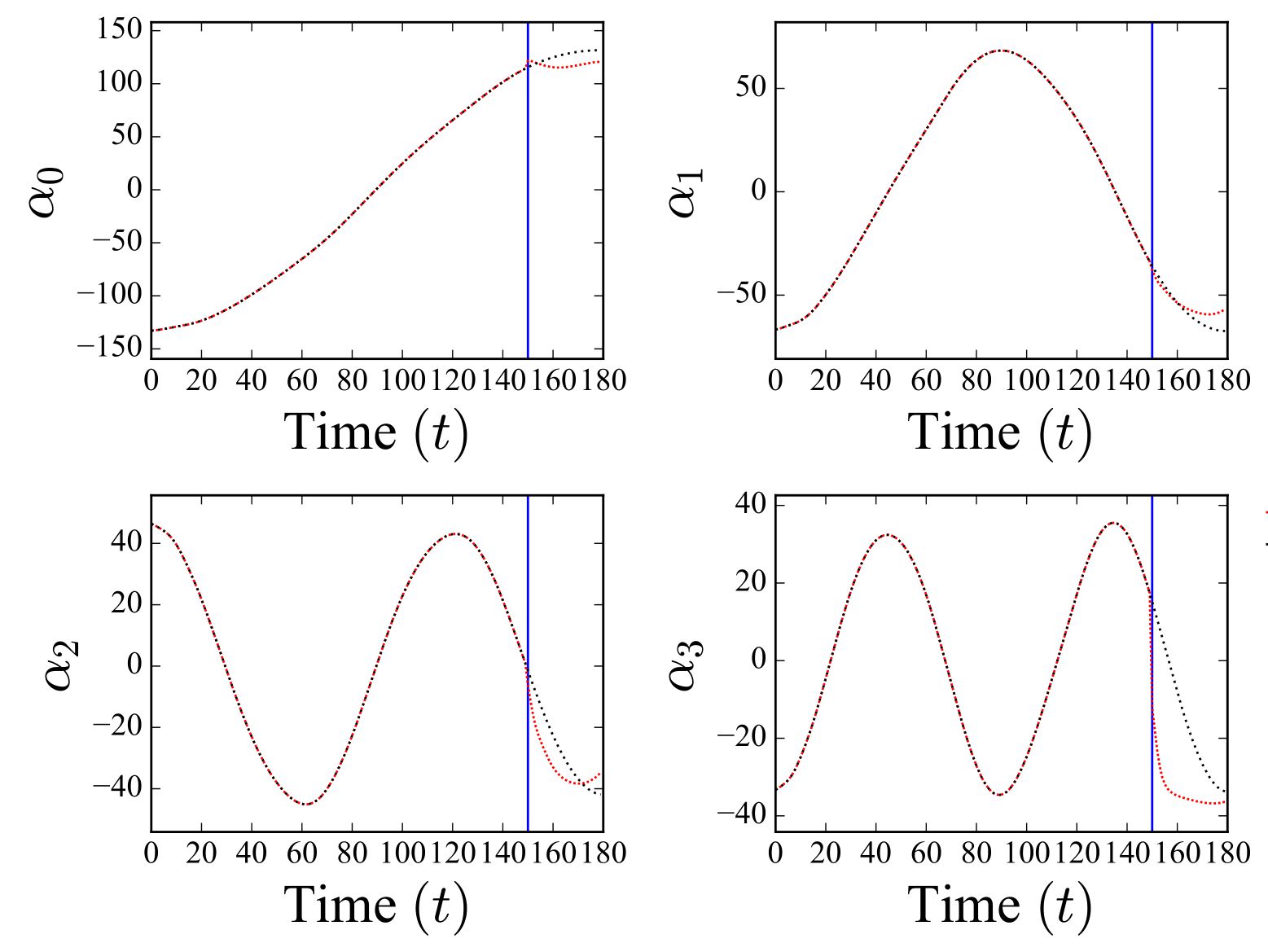}&
     %{img/final/ee/ee_param_node_mode_pred_val.pdf}
     \includegraphics[clip, trim=0.01cm 0.01cm 0.01cm 0.01cm, width=0.46\columnwidth]{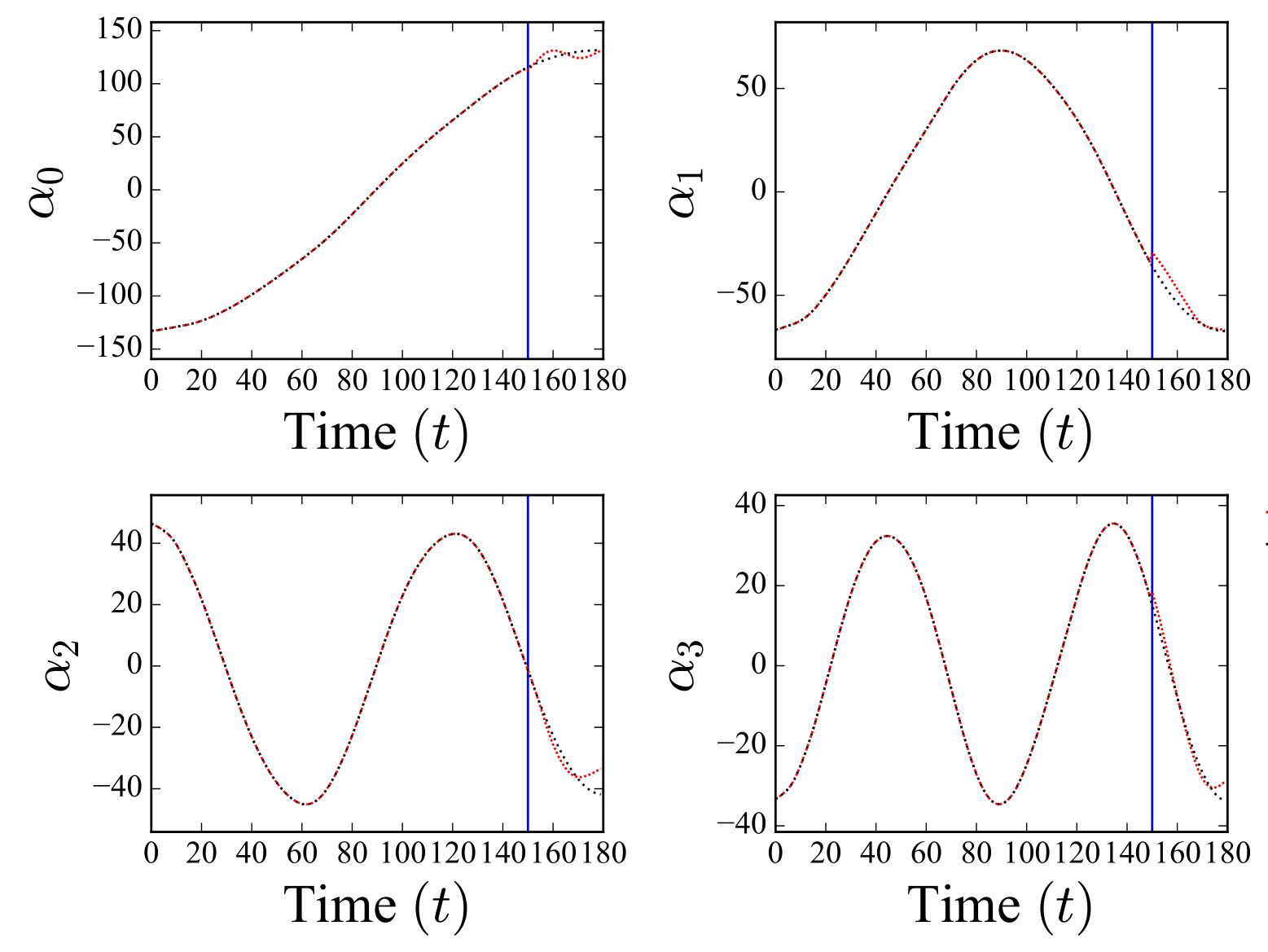}\\
     %{img/final/ee/ee_param_ghbnode_mode_pred_val.pdf}
     (a) NODE Validation Mode Recapture &
     (b) GHBNODE Validation Mode Recapture
     \end{tabular}
     \caption{For the randomly selected initial parameter $\eta_{\mathrm{valid}}$, we plot the dominant four modes and their predicted data using NODE and GHBNODE. The blue vertical line separates the input and output data. The ground truth is in black, while the prediction data is in dashed red. HBNODE can predict the data for unseen parameterizations much better than NODE.}
     \label{fig:ee_valid}
\end{figure}

We compare the prediction of NODE and GHBNODE for a randomly selected parameter $\eta_{\mathrm{train}}$ from the training set and a randomly selected parameter $\eta_{\mathrm{valid}}$ from the validation set. The modes for the training parameter $\eta_{\mathrm{train}}$ are shown in Fig.~\ref{fig:ee_train}, and the modes for the validation parameter $\eta_{\mathrm{valid}}$ are shown in Fig.~\ref{fig:ee_valid}. We observe that the POD modes for the parameter $\eta_{\mathrm{train}}$ in Fig.~\ref{fig:ee_train} differ from those for $\eta_{\mathrm{valid}}$ primarily in amplitude rather than shape. As a result, a poor prediction model will have a sudden discontinuity between the input and prediction values. The transition point between the input and the prediction is indicated in Fig.~\ref{fig:ee_train} and Fig.~\ref{fig:ee_valid} by the vertical blue line.

Figure~\ref{fig:ee_train} shows the dominant $4$ POD modes for $\eta_{\mathrm{train}}$ from the training set. The predictive capabilities of GHBNODE significantly outperform the NODE model. In particular, for $\alpha_2$ and $\alpha_3$, we observe that the NODE has a large jump discontinuity at the transition between the input and prediction. The GHBNODE modes are smoother in the transition region, which indicates the ability of the GHBNODE model to distinguish between separate parameters.

In Figure~\ref{fig:ee_valid}, we observe the same characteristics for NODE and GHBNODE. NODE is unable to accurately predict the output for the parameter $\eta_{\mathrm{valid}}$ from the validation set. In particular, for $\alpha_3$ of the validation parameter, the NODE prediction is even less smooth than that of the training parameter shown in Fig.~\ref{fig:ee_train}. In this experiment, we observe that NODE is unable to distinguish data with varying parameters as accurately as GHBNODE.

\section{Concluding Remarks}\label{sec:conclusion}
This paper employs the recently developed HBNODEs and their generalization for learning POD coefficients for model reduction. We analyze through simple linearized models and empirically verify the advantages of HBNODEs over existing NODEs. In particular, HBNODEs enjoy the following advantages that imply practical benefits for learning POD-based ROMs, including 1) The deep learning model is continuous-depth, providing flexibility in learning irregularly-sampled time series and faithful to the continuous profiling of the underlying physical models. 2) Both the forward and adjoint ODEs of HBNODEs are of the heavy-ball style, accelerating both training and testing of the machine learning procedure. And 3) HBNODEs can learn long-range dependencies effectively, capturing intrinsic patterns from data. There are numerous avenues for future works, and two particular interesting directions in our mind are 1) Improving HBNODEs, particularly replacing the fine-tuned or learned damping parameter with an adaptive one that are motivated by certain optimization algorithms with adaptive momentum \cite{wang2020scheduled,sun2021training,wang2020stochastic}, and 2) Applying HBNODE-based ROMs to model reduction arising from scientific challenges, especially when we do not have the ground truth governing equation of the dynamical systems.

\section{Data Availability Statement}
All data and code related to this paper are available at \url{https://github.com/JustinBakerMath/pod_hbnode/}.
%will be made available once the paper is accepted.

\section{Acknowledgement}
This material is based on research sponsored by NSF grants DMS-1848508, DMS-1924935, DMS-1952339, DMS-2110145, DMS-2111117, DMS-2152762, and DMS-2208361, DOE grant DE-SC0021142 and DE-SC0023490, and AFOSR FA9550-20-1-0338. We also acknowledge support from a seed grant from the College of Science at the University of Utah.

%\bibliographystyle{plain}
%\bibliography{refs.bib,refs2.bib}

\begin{thebibliography}{10}

\bibitem{antoulas_approximation_2005}
A.~Antoulas.
\newblock {\em Approximation of {Large}-{Scale} {Dynamical} {Systems}}.
\newblock Advances in {Design} and {Control}. Society for Industrial and
  Applied Mathematics, January 2005.

\bibitem{antoulas_interpolatory_2010}
Athanasios~C. Antoulas, Christopher~A. Beattie, and Serkan Gugercin.
\newblock Interpolatory {Model} {Reduction} of {Large}-{Scale} {Dynamical}
  {Systems}.
\newblock In Javad Mohammadpour and Karolos~M. Grigoriadis, editors, {\em
  Efficient {Modeling} and {Control} of {Large}-{Scale} {Systems}}, pages
  3--58. Springer US, Boston, MA, 2010.

\bibitem{github-animation}
Justin Baker, Elena Cherkaev, Akil Narayan, and Bao Wang.
\newblock Learning pod of complex dynamics usingheavy-ball neural odes:
  Animations.
\newblock
  \url{https://www.github.com/JustinBakerMath/pod_hbnode/blob/master/README.md#animations}.

\bibitem{bengio1994learning}
Yoshua Bengio, Patrice Simard, and Paolo Frasconi.
\newblock Learning long-term dependencies with gradient descent is difficult.
\newblock {\em IEEE Transactions on Neural Networks}, 5(2):157--166, 1994.

\bibitem{benner_survey_2015}
P.~Benner, S.~Gugercin, and K.~Willcox.
\newblock A {Survey} of {Projection}-{Based} {Model} {Reduction} {Methods} for
  {Parametric} {Dynamical} {Systems}.
\newblock {\em SIAM Review}, 57(4):483--531, January 2015.

\bibitem{benner2015survey}
Peter Benner, Serkan Gugercin, and Karen Willcox.
\newblock A survey of projection-based model reduction methods for parametric
  dynamical systems.
\newblock {\em SIAM review}, 57(4):483--531, 2015.

\bibitem{berkooz1993proper}
Gal Berkooz, Philip Holmes, and John~L Lumley.
\newblock The proper orthogonal decomposition in the analysis of turbulent
  flows.
\newblock {\em Annual review of fluid mechanics}, 25(1):539--575, 1993.

\bibitem{adjoint}
L.~Bittner.
\newblock {L. S. Pontryagin, V. G. Boltyanskii, R. V. Gamkrelidze, E. F.
  Mishechenko}, the mathematical theory of optimal processes. {VIII} + 360 {S.
  New York/London 1962. John Wiley \& Sons. Preis} 90/–.
\newblock {\em ZAMM - Journal of Applied Mathematics and Mechanics /
  Zeitschrift für Angewandte Mathematik und Mechanik}, 43(10-11):514--515,
  1963.

\bibitem{chen2018neural}
Ricky T.~Q. Chen, Yulia Rubanova, Jesse Bettencourt, and David~K Duvenaud.
\newblock Neural ordinary differential equations.
\newblock In S.~Bengio, H.~Wallach, H.~Larochelle, K.~Grauman, N.~Cesa-Bianchi,
  and R.~Garnett, editors, {\em Advances in Neural Information Processing
  Systems}, volume~31. Curran Associates, Inc., 2018.

\bibitem{cho2014learning}
Kyunghyun Cho, Bart Van~Merri{\"e}nboer, Caglar Gulcehre, Dzmitry Bahdanau,
  Fethi Bougares, Holger Schwenk, and Yoshua Bengio.
\newblock Learning phrase representations using rnn encoder-decoder for
  statistical machine translation.
\newblock {\em arXiv preprint arXiv:1406.1078}, 2014.

\bibitem{cohen1983absolute}
Michael~A. Cohen and Stephen Grossberg.
\newblock Absolute stability of global pattern formation and parallel memory
  storage by competitive neural networks.
\newblock {\em IEEE Transactions on Systems, Man, and Cybernetics},
  SMC-13(5):815--826, 1983.

\bibitem{craster2009dynamics}
Richard~V Craster and Omar~K Matar.
\newblock Dynamics and stability of thin liquid films.
\newblock {\em Reviews of modern physics}, 81(3):1131, 2009.

\bibitem{DORMAND198019}
J.R. Dormand and P.J. Prince.
\newblock A family of embedded runge-kutta formulae.
\newblock {\em Journal of Computational and Applied Mathematics}, 6(1):19--26,
  1980.

\bibitem{NEURIPS2019_21be9a4b}
Emilien Dupont, Arnaud Doucet, and Yee~Whye Teh.
\newblock Augmented neural odes.
\newblock In {\em Advances in Neural Information Processing Systems},
  volume~32. Curran Associates, Inc., 2019.

\bibitem{dutta2021data}
Sourav Dutta, Peter Rivera-Casillas, Orie~M Cecil, Matthew~W Farthing, Emma
  Perracchione, and Mario Putti.
\newblock Data-driven reduced order modeling of environmental hydrodynamics
  using deep autoencoders and neural odes.
\newblock {\em arXiv preprint arXiv:2107.02784}, 2021.

\bibitem{dutta2021neural}
Sourav Dutta, Peter Rivera-Casillas, and Matthew~W Farthing.
\newblock Neural ordinary differential equations for data-driven reduced order
  modeling of environmental hydrodynamics.
\newblock {\em arXiv preprint arXiv:2104.13962}, 2021.

\bibitem{germano1991dynamic}
Massimo Germano, Ugo Piomelli, Parviz Moin, and William~H Cabot.
\newblock A dynamic subgrid-scale eddy viscosity model.
\newblock {\em Physics of Fluids A: Fluid Dynamics}, 3(7):1760--1765, 1991.

\bibitem{gugercin_survey_2004}
Serkan Gugercin and Athanasios~C. Antoulas.
\newblock A {Survey} of {Model} {Reduction} by {Balanced} {Truncation} and
  {Some} {New} {Results}.
\newblock {\em International Journal of Control}, 77(8):748--766, May 2004.

\bibitem{harten_upstream_1983}
Amiram Harten, Peter~D. Lax, and Bram~van Leer.
\newblock On {Upstream} {Differencing} and {Godunov}-{Type} {Schemes} for
  {Hyperbolic} {Conservation} {Laws}.
\newblock {\em SIAM Review}, 25(1):35--61, 1983.

\bibitem{he2016identity}
Kaiming He, Xiangyu Zhang, Shaoqing Ren, and Jian Sun.
\newblock Identity mappings in deep residual networks.
\newblock In {\em European Conference on Computer Vision}, pages 630--645,
  2016.

\bibitem{LSTM}
S.~Hochreiter and J.~Schmidhuber.
\newblock Long short-term memory.
\newblock {\em Neural Computation}, 9(8):1735--1780, 1997.

\bibitem{hochreiter1997long}
Sepp Hochreiter and J{\"u}rgen Schmidhuber.
\newblock Long short-term memory.
\newblock {\em Neural Computation}, 9(8):1735--1780, 1997.

\bibitem{jacot2018neural}
Arthur Jacot, Franck Gabriel, and Cl{\'e}ment Hongler.
\newblock Neural tangent kernel: convergence and generalization in neural
  networks.
\newblock In {\em Proceedings of the 32nd International Conference on Neural
  Information Processing Systems}, pages 8580--8589, 2018.

\bibitem{kani2017dr}
J~Nagoor Kani and Ahmed~H Elsheikh.
\newblock Dr-rnn: A deep residual recurrent neural network for model reduction.
\newblock {\em arXiv preprint arXiv:1709.00939}, 2017.

\bibitem{kani2019reduced}
J~Nagoor Kani and Ahmed~H Elsheikh.
\newblock Reduced-order modeling of subsurface multi-phase flow models using
  deep residual recurrent neural networks.
\newblock {\em Transport in Porous Media}, 126(3):713--741, 2019.

\bibitem{kingma2013auto1}
Diederik~P Kingma and Max Welling.
\newblock Auto-encoding variational bayes.
\newblock {\em arXiv preprint arXiv:1312.6114}, 2013.

\bibitem{10.1143/PTPS.64.346}
Yoshiki Kuramoto.
\newblock {Diffusion-Induced Chaos in Reaction Systems}.
\newblock {\em Progress of Theoretical Physics Supplement}, 64:346--367, 02
  1978.

\bibitem{kurganov_adaptive_2007}
Alexander Kurganov, Guergana Petrova, and Bojan Popov.
\newblock Adaptive {Semidiscrete} {Central}-{Upwind} {Schemes} for {Nonconvex}
  {Hyperbolic} {Conservation} {Laws}.
\newblock {\em SIAM Journal on Scientific Computing}, 29(6):2381--2401, January
  2007.

\bibitem{lechner2020learning}
Mathias Lechner and Ramin Hasani.
\newblock Learning long-term dependencies in irregularly-sampled time series.
\newblock {\em arXiv preprint arXiv:2006.04418}, 2020.

\bibitem{LIANG2002527}
Y.C. Liang, H.P. Lee, S.P. Lim, W.Z. Lin, K.H. Lee, and C.G. Wu.
\newblock Proper orthogonal decomposition and its applications—part i:
  Theory.
\newblock {\em Journal of Sound and Vibration}, 252(3):527--544, 2002.

\bibitem{lui2019construction}
Hugo~FS Lui and William~R Wolf.
\newblock Construction of reduced-order models for fluid flows using deep
  feedforward neural networks.
\newblock {\em Journal of Fluid Mechanics}, 872:963--994, 2019.

\bibitem{ma2018model}
Chao Ma, Jianchun Wang, et~al.
\newblock Model reduction with memory and the machine learning of dynamical
  systems.
\newblock {\em arXiv preprint arXiv:1808.04258}, 2018.

\bibitem{mannarino2014nonlinear}
Andrea Mannarino and Paolo Mantegazza.
\newblock Nonlinear aeroelastic reduced order modeling by recurrent neural
  networks.
\newblock {\em Journal of Fluids and Structures}, 48:103--121, 2014.

\bibitem{massaroli2020dissecting}
Stefano Massaroli, Michael Poli, Jinkyoo Park, Atsushi Yamashita, and Hajime
  Asama.
\newblock Dissecting neural odes.
\newblock In H.~Larochelle, M.~Ranzato, R.~Hadsell, M.~F. Balcan, and H.~Lin,
  editors, {\em Advances in Neural Information Processing Systems}, volume~33,
  pages 3952--3963. Curran Associates, Inc., 2020.

\bibitem{maulik2021reduced}
Romit Maulik, Bethany Lusch, and Prasanna Balaprakash.
\newblock Reduced-order modeling of advection-dominated systems with recurrent
  neural networks and convolutional autoencoders.
\newblock {\em Physics of Fluids}, 33(3):037106, 2021.

\bibitem{https://doi.org/10.1002/fld.4684}
M.~Mohebujjaman, L.G. Rebholz, and T.~Iliescu.
\newblock Physically constrained data-driven correction for reduced-order
  modeling of fluid flows.
\newblock {\em International Journal for Numerical Methods in Fluids},
  89(3):103--122, 2019.

\bibitem{moin1998direct}
Parviz Moin and Krishnan Mahesh.
\newblock Direct numerical simulation: a tool in turbulence research.
\newblock {\em Annual review of fluid mechanics}, 30(1):539--578, 1998.

\bibitem{mou2020data}
Changhong Mou, Honghu Liu, David~R Wells, and Traian Iliescu.
\newblock Data-driven correction reduced order models for the quasi-geostrophic
  equations: A numerical investigation.
\newblock {\em International Journal of Computational Fluid Dynamics},
  34(2):147--159, 2020.

\bibitem{murata_fukami_fukagata_2020}
Takaaki Murata, Kai Fukami, and Koji Fukagata.
\newblock Nonlinear mode decomposition with convolutional neural networks for
  fluid dynamics.
\newblock {\em Journal of Fluid Mechanics}, 882:A13, 2020.

\bibitem{MomentumRNN}
Tan Nguyen, Richard Baraniuk, Andrea Bertozzi, Stanley Osher, and Bao Wang.
\newblock {MomentumRNN}: Integrating momentum into recurrent neural networks.
\newblock In H.~Larochelle, M.~Ranzato, R.~Hadsell, M.~F. Balcan, and H.~Lin,
  editors, {\em Advances in Neural Information Processing Systems}, volume~33,
  pages 1924--1936. Curran Associates, Inc., 2020.

\bibitem{norcliffe2020_sonode}
Alexander Norcliffe, Cristian Bodnar, Ben Day, Nikola Simidjievski, and Pietro
  Li\'{o}.
\newblock On second order behaviour in augmented neural odes.
\newblock In H.~Larochelle, M.~Ranzato, R.~Hadsell, M.~F. Balcan, and H.~Lin,
  editors, {\em Advances in Neural Information Processing Systems}, volume~33,
  pages 5911--5921. Curran Associates, Inc., 2020.

\bibitem{pascanu2013difficulty}
Razvan Pascanu, Tomas Mikolov, and Yoshua Bengio.
\newblock On the difficulty of training recurrent neural networks.
\newblock In {\em International Conference on Machine Learning}, pages
  1310--1318, 2013.

\bibitem{pearson1901liii}
Karl Pearson.
\newblock Liii. on lines and planes of closest fit to systems of points in
  space.
\newblock {\em The London, Edinburgh, and Dublin philosophical magazine and
  journal of science}, 2(11):559--572, 1901.

\bibitem{polyak1964some}
Boris~T Polyak.
\newblock Some methods of speeding up the convergence of iteration methods.
\newblock {\em USSR Computational Mathematics and Mathematical Physics},
  4(5):1--17, 1964.

\bibitem{rom_node}
Carlos J.~G. Rojas, Andreas Dengel, and Mateus~Dias Ribeiro.
\newblock Reduced-order {Model} for {Fluid} {Flows} via {Neural} {Ordinary}
  {Differential} {Equations}.
\newblock {\em arXiv:2102.02248 [physics]}, February 2021.
\newblock arXiv: 2102.02248.

\bibitem{rosenblatt1961principles}
Frank Rosenblatt.
\newblock Principles of neurodynamics. perceptrons and the theory of brain
  mechanisms.
\newblock Technical report, Cornell Aeronautical Lab Inc Buffalo NY, 1961.

\bibitem{latentODE}
Yulia Rubanova, Ricky T.~Q. Chen, and David~K Duvenaud.
\newblock Latent ordinary differential equations for irregularly-sampled time
  series.
\newblock In {\em Advances in Neural Information Processing Systems},
  volume~32. Curran Associates, Inc., 2019.

\bibitem{san2017neural}
Omer San and Romit Maulik.
\newblock Neural network closures for nonlinear model order reduction.
\newblock {\em arXiv preprint arXiv:1705.08532}, 2017.

\bibitem{SAN2018681}
Omer San and Romit Maulik.
\newblock Machine learning closures for model order reduction of thermal
  fluids.
\newblock {\em Applied Mathematical Modelling}, 60:681--710, 2018.

\bibitem{SAN2019271}
Omer San, Romit Maulik, and Mansoor Ahmed.
\newblock An artificial neural network framework for reduced order modeling of
  transient flows.
\newblock {\em Communications in Nonlinear Science and Numerical Simulation},
  77:271--287, 2019.

\bibitem{schmid_2010}
Peter~J. Schmid.
\newblock Dynamic mode decomposition of numerical and experimental data.
\newblock {\em Journal of Fluid Mechanics}, 656:5–28, 2010.

\bibitem{10.2307/2100687}
G.~I. Sivashinsky.
\newblock On flame propagation under conditions of stoichiometry.
\newblock {\em SIAM Journal on Applied Mathematics}, 39(1):67--82, 1980.

\bibitem{SIVASHINSKY19771177}
G.I. Sivashinsky.
\newblock Nonlinear analysis of hydrodynamic instability in laminar flames—i.
  derivation of basic equations.
\newblock {\em Acta Astronautica}, 4(11):1177--1206, 1977.

\bibitem{sun2021training}
Tao Sun, Huaming Ling, Zuoqiang Shi, Dongsheng Li, and Bao Wang.
\newblock Training deep neural networks with adaptive momentum inspired by the
  quadratic optimization.
\newblock {\em arXiv preprint arXiv:2110.09057}, 2021.

\bibitem{wang2020scheduled}
Bao Wang, Tan~M Nguyen, Andrea~L Bertozzi, Richard~G Baraniuk, and Stanley~J
  Osher.
\newblock Scheduled restart momentum for accelerated stochastic gradient
  descent.
\newblock {\em arXiv preprint arXiv:2002.10583}, 2020.

\bibitem{wang2021does}
Bao Wang, Hedi Xia, Tan Nguyen, and Stanley Osher.
\newblock How does momentum benefit deep neural networks architecture design? a
  few case studies.
\newblock {\em arXiv preprint arXiv:2110.07034}, 2021.

\bibitem{wang2020stochastic}
Bao Wang and Qiang Ye.
\newblock Stochastic gradient descent with nonlinear conjugate gradient-style
  adaptive momentum.
\newblock {\em arXiv preprint arXiv:2012.02188}, 2020.

\bibitem{7572934}
Mingliang Wang, Han-Xiong Li, Xin Chen, and Yun Chen.
\newblock Deep learning-based model reduction for distributed parameter
  systems.
\newblock {\em IEEE Transactions on Systems, Man, and Cybernetics: Systems},
  46(12):1664--1674, 2016.

\bibitem{HBNODE:2021}
Hedi Xia, Vai Suliafu, Hangjie Ji, Tan Nguyen, Andrea Bertozzi, Stanley Osher,
  and Bao Wang.
\newblock Heavy ball neural ordinary differential equation.
\newblock In {\em Advances in Neural Information Processing Systems},
  volume~34. Curran Associates, Inc., 2021.

\bibitem{you2007dynamic}
Donghyun You and Parviz Moin.
\newblock A dynamic global-coefficient subgrid-scale eddy-viscosity model for
  large-eddy simulation in complex geometries.
\newblock {\em Physics of Fluids}, 19(6):065110, 2007.

\end{thebibliography}

\end{document}